\documentclass[ijds,nonblindrev]{informs-ijds}

\OneAndAHalfSpacedXI

\usepackage{natbib}
 \bibpunct[, ]{(}{)}{,}{a}{}{,}%
 \def\bibfont{\small}%
\TheoremsNumberedThrough     
\ECRepeatTheorems

\EquationsNumberedThrough    

\MANUSCRIPTNO{}

\usepackage{subfigure}
\usepackage{bm}
\usepackage{wrapfig}
\usepackage[font=footnotesize]{caption}

\usepackage{mathtools}
\DeclarePairedDelimiter\ceil{\lceil}{\rceil}

\usepackage[ruled,vlined]{algorithm2e}
\usepackage{amsmath}
\usepackage[makeroom]{cancel}

\usepackage{makecell}
\DeclarePairedDelimiterX{\infdivx}[2]{(}{)}{%
  #1\;\delimsize\|\;#2%
}

\newcommand{\norm}[1]{\left\lVert#1\right\rVert}

\newcommand{\btheta}{\boldsymbol{\theta}}

\usepackage[ruled,vlined]{algorithm2e}
\usepackage{amsmath}
\usepackage[makeroom]{cancel}
\usepackage{makecell}

\begin{document}



\RUNTITLE{Xubo Yue, Maher Nouiehed, Raed Al Kontar}

\TITLE{GIFAIR-FL: A Framework for Group and Individual Fairness in Federated Learning}

\ARTICLEAUTHORS{%
\AUTHOR{Xubo Yue}
\AFF{Industrial \& Operations Engineering, University of Michigan, Ann Arbor, MI 48109, USA, \EMAIL{maxyxb@umich.edu}} 
\AUTHOR{Maher Nouiehed}
\AFF{Industrial Engineering and Management, American University of Beirut, Beirut, Lebanon, \EMAIL{mn102@aub.edu.lb}}
\AUTHOR{Raed Al Kontar}
\AFF{Industrial \& Operations Engineering, University of Michigan, Ann Arbor, MI 48109, USA, \EMAIL{alkontar@umich.edu}}
} 

\ABSTRACT{%
In this paper we propose \texttt{GIFAIR-FL}: a framework that imposes \textbf{G}roup and \textbf{I}ndividual \textbf{FAIR}ness to \textbf{F}ederated \textbf{L}earning settings. By adding a regularization term, our algorithm penalizes the spread in the loss of client groups to drive the optimizer to fair solutions. Our framework \texttt{GIFAIR-FL} can accommodate both global and personalized settings. Theoretically,  we show convergence in non-convex and strongly convex settings. Our convergence guarantees hold for both $i.i.d.$ and non-$i.i.d.$ data. To demonstrate the empirical performance of our algorithm, we apply our method to image classification and text prediction tasks. Compared to existing algorithms, our method shows improved fairness results while retaining superior or similar prediction accuracy.
}%



\maketitle

%


\section{Introduction}
\label{sec:intro}

A critical change is happening in today's Internet of Things (IoT). The computational power of edge devices is steadily increasing. AI chips are rapidly infiltrating the market, smart phones nowadays have compute power comparable to everyday use laptops \citep{phones}, Tesla just boasted that its autopilot system has computing power of more than 3000 MacBook pros \citep{Tesla} and small local computers such as Raspberry Pis have become common place in many applications especially manufacturing \citep{al2018cyber}. This opens a new paradigm for data analytics in IoT; one that exploits local computing power to process more of the user's data where it is created. This future of IoT has been recently termed as the ``The Internet of Federated Things (IoFT)'' \citep{kontar2021internet} where the term federated, refers to some autonomy for IoT devices and is inspired by the explosive recent interest in federated data science. 

\begin{wrapfigure}{r}{0.55\textwidth}
  \begin{center}
     \includegraphics[width=0.55\textwidth]{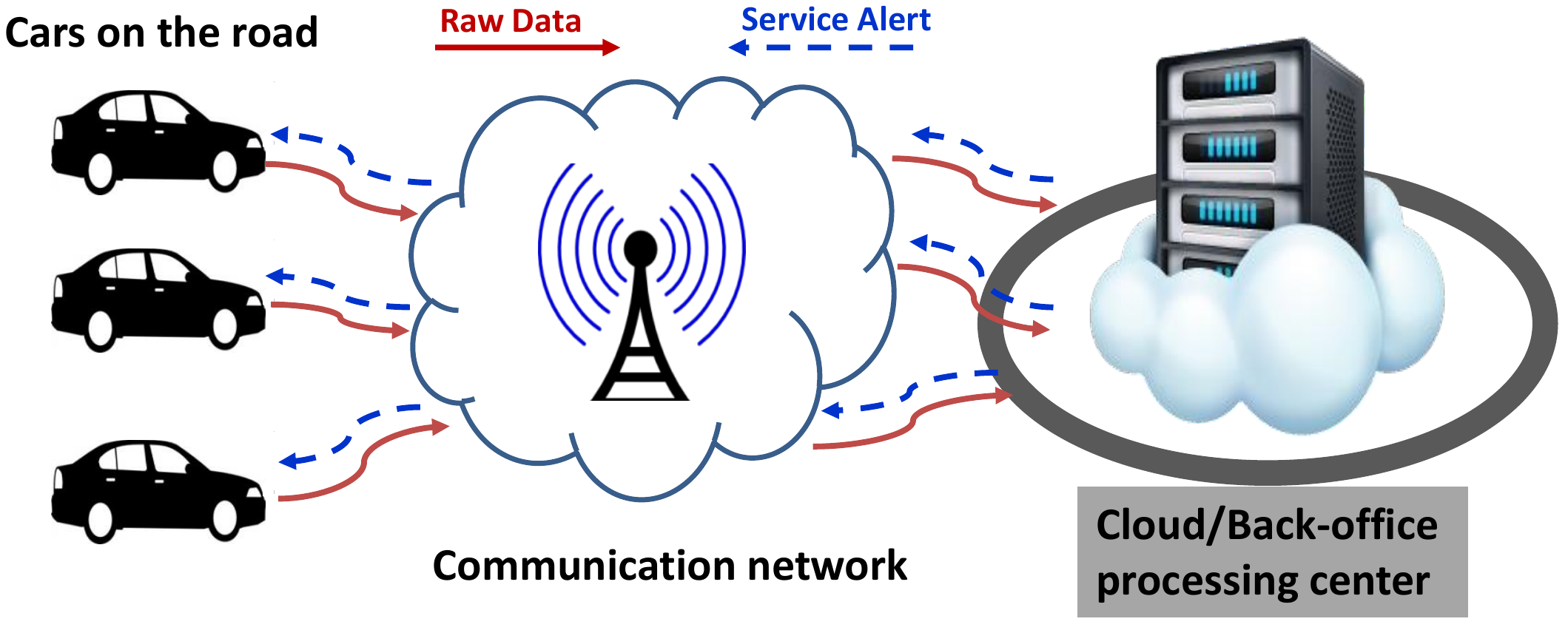}
  \end{center}
    \caption{Example of the traditional IoT-enabled System}
    \label{fig:present}
\end{wrapfigure}

To give a microcosm of current IoT and its future, consider the IoT teleservice system shown in Fig. \ref{fig:present}. Vehicles enrolled in this tele-service system often have their data in the form of condition monitoring signals uploaded to the cloud at regular intervals. The cloud acts as a data processing center that analyzes data for continuous improvement and to keep drivers informed about the health of their vehicles. IoT companies and services, such as Ford's SYNC and General Motors Onstar services, have long adopted this centralized approach to IoT. However, this state where data is amassed on the cloud yields significant challenges. The need to upload large amounts of data to the cloud incurs high communication and storage costs, demands large internet bandwidth \citep{jiang2020bacombo, yang2020federated}, and leads to latency in deployment as well as reliability risks due to unreliable connection \citep{zhang2020federated}. Further, such a system does not foster trust or privacy as users need to share their raw data which is often sensitive or confidential \citep{li2020review}.

With the increasing computational power of edge devices, the discussed challenges can be circumvented by moving part of the model learning to the edge. More specifically, rather than processing the data at the cloud, each device performs small local computations and only shares the minimum information needed to allow devices to borrow strength from each other and collaboratively extract knowledge to build smart analytics. In turn, such an approach (i) improves privacy as raw data is never shared, (ii) reduces cost and storage needs as less information is transmitted, (iii) enables learning parallelization and (iv) reduces latency in decisions as many decisions can now be achieved locally. Hereon we will use edge device and client interchangeably, also, the cloud or data processing center is referred to as the central server. 

\begin{figure*}[!htbp]
    \centering
    \centerline{\includegraphics[width=0.6\columnwidth]{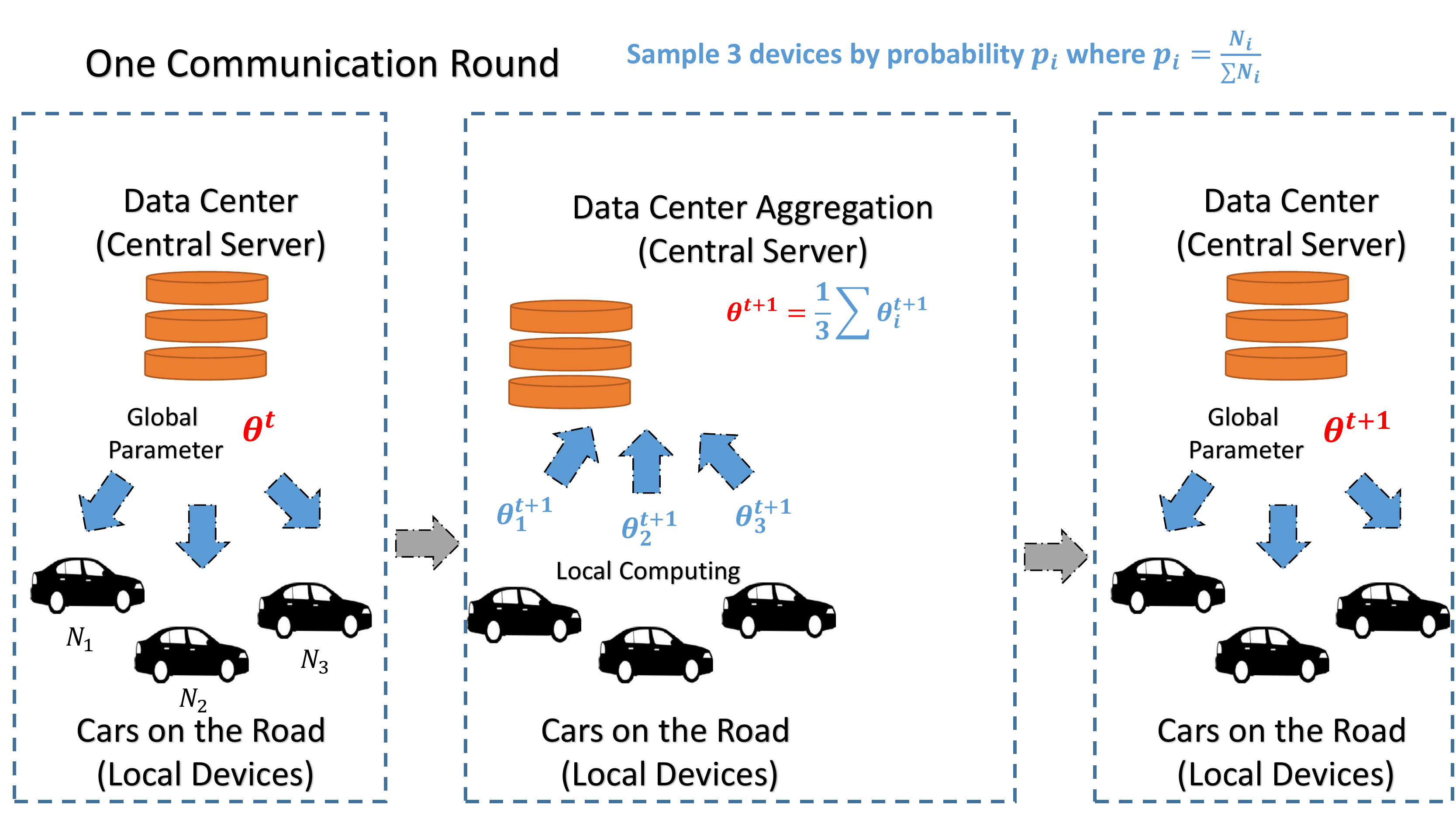}}
    \caption{Illustrative Example of FL with \texttt{FedAvg} }
    \label{fig:fed_example}
\end{figure*}

This idea of exploiting the computational power of edge devices by locally training models without recourse to data sharing gave rise to federated learning (FL). In particular, FL is a data analytics approach that allows distributed model learning without access to private data. Although the main concept of FL dates back a while ago, it was brought to the forefront of data science in 2017 by a team at Google which proposed Federated Averaging (\texttt{FedAvg}) \citep{mcmahan2017communication}. In \texttt{FedAvg}, a central server distributes the model architecture (e.g., neural network, linear model) and a global model parameter (e.g., model weights) to selected devices. Devices run local computations to update model parameters and send updated parameters to the server. The server then takes a weighted average of the resulting local parameters to update the global model. This whole process is termed as one communication round and the process is iterated over multiple rounds until an exit condition is met. Figure \ref{fig:fed_example} provides one illustrative example of \texttt{FedAvg}. Since then FL has seen immense success in various fields such as text prediction \citep{hard2018federated, ramaswamy2019federated}, Bayesian optimization \citep{dai2020federated, khodak2021federated}, Multi-fidelity modeling \citep{yue2021federated}, environment monitoring \citep{hu2018federated} and healthcare \citep{li2020review}. 

Over the last few years, literature has been proposed to improve the performance of FL algorithms; be it speeding up FL algorithms to enable faster convergence \citep{karimireddy2020scaffold, yuan2020federated}, tackling heterogeneous data both in size and distribution \citep{zhao2018federated, li2018federated, sattler2019robust, ghosh2019robust, li2019fedmd}, improving the parameter aggregation strategies at the central server \citep{wang2020federated}, designing personalized FL algorithms \citep{fallah2020personalized}, protecting federated systems from adversarial attacks \citep{bhagoji2019analyzing, wang2020attack}, and promoting fairness \citep{li2019fair}. Please refer to \cite{kontar2021internet} for a detailed literature review. Among those advances, fairness is a critical yet under-investigated area.

In the training phase of FL algorithms, devices with few data, limited bandwidth/memory, or unreliable connection may not be favored by conventional FL algorithms. For instance, as shown in Figure \ref{fig:fed_example}, \texttt{FedAvg} samples devices using the weight coefficient $p_k$ proportional to the sample size on the device $k$. Scant data on device $k$ will render $p_k$ insignificant and this device less favorable by the resulting global model. As a result, such device(s) can potentially incur higher error rate(s). This vicious cycle often relinquishes the opportunity of these devices to significantly contribute in the training process.  Indeed, many recent papers have shown the large variety in model performance across devices under FL \citep{jiang2019improving,hard2018federated,wang2019federated,smith2017federated, kairouz2019advances}, with some clients showing extremely bad model performance. Besides this aforementioned notion of individual fairness, group fairness also deserves attention in FL. As FL penetrates practical applications, it is important to achieve fair performance across groups of clients characterized by their gender, ethnicity, socio-economic status, geographic location, etc. Despite the importance of this notion of group fairness, unfortunately, \textit{no work exists along this line in FL}.

\paragraph{Contribution:} We propose a framework, \texttt{GIFAIR-FL}, that aims for fairness in FL. \texttt{GIFAIR-FL} resorts to regularization techniques by penalizing the spread in the loss of clients/groups to drive the optimizer to fair solutions. We show that our regularized formulation can be viewed as a dynamic client re-weighting technique that adaptively gives higher weights to low-performing individuals or groups. Our proposed method adapts the client weights at every communication round accordingly. One key feature of \texttt{GIFAIR-FL} is that it can handle both \textbf{group-level and individual-level fairness}. Also, \texttt{GIFAIR-FL} can be naturally tailored to either a global FL algorithm or a personalized FL algorithm. We then prove that, under reasonable conditions, our algorithm converges to an optimal solution for strongly convex objective functions and to a stationary solution for non-convex functions under \textbf{non-$\bm{i.i.d}$. settings}. Through empirical results on image classification and text prediction datasets, we demonstrate that \texttt{GIFAIR-FL} can \textbf{promote fairness while achieving superior or similar prediction accuracy} relative to recent state-of-the-art fair FL algorithms. Besides that, \texttt{GIFAIR-FL} can be easily plugged into other FL algorithms for different purposes. 

\paragraph{Organization:} The rest of the paper is organized as follows. In Sec. \ref{sec:back}, we introduce important notations/definitions and briefly review FL.  Related work is highlighted in Sec. \ref{sec:literature}. In Sec. \ref{sec:fairness}, we present \texttt{GIFAIR-FL-Global} which is a global modeling approach for fairness in FL. We then briefly discuss the limitation of \texttt{GIFAIR-FL-Global} and introduce \texttt{GIFAIR-FL-Per} which is a personalized alternative for fairness, in Sec. \ref{sec:fairness-Per}. Meanwhile, we provide convergence guarantees for both methods. Experiments on image classification and text prediction tasks are then presented in Sec. \ref{sec:exp}. Finally, Sec. \ref{sec:con} concludes the paper with a brief discussion.

\section{Background}
\label{sec:back}

We start by introducing needed background and notation for model development. Then we provide a brief overview of current literature. 


\textbf{Notation:} Suppose there are $K\geq 2$ local devices and each device has $N_k$ datapoints. Denote by $D_k=\big((x_{k,1},y_{k,1}),(x_{k,2},y_{k,2}),\ldots,(x_{k,N_k},y_{k,N_k})\big)$ the data stored at device $k$ where $x\in\mathcal{X}$ is the input, $\mathcal{X}$ is the input space, $y\in\mathcal{Y}$ is the output/label and $\mathcal{Y}$ is the output space. Denote by $\Delta_\mathcal{Y}$ the simplex over $\mathcal{Y}$, $h:\mathcal{X}\mapsto\Delta_\mathcal{Y}$ the hypothesis and $\mathcal{H}$ a family of hypotheses $h$. Let $\ell$ be a loss function defined over $\Delta_\mathcal{Y}\times\mathcal{Y}$. Without loss of generality, assume $\ell\geq 0$. The loss of $h$ is therefore given by $\ell(h(x),y)$. Let $\btheta\in\bm{\Theta}$ be a vector of parameters defining a hypothesis $h$ and $\bm{\Theta}$ is a parameter space. For instance, $\btheta$ can be the model parameters of a deep neural network. In the following section we use $h_{\btheta}$ to represent the hypothesis.

\textbf{Brief background on FL with \texttt{FedAvg}:} In FL, clients collaborate to learn a model that yields better performance relative to each client learning in isolation. This model is called the global model where the global objective function is to minimize the average loss over all clients:
\begin{align*}
    \min_{\btheta}F(\btheta)\coloneqq \sum_{k=1}^Kp_kF_k(\btheta),
\end{align*}
where $p_k=\frac{N_k}{\sum_k N_k}$, $F_k(\btheta)=\mathbb{E}_{(x_{k,i},y_{k,i})\sim\mathcal{D}_k}[\ell(h_{\btheta}(x_{k,i}),y_{k,i})]\approx\frac{1}{N_k}\sum_{j=1}^{N_k}[\ell(h_{\btheta}(x_{k,j}),y_{k,j})]$ and $\mathcal{D}_k$ indicates the data distribution of the $k$-th device's data observations $(x_{k,i},y_{k,i})$. During training, all devices collaboratively learn global model parameters $\btheta$ to minimize $F(\btheta)$. The most commonly used method to learn the global objective is \texttt{FedAvg} \citep{mcmahan2017communication}. Details of \texttt{FedAvg} are highlighted in Algorithm \ref{algo:fed_avg_1} as our work will build upon it for fairness. As shown in Algorithm \ref{algo:fed_avg_1}, \texttt{FedAvg} aims to learn a global parameter $\bm{\theta}$, by iteratively averaging local updates $\bm{\theta}_k$ learned by performing $E$ steps of stochastic gradient descent (SGD) on each client's local objective $F_k$. 
 
\begin{algorithm}[!htbp]
	\SetAlgoLined
	\KwData{number of communication rounds $C$, number of local updates $E$, SGD learning rate schedule $\{\eta^{(t)}\}_{t}$, initial model parameter $\bm{\theta}$}
	\For{$c=0:(C-1)$}{
	    Select some clients by sampling probability $p_k$ and denote by $\mathcal{S}_c$ the set of selected clients\;
	    Server broadcasts $\bm{\theta}$\;
	    \For{all selected devices}{
	        $\bm{\theta}_k^{(cE)}=\bm{\theta}$\;
    		\For{$t=cE:((c+1)E-1)$}{
    		    Randomly sample a subset of data and denote it as $\zeta^{(t)}_k$\;
    		    Local Training $\bm{\theta}^{(t+1)}_k=\bm{\theta}^{(t)}_k-\eta^{(t)} g_k(\bm{\theta}^{(t)}_k;\zeta^{(t)}_k)$ \;
    		}
		}
		Aggregation $\bm{\bar{\theta}}_c=\frac{1}{|\mathcal{S}_c|}\sum_{k\in\mathcal{S}_c}\bm{\theta}^{((c+1)E)}_k$, Set $\bm{\theta}=\bm{\bar{\theta}}_c$\;
	}
	Return $\bm{\bar{\theta}}_C$.
	\caption{\texttt{FedAvg} Algorithm \citep{mcmahan2017communication}}
	\label{algo:fed_avg_1}
\end{algorithm}





In Algorithm \ref{algo:fed_avg_1}, $\zeta_k$ denotes the set of indices corresponding to a subset of training data on device $k$ and $g_k(\cdot;\zeta_k)$ denotes the stochastic gradient of $F_k(\cdot)$ evaluated on the subset of data indexed by $\zeta_k$. Also, $|\mathcal{S}_c|$ denotes the cardinality of $\mathcal{S}_c$. One should note that it is also common for the central server to sample clients uniformly and then take a weighted average using $p_k$ \citep{li2019convergence}. Whichever method used, the resulting model may not be fair as small a $p_k$ implies a lower weight for client $k$.

\textbf{Defining fairness in FL:} Suppose there are $d\in [2,K]$ groups and each client can be assigned to one of those groups $s\in[d]\coloneqq\{1,\ldots,d\}$. \textbf{Note that clients from different groups are typically non-$\bm{i.i.d.}$} Denote by $k^i, k\in[K], i\in[d]$ the index of $k$-th local device in group $i$. Throughout this paper, we drop the superscript $i$ unless we want to emphasize $i$ explicitly. Group fairness can be defined as follows.
\begin{definition}
\label{definition:1}
Denote by $\{a^i_{1}\}_{1\leq i\leq d}$ and $\{a^i_{2}\}_{1\leq i\leq d}$ the sets of performance measures (e.g., testing accuracy) of trained models 1 and 2 respectively. We say model 1 is more fair than model 2 if $Var(\{a^i_{1}\}_{1\leq i\leq d})<Var(\{a^i_{2}\}_{1\leq i\leq d})$, where $Var$ is variance.
\end{definition}

Definition \ref{definition:1} is straightforward: a model is fair if it yields small discrepancies among testing accuracies of different groups. It can be seen that when $d=K$, Definition \ref{definition:1} is equivalent to individual fairness \citep{li2019fair}. Definition 1 is widely adopted in FL literature \citep{mohri2019agnostic, li2019fair, li2020tilted, li2021ditto}. This notion of fairness might be different from traditional definitions such as demographic disparity \citep{feldman2015certifying}, equal opportunity and equalized odds \citep{hardt2016equality} in centralized systems. The reason is that those definitions cannot be extended to FL as there is no clear notion of an outcome which is ``good" for a device \citep{kairouz2019advances}. Instead, fairness in FL can be reframed as equal access to effective models (e.g., the accuracy parity \citep{zafar2017fairness} or the representation disparity \citep{li2019fair}). Specifically, the goal is to train models that incur a uniformly good performance across all devices \citep{kairouz2019advances}.



\subsection{Literature Overview}
\label{sec:literature}

Now we briefly review existing state-of-the-art fair and personalized FL algorithms.


\textbf{Fair FL:} \citet{mohri2019agnostic} propose a minimax optimization framework named agnostic FL (\texttt{AFL}). \texttt{AFL} optimizes the worst weighted combination of local devices and is demonstrated to be robust to unseen testing data. \citet{du2020fairness} further refine the notation of \texttt{AFL} and propose the \texttt{AgnosticFair} algorithm. Specifically, they linearly parametrize weight parameters by kernel functions and show that \texttt{AFL} can be viewed as a special case of \texttt{AgnosticFair}. Upon that, \citet{hu2020fedmgda+} combine minimax optimization with gradient normalization techniques to produce a fair algorithm \texttt{FedMGDA+}. Motivated by fair resource allocation problems, \citet{li2019fair} propose $q$-Fair FL (\texttt{q-FFL}). \texttt{q-FFL} reweights loss functions such that devices with poor performance will be given relatively higher weights. The \texttt{q-FFL} objective is proved to encourage individual fairness in FL. However, this algorithm requires accurate estimation of a local Lipschitz constant $L$. Later, \cite{li2020tilted} developed a tilted empirical risk minimization (\texttt{TERM}) algorithm to handle outliers and class imbalance in statistical estimation procedures. \texttt{TERM} has been shown to be superior to \texttt{q-FFL} in many FL applications. Along this line, \cite{huang2020fairness} propose to use training accuracy and frequency to adjust weights of devices to promote fairness. \citet{zhang2020fairfl} develop an algorithm to minimize the discrimination index of the global model to encourage fairness. 
Here we note that recent work study collaborative fairness in FL \citep{zhang2020hierarchically, xu2020towards,lyu2020collaborative}. The goal of this literature, which is perpendicular to our purpose, is to provide more rewards to high-contributing participants while penalizing free riders. 

\textbf{Personalized FL:} One alternative to global modeling is personalized FL which allows each client to retain their own individualized parameters $\{\btheta_k\}_{k=1}^K$. For instance, in Algorithm \ref{algo:fed_avg_1} and after training is done, each device $i$ can use $\bm{\theta} = \bm{\bar{\theta}}_C$ as the initial weight and run additional SGD steps to obtain a personalized solution $\btheta_i$. Though personalization techniques do not directly target fairness, recent papers have shown that personalized FL algorithms may improve fairness. \citet{arivazhagan2019federated} and \citet{liang2020think} use different layers of a network to represent global and personalized solutions. Specifically, they fit personalized layers to each local device such that each device will return a task-dependent solution based on its own local data. \citet{wang2019federated, yu2020salvaging, dinh2020personalized} and \citet{li2021ditto} resort to fine-tuning techniques to learn personalized models. Notably, \citet{li2021ditto} develops a multi-task personalized FL algorithm \texttt{Ditto}. After optimizing a global objective function, \texttt{Ditto} allows local devices to run more steps of SGD, subject to some constraints, to minimize their own losses. \citet{li2021ditto} have shown that \texttt{Ditto} can significantly improve testing accuracy among local devices and encourage fairness.  

\textbf{Features of \texttt{GIFAIR-FL}:} We here give a quick comparison to highlight the features of our proposed algorithm. The detailed formulation of \texttt{GIFAIR-FL} will be presented in the following section. \texttt{GIFAIR-FL} resorts to regularization to penalize the spread  in the loss of client groups. Interestingly, \texttt{GIFAIR-FL} can be seen as a dynamic re-weighting strategy based on the statistical ordering of client/group losses at each communication round. As such, our approach aligns with FL literature that uses re-weighting client schemes, yet existing work faces some limitations. Specifically, \texttt{AFL} and its variants \citep{mohri2019agnostic,li2019fair, hu2020fedmgda+, du2020fairness} exploit minimax formulations that optimize the worst-case distribution of weights among clients to promote fairness. Such approaches lead to overly pessimistic solutions as they only focus on the device with the largest loss. As will be shown in our case studies, \texttt{GIFAIR-FL} significantly outperforms such approaches. Adding to this key advantage, \texttt{GIFAIR-FL} enjoys convergence guarantees even for non-$i.i.d.$ data and is amenable to both global and personalized modeling.



\section{\texttt{GIFAIR-FL-Global}: a Global Model for Fairness}
\label{sec:fairness}

We start by detailing our proposed global modeling approach - \texttt{GIFAIR-FL-Global}. In this approach, all local devices collaborate to learn one global model parameter $\bm{\theta}$. Our fair FL formulation aims at imposing group fairness while minimizing the training error. More specifically, our goal is to minimize the discrepancies in the average group losses while achieving a low training error. By penalizing the spread in the loss among client groups, we propose a regularization framework for computing optimal parameters $\bm{\theta}$ that balances learning accuracy and fairness. This translates to solving the following optimization problem

\begin{align}
\label{eq:server_obj}
    \min_{\bm{\theta}} \; H(\bm{\theta})\triangleq \sum_{k=1}^K  p_kF_k(\bm{\theta})  + \lambda\sum_{1\leq i < j\leq d}| L_{i}(\bm{\theta})- L_{j}(\bm{\theta})|,
\end{align}
where $\lambda$ is a positive scalar that balances fairness and goodness-of-fit, and 
\begin{equation*}\label{eqn:group_loss}
L_i(\bm{\theta})  \triangleq \dfrac{1}{|{\cal A}_i|} \sum_{k \in {\cal A}_i} F_k(\bm{\theta})
\end{equation*}
is the average loss for client group $i$, ${\cal A}_i$ is the set of indices of devices that belong to group $i$, and $|\mathcal{A}|$ is the cardinality of the set $\mathcal{A}$. 

\begin{remark}
Objective \eqref{eq:server_obj} aims at ensuring fairness by reducing client loss spread when losses are evaluated at a single global parameter $\btheta$. This achieves fairness from the server perspective. Specifically, the goal is to find a single solution that yields small discrepancies among $\{L_i(\btheta)\}_{i=1}^d$.
\end{remark}

In typical FL settings, the global objective is given as $$H(\bm{\theta}) = \sum_{k=1}^K \,p_k H_k(\bm{\theta})$$ where each client uses local data to optimize a surrogate of the global objective function. For instance, \texttt{FedAvg} simply uses the local objective function $H_k(\bm{\theta}) = F_{k}(\bm{\theta})$ for a given client $k$. Interestingly, our global objective in (\ref{eq:server_obj}) can also be written as  $H(\bm{\theta}) = \sum_{k=1}^K \,p_k H_k(\bm{\theta})$ as shown in Lemma \ref{lemma:equivalence} below. 

\begin{lemma} \label{lemma:equivalence}
Let $s_k\in[d]$ denote the group index of device $k$. For any given $\bm{\theta}$, the global objective function $H(\bm{\theta})$ defined in~\eqref{eq:server_obj} can be expressed as
\begin{equation}\label{eqn:global_obj_weighted}
    H(\bm{\theta}) = \sum_{k=1}^K p_k\,\left(1 + \dfrac{\lambda}{p_k|{\cal A}_{s_k}|} r_k(\bm{\theta}) \right)F_k(\bm{\theta}),
\end{equation}
where 
\[r_k(\bm{\theta}) = \sum_{1\leq j \neq s_k \leq d } {\normalfont\mbox{sign}}(L_{s_k}(\bm{\theta}) - L_j(\bm{\theta})). \]
Consequently, 
\[H(\bm{\theta}) = \sum_{k=1}^K \, p_k H_k(\bm{\theta}) \]
such that the local client objective is, 
\begin{equation}\label{eqn:local_obj_weighted} 
    H_k(\bm{\theta}) \triangleq  \left(1 + \dfrac{\lambda}{p_k\, |{\cal A}_{s_k}|} r_k(\bm{\theta}) \right) F_k(\bm{\theta}).
\end{equation}
\end{lemma}

In Lemma \ref{lemma:equivalence}, $r_k(\bm{\theta}) \in \{-d+1, -d+3, \ldots, d-3, d-1\}$ is a scalar directly related to the statistical ordering of $L_{s_k}$ among client group losses. To illustrate that in a simple example, suppose that at a given $\bm{\theta}$, we have $L_1(\bm{\theta}) \geq L_2(\bm{\theta}) \geq \ldots \geq L_d(\bm{\theta})$. Then,
\[r_k(\bm{\theta}) = \begin{cases}
\begin{array}{ll}
d-1 \quad & \mbox{if }s_k = 1\\
d-3 \quad & \mbox{if }s_k = 2\\
\quad \vdots &\\
-d+1\quad & \mbox{if }s_k = d.
\end{array}
\end{cases}\]

According to Lemma 1, one can view our global objective as a parameter-based weighted sum of the client loss functions. Particularly, rather than using uniform weighting for clients, our assigned weights are functions of the parameter $\bm{\theta}$. More specifically, for a given parameter $\bm{\theta}$, our objective yields higher weights for groups with higher average group loss; hence, imposing group fairness. To illustrate this idea, we provide a simple concrete example.

\begin{example}
Without loss of generality and for a given $\bm{\theta}$, consider four different groups each having 10 clients with $L_1(\bm{\theta}) > L_2(\bm{\theta}) > L_3(\bm{\theta})> L_4(\bm{\theta})$. Then our global objective function $H(\bm{\theta})$ in ~\eqref{eqn:global_obj_weighted} can be expressed as
\begin{align*}
    &\sum_{k=1}^{40}p_kF_k(\bm{\theta})+\lambda\bigg(|L_{1}(\bm{\theta})- L_{2}(\bm{\theta})|+| L_{1}(\bm{\theta})- L_{3}(\bm{\theta})|\\
    &\qquad +|L_{1}(\bm{\theta})- L_{4}(\bm{\theta})| +|L_{2}(\bm{\theta})- L_{3}(\bm{\theta})| +|L_{2}(\bm{\theta})- L_{4}(\bm{\theta})|+|L_{3}(\bm{\theta})- L_{4}(\bm{\theta})|\bigg),
\end{align*}
which is equivalent to
\begin{align*}&\sum_{k \in {\cal A}_1} p_k \left(1 + \dfrac{3\lambda}{10p_k}\right)F_k(\bm{\theta})  + \sum_{k \in {\cal A}_2} p_k \left(1 + \dfrac{\lambda}{10p_k}\right)F_k(\bm{\theta})\\
&\hspace*{-0.3cm}+ \sum_{k \in {\cal A}_3} p_k \left(1 - \dfrac{\lambda}{10p_k}\right)F_k(\bm{\theta})+\sum_{k \in {\cal A}_4} p_k \left(1 - \dfrac{3\lambda}{10p_k}\right)F_k(\bm{\theta}).\end{align*}
The objective clearly demonstrates a higher weight applied to clients that belong to a group with a higher average loss.
\end{example}

According to~\eqref{eqn:local_obj_weighted}, the optimization problem solved by every selected client is a weighted version of the local objective in \texttt{FedAvg}. The objective imposes a higher weight for clients that belong to groups with higher average losses. These weights will be dynamically updated at every communication round. To assure positive weights for clients, we require the following bounds on $\lambda$
\[0 \leq \lambda < \lambda_{max} \triangleq \min_{k} \left\{\dfrac{p_k |{\cal A}_{s_k}|}{d-1}\right\}.   \]
When $\lambda = 0$, our approach is exactly \texttt{FedAvg}. Moreover, a  higher value of $\lambda$ imposes more emphasis on fairness.

Now, the above formulation can be readily extended to individual fairness; simply through considering each client to be a group. This translates to the global objective in (\ref{eqn:global_obj_weighted}) to be given as
\[H(\bm{\theta}) = \sum_{k=1}^K \left( 1 + \dfrac{\lambda}{p_k}r_k(\bm{\theta}) \right) F_k(\bm{\theta}).\]

In essence our approach falls in line with FL literature that exploit re-weighting of clients. For instance, \texttt{AFL} proposed by \cite{mohri2019agnostic} computes at every communication round the worst-case distribution of weights among clients. This approach promotes robustness but may be overly conservative in the sense that it focuses on the largest loss and thus causes very pessimistic performance to other clients. Our algorithm, however,  adaptively updates the weight of clients at every communication round based on the statistical ordering of client/group losses. Moreover, the dynamic update of the weights can potentially avoid over-fitting by impeding updates for clients with low loss. We will further demonstrate the advantages of our algorithm in Sec. \ref{sec:exp}. In the next subsection, we provide our detailed algorithm for solving our proposed objective.

\subsection{Algorithm}

\begin{algorithm}[!htbp]
	\SetAlgoLined
	\KwData{number of devices $K$, fraction $\alpha$,  number of communication rounds $C$, number of local updates $E$, SGD learning rate schedule $\{\eta^{(t)}\}_{t}$, initial model parameter $\bm{\theta}$, regularization parameter $\lambda$, initial loss $\{L_i\}_{1\leq i\leq d}$}
	\For{$c=0:(C-1)$}{
	    Select clients by sampling probability $p_k$ and denote by ${\cal S}_c$ the indices of these clients\;
	    Server broadcasts ${\color{red}\left(\bm{\theta},\left\{\dfrac{\lambda}{p_k|{\cal A}_{s_k}|}r^c_k(\bm{\theta})\right\}_{k\in {\cal S}_c}\right)}$\;
	    \For{$k\in\mathcal{S}_c$}{
	        $\bm{\theta}_k^{(cE)}=\bm{\theta}$\;
    		\For{$t=cE:((c+1)E-1)$}{
    		    Randomly sample a subset of data and denote it as $\zeta^{(t)}_k$\;
    		    $\bm{\theta}^{(t+1)}_k=\bm{\theta}^{(t)}_k-\eta^{(t)}{\color{red}\left(1+\dfrac{\lambda}{p_k|{\cal A}_{s_k}|}r^c_k(\bm{\theta})\right)} g_k(\bm{\theta}^{(t)}_k;\zeta^{(t)}_k)$ \;
    		    \tcp{Note that $r^c_k(\bm{\theta})$ is fixed during local update (See Remark \ref{remark:rk})}
    		}
		}
		Aggregation $\bm{\bar{\theta}}_c=\frac{1}{|\mathcal{S}_c|}\sum_{k\in\mathcal{S}_c}\bm{\theta}^{((c+1)E)}_k$, Set $\bm{\theta}=\bm{\bar{\theta}}_c$\;
		{\color{red}Calculate $L_i=\frac{1}{|{\cal A}_i|}\sum_{k \in {\cal A}_i}F_k(\bm{\theta}^{((c+1)E)}_k)$ for all $i\in[d]$ and update $r_k^{c+1}(\bm{\theta})$}\; 
		$c \leftarrow c+1$\;
	}
	Return $\bm{\bar{\theta}}_C$.
	\caption{\texttt{GIFAIR-FL-Global} Algorithm}
	\label{algo:1}
\end{algorithm}

In this section, we describe our proposed algorithm \texttt{GIFAIR-FL-Global} which is detailed in Algorithm~\ref{algo:1}. We highlight the difference between \texttt{GIFAIR-FL-Global} and \texttt{FedAvg} in red color. At every communication round $c$, our algorithm selects a set of clients to participate in the training and shares $r_k^c$ with each selected client. For each client, multiple SGD steps are then applied to a weighted client loss function. The updated parameters are then passed to the server that aggregates these results and computes $r_k^{c+1}$.

Computationally, our approach requires evaluating the client loss function at every communication round to compute $r_{k}^c$. Compared to existing fair FL approaches, \texttt{GIFAIR-FL-Global} is simple and computationally efficient. For instance, \texttt{q-FFL} proposed by \cite{li2019fair} first runs \texttt{FedAvg} to obtain a well-tuned learning rate and uses this learning rate to roughly estimate the Lipschitz constant $L$. Another example is \texttt{AFL} which requires running two gradient calls at each iteration to estimate the gradients of model and weight parameters. Similarly, \texttt{Ditto} requires running additional steps of SGD, at each communication round, to generate personalized solutions. In contrast, our proposed method can be seen as a fairness-aware weighted version of \texttt{FedAvg}.



\begin{remark}
In Algorithm \ref{algo:1}, we sample local devices by sampling probability $p_k$ and aggregate model parameters by an unweighted average $\frac{1}{|\mathcal{S}_c|}\sum_{k\in\mathcal{S}_c}\bm{\theta}^{((c+1)E)}_k$.  Alternatively, one may choose to uniformly sample clients. Then, the  aggregation strategy should be replaced by $\bm{\bar{\theta}}_c=\frac{K}{|\mathcal{S}_c|}\sum_{k\in\mathcal{S}_c}p_k\bm{\theta}^{((c+1)E)}_k$ \citep{li2019convergence}.
\end{remark}


\begin{remark}
Instead of broadcasting $p_k$ and $|{\cal A}_{s_k}|$ separately to local devices, the central server broadcasts the product $\frac{\lambda}{p_k|{\cal A}_{s_k}|}r_k^c(\bm{\theta})$ to client $k$. Hence, the local device $k$ cannot obtain any information about $p_k$, $|{\cal A}_{s_k}|$ and $r_k^c(\bm{\theta})$. This strategy can protect privacy of other devices.  
\end{remark}

\begin{remark}
\label{remark:rk}
Notice that in \eqref{eqn:local_obj_weighted}, $H_k(\bm{\theta})$ is not differentiable due to the $r_k(\bm{\theta})$ component. However, $r^c_k$ is fixed during local client training as it is calculated on the central server. Also, local devices do not have any information about other devices hence they cannot update $r^c_k$ during local training.
\end{remark}

\subsection{Convergence Guarantees}

In this section, we first show that, under mild conditions, \texttt{GIFAIR-FL-Global} converges to the global optimal solution at a rate of $\mathcal{O}(\frac{E^2}{T})$ for strongly convex functions and to a stationary point at a rate of $\mathcal{O}(\frac{(E-1)\log(T+1)}{\sqrt{T}})$, up to a logarithmic factor, for non-convex functions. Here $T\coloneqq CE$ denotes the total number of iterations across all devices. \textbf{Our theorems hold for both $\bm{i.i.d.}$ and non-$\bm{i.i.d.}$ data.} Due to space limitation, we defer proof details to the Appendix. 

\subsubsection{Strongly Convex Functions}

We assume each device performs $E$ steps of local updates 
and make the following assumptions. Here, our assumptions are based on $F_k$ rather than $H_k$. These assumptions are very common in many FL papers \citep{li2019communication, li2018federated, li2019convergence}.
\begin{assumption}
\label{assumption:1}
$F_k$ is $L$-smooth and $\mu$-strongly convex for all $ k\in[K]$.
\end{assumption}



\begin{assumption}
\label{assumption:2}
 The variance of stochastic gradient is bounded. Specifically,
\begin{align*}
    \mathbb{E}\bigg\{\norm{g_k(\bm{\theta}^{(t)}_k;\zeta^{(t)}_k)-\nabla F_k(\bm{\theta}^{(t)}_k)}^2\bigg\}\leq\sigma_k^2,\;\; \forall k\in[K].
\end{align*}
\end{assumption}
\begin{assumption}
\label{assumption:3}
The expected squared norm of the stochastic gradient is bounded. Specifically,
\begin{align*}
    \mathbb{E}\bigg\{\norm{g_k(\bm{\theta}^{(t)}_k;\zeta^{(t)}_k)}^2\bigg\}\leq G^2, \forall k\in[K].
\end{align*}
\end{assumption}

Typically, data from different groups are \textbf{non-$\bm{i.i.d.}$}. We modify the definition in \citet{li2019convergence} to roughly quantify the degree of non-$i.i.d.$-ness. Specifically, 
\begin{align*}
    \Gamma_K = H^* - \sum_{k=1}^Kp_kH^*_k= \sum_{k=1}^K p_k(H^*-H^*_k),
\end{align*}
where $H^*\triangleq H(\bm{\theta}^*)= \sum_{k=1}^KH_k(\bm{\theta}^*)$ is the optimal value of the global objective function and $H_k^* \triangleq H_k(\bm{\theta}_k^*)$ is the optimal value of the local loss function. If data are $i.i.d.$, then $\Gamma_K\to 0$ as the number of samples grows. Otherwise, $\Gamma_K \neq 0$ \citep{li2019convergence}. Given all aforementioned assumptions, we next prove the convergence of our proposed algorithm. We first assume all devices participate in each communication round (i.e., $|\mathcal{S}_c|=K, \forall c$).


\begin{theorem}
\label{theorem:conv_full}
Assume Assumptions \ref{assumption:1}-\ref{assumption:3} hold and $|\mathcal{S}_c|=K$. If $\eta^{(t)}$ is decreasing in a rate of $\mathcal{O}(\frac{1}{t})$ and $\eta^{(t)}\leq\mathcal{O}(\frac{1}{L})$, then for $\gamma,\mu,\epsilon>0$, we have
\begin{align*}
    &\mathbb{E}\bigg\{H(\bm{\bar{\theta}}_C)\bigg\} - H^*\leq \frac{L}{2}\frac{1}{\gamma+T}\bigg\{ \frac{4\xi}{\epsilon^2\mu^2}+(\gamma+1)\norm{\bm{\bar{\theta}}^{(0)}-\bm{\theta}^*}^2\bigg\},
\end{align*}
where $\xi=8(E-1)^2G^2+4L\Gamma+2\frac{\Gamma_{max}}{\eta^{(t)}}+4\sum_{k=1}^Kp_k^2\sigma_k^2$ and $\Gamma_{max}\coloneqq\sum_{k=1}^Kp_k|(H^*-H^*_k)|\geq|\sum_{k=1}^K p_k(H^*-H^*_k)|= |\Gamma_K|$. Here $\bm{\bar{\theta}}^{(0)}\coloneqq\btheta^{(0)}$ where $\btheta^{(0)}$ is the initial model parameter in the central server.
\end{theorem}

\begin{remark}
Theorem \ref{theorem:conv_full} shows an $\mathcal{O}(\frac{E^2}{T})$ convergence rate which is similar to that obtained from  \texttt{FedAvg}. However, the rate is also affected by $\xi$, which contains the degree of non-$i.i.d.$-ness. Under a fully $i.i.d.$ settings where $\Gamma_K=\Gamma_{max}=0$, we retain the typical \texttt{FedAvg} for strongly convex functions.
\end{remark}



Next, we assume only a fraction of devices participate in each communication round (i.e., $|\mathcal{S}_c|=\alpha K, \forall c, \alpha\in(0,1)$). 
As per Algorithm \ref{algo:1}, all local devices are sampled according to the sampling probability $p_k$ \citep{li2018federated}. Our Theorem can similarly be extended to the scenario where devices are sampled uniformly (i.e., with the same probability). Recall, the aggregation strategy becomes $\bm{\bar{\theta}}_c=\frac{K}{|\mathcal{S}_c|}\sum_{k\in\mathcal{S}_c}p_k\bm{\theta}_k$ \citep{li2019convergence}.

\begin{theorem}
\label{theorem:conv_partial}
Assume at each communication round, the central server samples a fraction $|\mathcal{S}_c|$ of devices according to the sampling probability $p_k$. Additionally, assume Assumptions \ref{assumption:1}-\ref{assumption:3} hold. If $\eta^{(t)}$ is decreasing at a rate of $\mathcal{O}(\frac{1}{t})$ and $\eta^{(t)}\leq\mathcal{O}(\frac{1}{L})$, then for $\gamma,\mu,\epsilon>0$, we have
\begin{align*}
    &\mathbb{E}\bigg\{H(\bm{\bar{\theta}}_C)\bigg\} - H^*\leq\frac{L}{2}\frac{1}{\gamma+T}\bigg\{ \frac{4(\xi+\tau')}{\epsilon^2\mu^2}+(\gamma+1)\norm{\bm{\bar{\theta}}^{(0)}-\bm{\theta}^*}^2\bigg\},
\end{align*}
where $\tau'=\frac{4G^2E^2}{|\mathcal{S}_c|}$.
\end{theorem}

\begin{remark}
Under the partial device participation scenario, the same convergence rate $\mathcal{O}(\frac{E^2}{T})$ holds. The only difference is that there is a term $\tau'=\frac{4G^2E^2}{|\mathcal{S}_c|}$ that appears in the upper bound. This ratio slightly impedes the convergence rate when the number of sampled devices $|\mathcal{S}_c|$ is small.
\end{remark}

\subsubsection{Non-convex Functions}

To prove the convergence result on non-convex functions, we replace Assumption \ref{assumption:1} by the following assumption.
\begin{assumption}
\label{assumption:4}
$F_k$ is $L$-smooth for all $ k\in[K]$.
\end{assumption}

\begin{theorem}
\label{theorem:nonconvex_full}
Assume Assumptions \ref{assumption:2}-\ref{assumption:4} hold and $|\mathcal{S}_c|=K$. If $\eta^{(t)}=\mathcal{O}(\frac{1}{\sqrt{t}})$ and $\eta^{(t)}\leq\mathcal{O}(\frac{1}{L})$, then our algorithm converges to a stationary point. Specifically,
\begin{align*}
    &\min_{t=1,\ldots,T} \mathbb{E}\bigg\{\norm{\nabla H(\bm{\bar{\theta}}^{(t)})}^2 \bigg\}\\ &\qquad \leq \frac{\bigg\{ 2\big(1+2L^2\log(T+1)\big)\mathbb{E}\big\{H(\bm{\bar{\theta}}^{(0)}) -H^*\big\} + 2\xi_{\Gamma_K}\bigg\}}{\sqrt{T} },
\end{align*}
where $$\xi_{\Gamma_K}=\mathcal{O}\bigg(\big(2L^2\Gamma_K+8(E-1)LG^2 + 10L\sum_{k=1}^Kp_k\sigma_k^2\big)\log(T+1)\bigg),$$
and $\bm{\bar{\theta}}^{(t)}=\frac{1}{|\mathcal{S}_c|}\sum_{k\in\mathcal{S}_c}\bm{\theta}^{(t)}_k$.
\end{theorem}

\begin{remark}
Our results show that \texttt{GIFAIR-FL} converges to a stationary point at a rate of $\mathcal{O}(\frac{(E-1)\log(T+1)}{\sqrt{T}})$. Similar to the strongly-convex setting, this convergence rate is affected by the degree of non-$i.i.d.$-ness $\Gamma_K$.
\end{remark}

\subsection{Discussion and Limitations}


We here note our theoretical results require exact computation of $r_k$. However, Algorithm \ref{algo:1} uses an estimate of $r_k$ at every communication round using the local client loss prior to aggregation. This procedure might generate an inexact estimate of $r_k$ as one cannot guarantee $F_k(\bar{\btheta}_c)=F_k(\btheta_k^{((c+1)E)})$ at each communication round. Here recall that $r_k$ in Eq. \eqref{eqn:global_obj_weighted} is calculated based on the order of $F_k(\bar{\btheta}_c)$. To guarantee exact an $r_k$, the server can ask clients to share the local losses evaluated at the global parameters. Specifically, after sharing $\bar{\btheta}_c$ to selected local devices, those devices calculate $\{F_k(\bar{\btheta}_c)\}_k$ and send loss values back to the central server to update $r_k^{c+1}$. This, however, requires more communication rounds. One approach to remedy the limitation of \texttt{GIFAIR-FL-Global} is to develop a personalized counterpart to circumvent additional communication rounds. We will detail this idea in the coming section. 




\section{\texttt{GIFAIR-FL-Per}: A Personalized Model for Fairness}
\label{sec:fairness-Per}

In this section, we slightly tailor \texttt{GIFAIR-FL-Global} to a personalized fair algorithm \texttt{GIFAIR-FL-Per}. While still aiming to minimize the spread in the loss among client groups, our proposed objective evaluates the loss at the client-specific (i.e. personalized) solution . Formally speaking, our objective function is

\begin{align}
\label{eq:server_obj_per}
    &\min_{\bm{\theta}} \; H(\bm{\theta},\bm{\theta}_1,\ldots,\bm{\theta}_K)\triangleq \sum_{k=1}^K  p_kF_k(\bm{\theta}) + \lambda\sum_{1\leq i < j\leq d}\left| L_{i}(\{\bm{\theta}_k\}_{k\in\mathcal{A}_i})- L_{j}(\{\bm{\theta}_k\}_{k\in\mathcal{A}_j})\right|,
\end{align}
where
\begin{equation*}\label{eqn:group_loss_2}
L_i(\{\bm{\theta}_k\}_{k \in {\cal A}_i})  \triangleq \dfrac{1}{|{\cal A}_i|} \sum_{k \in {\cal A}_i} F_k(\bm{\theta}_k)
\end{equation*}
is the average loss for client group $i$ and $\sum_{k=1}^Kp_k\btheta_k=\btheta$. 


\begin{remark}
Different from \eqref{eq:server_obj}, objective \eqref{eq:server_obj_per} achieves fairness from the device perspective. By optimizing \eqref{eq:server_obj_per}, we can obtain device-specific solutions $\{\btheta_k\}_{k=1}^K$ that yield small discrepancies among $\{L_i(\{\btheta_k\}_{k\in\mathcal{A}_i})\}_{i=1}^d$. Although \eqref{eq:server_obj} and \eqref{eq:server_obj_per} have different perspectives, their ultimate goals are aligned with Definition \ref{definition:1}.
\end{remark}

Objective \eqref{eq:server_obj_per} has many notable features: (i) first, compared to \eqref{eq:server_obj}, the new objective function \eqref{eq:server_obj_per} evaluates group losses $\{L_i\}_{i=1}^d$ with respect to personalized solutions $\{\btheta_k\}_{k=1}^K$. This formulation circumvents the need to collect losses evaluated at global parameter and therefore requires no extra communication rounds to calculate $r_k$ exactly; (ii) second, the global parameter $\btheta=\sum_{k=1}^Kp_k\btheta_k$ ensures aggregation happens at every communication round. This can safeguard against over-fitting on local devices. Otherwise, each device will simply minimize its own local loss, without communication, and obtain a small loss value; (iii) before discussing the third property, we first need to present the convergence results of \texttt{GIFAIR-FL-Per}.

\begin{theorem}
\label{theorem:conv_full_per}
Assume Assumptions \ref{assumption:1}-\ref{assumption:3} hold and $|\mathcal{S}_c|=K$. If $\eta^{(t)}$ is decreasing in a rate of $\mathcal{O}(\frac{1}{t})$ and $\eta\leq\mathcal{O}(\frac{1}{L})$, then for $\gamma,\mu,\epsilon>0$, we have
\begin{align*}
    &\mathbb{E}\bigg\{H(\bm{\bar{\theta}}_C,\{\btheta_k^{(T)}\}_{k=1}^K)\bigg\} - H^*\leq \mathcal{O}(\frac{E^2}{T})
\end{align*}
where 
$\sum_{k=1}^Kp_k\btheta_k^{(T)}=\bm{\bar{\theta}}_C$ and $H^*\coloneqq H(\bm{\theta}^{^*},\{\btheta_k\}_{k=1}^K)$ such that $\sum_{k=1}^Kp_k\bm{\theta}_k=\bm{\theta}^{^*}$. A same convergence rate holds for the partial device participation scenario.

Under non-convex condition, we have
\begin{align*}
    \min_{t=1,\ldots,T} \mathbb{E}\bigg\{\norm{\nabla H(\bm{\bar{\theta}}^{(t)})}^2 \bigg\}\leq\mathcal{O}(\frac{(E-1)\log(T+1)}{\sqrt{T}}).
\end{align*}

\end{theorem}

The proof here follows a similar scheme to those in \texttt{GIFAIR-FL-Global}. Theorem \ref{theorem:conv_full_per} implies that \texttt{GIFAIR-FL-Per} drives aggregated parameter $\bm{\bar{\theta}}_C$ to the global optimal solution $\btheta^*$ at a rate of $\mathcal{O}(\frac{E^2}{T})$. This aggregated parameter is obtained from taking the weighed average of personalized solutions $\{\btheta_k^{(T)}\}_{k=1}^K$. This leads to a new interpretation of \texttt{GIFAIR-FL-Per}: once the optimizer reaches $\bm{\bar{\theta}}_C$, device $k$ retains personalized solution $\btheta_k^{(T)}$ that stays in the vicinity of
the global model parameter to balance each client’s shared knowledge and unique characteristics. One can link this idea to \texttt{Ditto} \citep{li2021ditto} - the recent state-of-the-art personalized FL algorithm. \texttt{Ditto} allows local devices to run more steps of SGD, subject to some constraints such that local solutions will not move far away from the global solution. \texttt{GIFAIR-FL-Per}, on the other hand, scales the magnitude of gradients based on the statistical ordering of client/group losses. 

\begin{algorithm}[!htbp]
	\SetAlgoLined
	\KwData{number of devices $K$, fraction $\alpha$,  number of communication rounds $C$, number of local updates $E$, SGD learning rate schedule $\{\eta^{(t)}\}_{t=1}^E$, initial model parameter $\bm{\theta}$, regularization parameter $\lambda$, initial loss $\{L_i\}_{1\leq i\leq d}$}
	\For{$c=0:(C-1)$}{
	    Select $|\mathcal{S}_c|$ clients by sampling probability $p_k$ and denote by ${\cal S}_c$ the indices of these clients\;
	    Server broadcasts $\left(\bm{\theta},\left\{\dfrac{\lambda}{p_k|{\cal A}_{s_k}|}r^c_k(\{\bm{\theta}_k\}_{k=1}^K)\right\}_{k\in {\cal S}_c}\right)$\;
	    \For{$k\in\mathcal{S}_c$}{
	        $\bm{\theta}_k^{(cE)}=\bm{\theta}$\;
    		\For{$t=cE:((c+1)E-1)$}{
    		    $\bm{\theta}^{(t+1)}_k=\bm{\theta}^{(t)}_k-\eta^{(t)}\left(1+\dfrac{\lambda}{p_k|{\cal A}_{s_k}|}r^c_k(\{\bm{\theta}_k\}_{k=1}^K)\right)\nabla F_k(\bm{\theta}^{(t)}_k)$ \;
    		}
		}
		Aggregation $\bm{\bar{\theta}}_c=\frac{1}{|\mathcal{S}_c|}\sum_{k\in\mathcal{S}_c}\bm{\theta}^{((c+1)E)}_k$, Set $\bm{\theta}=\bm{\bar{\theta}}_c$\;
		Calculate $L_i=\frac{1}{|{\cal A}_i|}\sum_{k \in {\cal A}_i}F_k(\bm{\theta}^{((c+1)E)}_k)$ for all $i\in[d]$\;
		Set $\btheta_k=\btheta_k^{((c+1)E)}$ for all $k\in\mathcal{S}_c$. Remain $\btheta_k$ unchanged otherwise\;
		update $r_k^{c+1}(\{\btheta_k\}_{k=1}^K)$\; 
		$c \leftarrow c+1$\;
	}
	Return $\{\btheta_k\}_{k=1}^K$.
	\caption{\texttt{GIFAIR-FL-Per} Algorithm}
	\label{algo:1-per}
\end{algorithm}


Finally, we detail \texttt{GIFAIR-FL-Per} in Algorithm~\ref{algo:1-per}. In the algorithm, $r_k$ is defined as 
\begin{align}
\label{eq:rank_k_per}
    &r_k(\{\bm{\theta}_k\}_{k=1}^K)= \sum_{1\leq j \neq s_k \leq d } {\normalfont\mbox{sign}}(L_{s_k}(\{\bm{\theta}_m\}_{m\in\mathcal{A}_{s_k}}) - L_j(\{\bm{\theta}_m\}_{m\in\mathcal{A}_j})).
\end{align}
In other words, $r_k$ is computed based on the ordering of losses evaluated \textbf{on the personalized solutions}.

\section{Experiments}
\label{sec:exp}

In this section, we test \texttt{GIFAIR-FL} on image classification and text prediction tasks.

We benchmark our model with the following algorithms: \texttt{q-FFL} \citep{li2019fair}, \texttt{TERM} \citep{li2020tilted}, \texttt{FedMGDA+} \citep{hu2020fedmgda+}, \texttt{AFL} \citep{mohri2019agnostic}, and \texttt{FedMGDA+} \citep{hu2020fedmgda+}. To the best of our knowledge, those are the well-known current state-of-the-art FL algorithms for fairness. We also benchmark our model with \texttt{Ditto} \citep{li2021ditto} which is a personalized FL approach using multi-task learning.



\subsection{Image Classification}

\begin{figure*}[!htbp]
    \centering
    \centerline{\includegraphics[width=0.6\columnwidth]{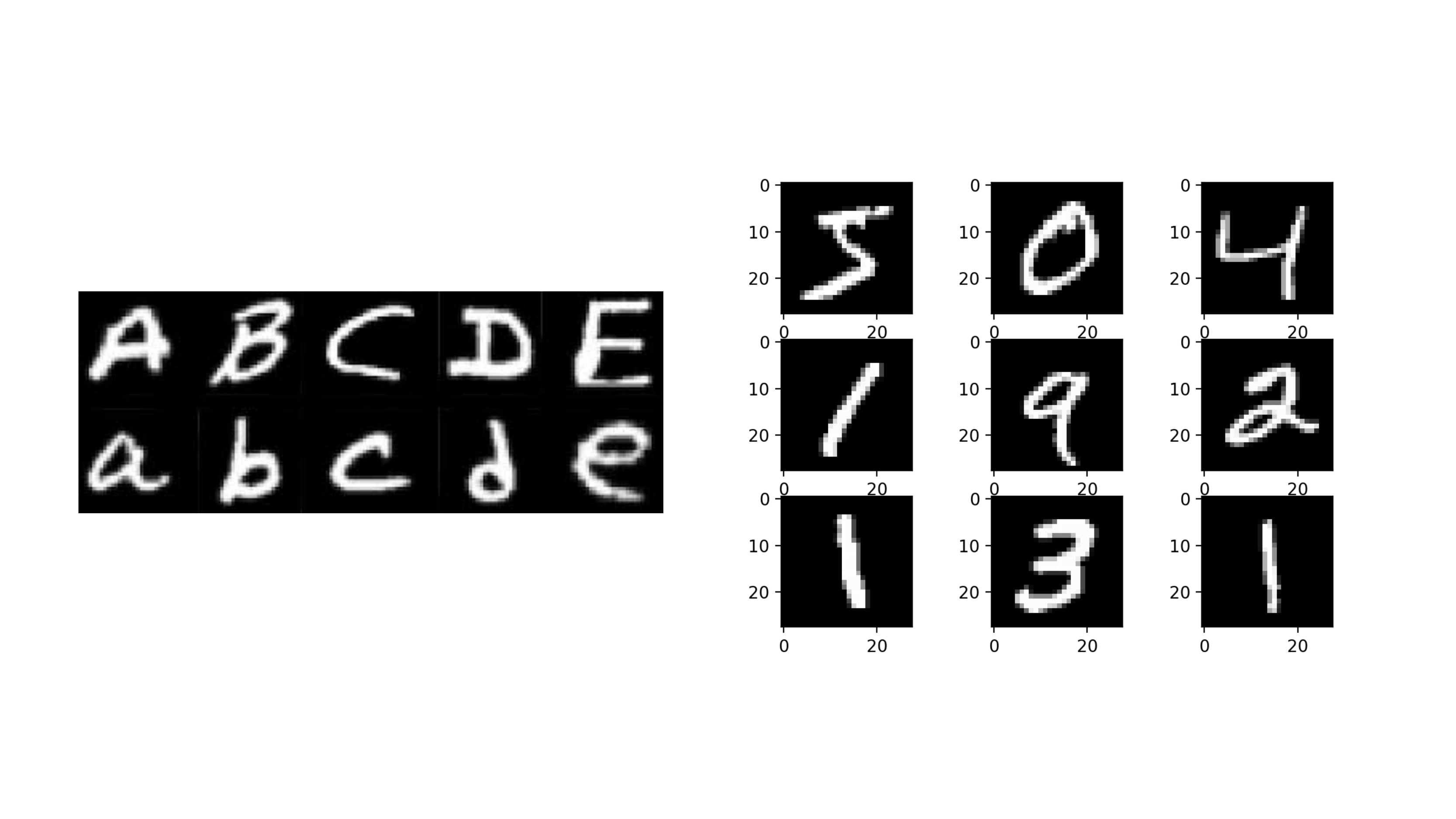}}
    \caption{Example of Images from FEMNIST.}
    \label{fig:femnist}
\end{figure*}

We start by considering a federated image classification dataset FEMNIST (Federated Extended MNIST) \citep{caldas2018leaf}. FEMNIST consists of images of digits (0-9) and English characters (A-Z, a-z) with 62 classes (Figure \ref{fig:femnist}) written by different people. Images are 28 by 28 pixels. All images are partitioned and distributed to 3,550 devices by the dataset creators \citep{caldas2018leaf}. 

\textbf{Individual Fairness} (\textbf{FEMNIST-skewed}, $d=100$) Following the setting in \citep{li2018federated}, we first sample 10 lower case characters (`a'-`j') from Extended MNIST (EMNIST) \citep{cohen2017emnist} and distribute 5 classes of images to each device. Each local device has 500 images. There are 100 devices in total. 
Results are reported in Table \ref{table:femnist-1}. (\textbf{FEMNIST-original}, $d=500$) Following the setting in \citep{li2021ditto}, we sample 500 devices and train models using the default data stored in each device. 
Results are reported in Table \ref{table:femnist-2}. 


\textbf{Group Fairness} (\textbf{FEMNIST-3-groups}, $d=3$) We manually divide FEMNIST data into three groups. See Table \ref{table:assign} for the detailed assignment. \textit{This assignment is inspired by the statistic that most people prefer to write in lowercase letters while a small amount of people use capital letters or a mixed of two types \citep{jones2004case}}. In such cases, it is important to assure that an FL algorithm is capable of achieving similar performance between such groups. Results are reported in Table \ref{table:femnist-3}.   


\begin{table}[!htbp]
\centering
\small
\begin{tabular}{ccccc}
\hline
\textbf{Group} & \textbf{Data Type} & \textbf{Number of Images} & \textbf{Number of Devices}  \\ \hline
Group 1 & Capital Letters + Digits  & 800 & 60 \\ \hline
Group 2 & Lowercase Letters + Digits & 1,000 & 100 \\ \hline
Group 3 & Capital/Lowercase Letters + Digits & 600 & 40 \\ \hline
\end{tabular}
\caption{Data Structure of FEMNIST-3-groups}
\label{table:assign}
\vspace{-0.6cm}
\end{table}
\textbf{Implementation:} For all tasks, we randomly split the data on each local device into a $70\%$ training set, a $10\%$ validation set and a $20\%$ testing set. This is a common data splitting strategy used in many FL papers \citep{li2018federated, chen2018federated, reddi2020adaptive}. The batch size is set to be 32. We use the tuned initial learning rate $0.1$ and decay rate $0.99$ for each method. During each communication round, $10$ devices are randomly selected and each device will run 2 epochs of SGD. We use a CNN model with 2 convolution layers followed by 2 fully connected layers. All benchmark models are well-tuned. Specifically, we solve \texttt{q-FFL} with $q\in\{0,0.001,0.01,0.1,1,2,5,10\}$ \citep{li2019fair} in parallel and select the best $q$. Here, the best $q$ is defined as the $q$ value where the variance decreases the most while the averaged testing accuracy is superior or similar to \texttt{FedAvg}. This definition is borrowed from the original \texttt{q-FFL} paper \citep{li2019fair}. Similarly, we train \texttt{TERM} with $t\in\{1,2,5\}$ and select the best $t$ \citep{li2020tilted}. For \texttt{Ditto}, we tune the regularization parameter $\lambda_{Ditto}\in\{0.01, 0.05, 0.1, 0.5, 1, 2, 5\}$. In \texttt{GIFAIR-FL}, we tune the parameter $\lambda\in\{0, 0.1\lambda_{max}, 0.2\lambda_{max}, \ldots, 0.8\lambda_{max}, 0.9\lambda_{max}\}$. Here kindly note that $\lambda_{max}$ is a function of $p_k, |{\cal A}_{s_k}|$ and $d$ (i.e., data-dependent). 

\textbf{Performance metrics:} Denote by $a_k$ the prediction accuracy on device $k$. We define (1) individual-level mean accuracy as $\bar{a}\coloneqq\frac{1}{K}\sum_{k=1}^K a_k$ and (2) individual-level variance as $Var(a)\coloneqq\frac{1}{K}\sum_{k=1}^K(a_k-\bar{a})^2$. 

\begin{table}[!htbp]
\centering
\footnotesize
\begin{tabular}{ccccccccc}
\hline
\textbf{Algorithm} & \texttt{FedAvg} & \texttt{q-FFL} & \texttt{TERM} & \texttt{FedMGDA+} & \texttt{Ditto} & \texttt{GIFAIR-FL-Global} & \texttt{GIFAIR-FL-Per} \\ \hline
$\bar{a}$ & 79.2 (1.0)  & 84.6 (1.9) & 84.2 (1.3) & 85.0 (1.7) & 92.5 (3.1) & 87.9 (0.9) & \textbf{93.0} (1.1) \\ \hline
$\sqrt{Var(a)}$ & 22.3 (1.1) & 18.5 (1.2) & 13.8 (1.0) & 14.9 (1.6) & 14.3 (1.0) & \textbf{5.7} (0.8) & 6.2 (0.9) \\ \hline
\end{tabular}
\caption{Empirical results on FEMNIST-skewed. Each experiment is repeated 5 times.}
\label{table:femnist-1}
\end{table}

\begin{table}[!htbp]
\centering
\footnotesize
\begin{tabular}{ccccccccc}
\hline
\textbf{Algorithm} & \texttt{FedAvg} & \texttt{q-FFL} & \texttt{TERM} & \texttt{AFL} & \texttt{Ditto} & \texttt{GIFAIR-FL-Global} & \texttt{GIFAIR-FL-Per}  \\ \hline
$\bar{a}$ & 80.4 (1.3) & 80.9 (1.1) & 81.0 (1.0) & 82.4 (1.0) & 83.7 (1.9) & 83.2 (0.7) & \textbf{84.1 (1.2)} \\ \hline
$\sqrt{Var(a)}$ & 11.1 (1.4) & 10.6 (1.3) & 10.3 (1.2) & 9.85 (0.9) & 10.1 (1.6) & 5.2 (0.8) & \textbf{4.5 (0.8)} \\ \hline
\end{tabular}
\caption{Test accuracy on FEMNIST-original. Each experiment is repeated 5 times.}
\label{table:femnist-2}
\end{table}

\begin{table}[!htbp]
\centering
\small
\begin{tabular}{cccccccc}
\hline
\textbf{Algorithm} & \texttt{FedAvg} & \texttt{q-FFL} & \texttt{TERM} & \texttt{FedMGDA+} & \texttt{Ditto}  \\ \hline
Group 1 & 79.72 (2.08) & 81.15 (1.97) & 81.29 (1.45) & 81.03 (2.28) & 82.37 (2.06) \\ \hline
Group 2 & 90.93 (2.35) & 88.24 (2.13) & 88.08 (1.09) & 89.12 (1.74) & \textbf{92.05 (2.00)} \\ \hline
Group 3 & 80.21 (2.91) & 80.93 (1.86) & 81.84 (1.44) & 81.33 (1.59) & 83.03 (2.18)\\ \hline
Discrepancy & 11.21 & 7.31 & 6.79 & 8.09 & 9.02 \\ \hline
\end{tabular}
\begin{tabular}{cccccccc}
\hline
\textbf{Algorithm} & \texttt{GIFAIR-FL-Global} & \texttt{GIFAIR-FL-Per}  \\ \hline
Group 1 & 83.41 (1.34) & \textbf{83.96 (1.22)}\\ \hline
Group 2 & 88.29 (1.22) & 91.05 (1.31)\\ \hline
Group 3 & 84.37 (1.85) & \textbf{84.98 (0.99)}\\ \hline
Discrepancy & \textbf{6.07} & 7.09 \\ \hline
\end{tabular}
\caption{Test accuracy on FEMNIST-3-groups. Each experiment is repeated 5 times. Discrepancy is the difference between the largest accuracy and the smallest accuracy.}
\label{table:femnist-3}
\end{table}





\subsection{Text Data}

\textbf{Individual Fairness:} We train a RNN to predict the next character using text data built from ``The Complete Works of William Shakespeare". In this dataset, there are about 1,129 speaking roles. Naturally, each speaking role in the play is treated as a device. Each device stored several text data and those information will be used to train a RNN on each device. The dataset is available on the LEAF website \citep{caldas2018leaf}.

Following the setting in \cite{mcmahan2017communication} and \cite{ li2019fair}, we subsample 31 roles ($d=31$). The RNN model takes a 80-character sequence as the input, and outputs one character after two LSTM layers and one densely-connected layer. For \texttt{FedAvg}, \texttt{q-FFL} and \texttt{Ditto}, the best initial learning rate is 0.8 and decay rate is 0.95 \citep{li2021ditto}. We also adopt this setting to \texttt{GIFAIR-FL-Global} and \texttt{GIFAIR-FL-Per}. The batch size is set to be 10. The number of local epochs is fixed to be 1 and all models are trained for 500 epochs. Results are reported in Table \ref{table:shakes-1}.

\begin{table}[!htbp]
\centering
\footnotesize
\begin{tabular}{cccccccc}
\hline
\textbf{Algorithm} & \texttt{FedAvg} & \texttt{q-FFL} &  \texttt{AFL} & \texttt{Ditto} & \texttt{GIFAIR-FL-Global} & \texttt{GIFAIR-FL-Per} \\ \hline
$\bar{a}$ & 53.21 (0.31)  & 53.90 (0.30) & 54.58 (0.14) & 60.74 (0.42) & 57.04 (0.23)  &  \textbf{61.58 (0.14)}  \\ \hline
$\sqrt{Var(a)}$ & 9.25 (6.17) & 7.52 (5.10) & 8.44 (5.65) & 8.32 (4.77) & \textbf{3.14 (1.25)} & 4.33 (1.25)\\ \hline
\end{tabular}
\caption{Mean and standard deviation of test accuracy on Shakespeare $(d=31)$. Each experiment is repeated 5 times.}
\label{table:shakes-1}
\end{table}

\textbf{Group Fairness:} We obtain the gender information from \texttt{https://shakespeare.folger.edu/} and group speaking roles based on gender ($d=2$). It is known that the majority of characters in Shakespearean drama are males. Simply training a \texttt{FedAvg} model on this dataset will cause implicit bias towards male characters. On a par with this observation, we subsample 25 males and 10 females from ``The Complete Works of William Shakespeare". Here we note that each device in the male group implicitly has more text data. The setting of hyperparameters is same as that of individual fairness. Results are reported in Table \ref{table:shakes-2}.

\begin{table}[!htbp]
\centering
\footnotesize
\begin{tabular}{ccccccccc}
\hline
\textbf{Algorithm} & \texttt{FedAvg} & \texttt{q-FFL} &  \texttt{FedMGDA+} & \texttt{Ditto} &  \texttt{GIFAIR-FL-Global} & \textbf{GIFAIR-FL-Per}  \\ \hline
Male & 72.95 (1.70) & 67.14 (2.18) & 67.07 (2.11) & \textbf{74.19 (3.75)} & 67.42(0.98)  & 73.95 (0.59)\\ \hline
Female & 40.39 (1.49) & 43.26 (2.05) & 43.85 (2.32) & 45.73 (4.01) & 52.04 (1.10) &  \textbf{54.88 (1.12)}\\ \hline
Discrepancy & 32.56 & 23.88 & 23.22 & 28.46 & \textbf{15.38} & 19.07 \\ \hline
\end{tabular}
\caption{Test accuracy on Shakespeare ($d=2$). Each experiment is repeated 5 times. }
\label{table:shakes-2}
\end{table}

\subsection{Analysis of Results}

Based on Table \ref{table:femnist-1}-\ref{table:shakes-2}, we can obtain important insights. First, compared to other benchmark models, \texttt{GIFAIR-FL-Global}/\texttt{GIFAIR-FL-Per} lead to significantly more fair solutions. As shown in Tables \ref{table:femnist-1}, \ref{table:femnist-2} and \ref{table:shakes-1}, our algorithm significantly reduces the variance of testing accuracy of all devices (i.e., $Var(a)$) while the average testing accuracy remains consistent. Second, from Tables \ref{table:femnist-3} and \ref{table:shakes-2}, it can be seen that \texttt{GIFAIR-FL-Global}/\texttt{GIFAIR-FL-Per} boosted the performance of the group with the worst testing accuracy and achieved the smallest discrepancy. Notably, this boost did not affect the performance of other groups. This indicates that \texttt{GIFAIR-FL-Global}/\texttt{GIFAIR-FL-Per} is capable of ensuring fairness among different groups while retaining a superior or similar prediction accuracy compared to existing benchmark models. Finally, we note that \texttt{GIFAIR-FL-Global} sometimes achieves lower prediction performance than \texttt{Ditto}. This is understandable as \texttt{Ditto} provides a personalized solution to each device $k$ while our model only returns a global parameter $\bm{\bar{\theta}}$. Yet, as shown in the last column, if we use \texttt{GIFAIR-FL-Per}, then the prediction performance can be significantly improved without sacrificing fairness. However, even without personalization, \texttt{GIFAIR-FL-Global} achieves superior testing performance compared to existing fair FL benchmark models.



\subsection{Sensitivity Analysis}
\label{subsec:lambda}

 \begin{wrapfigure}{R}{0.4\textwidth}
    \vskip -0.2in
    \centering
    \centerline{\includegraphics[width=0.4\columnwidth]{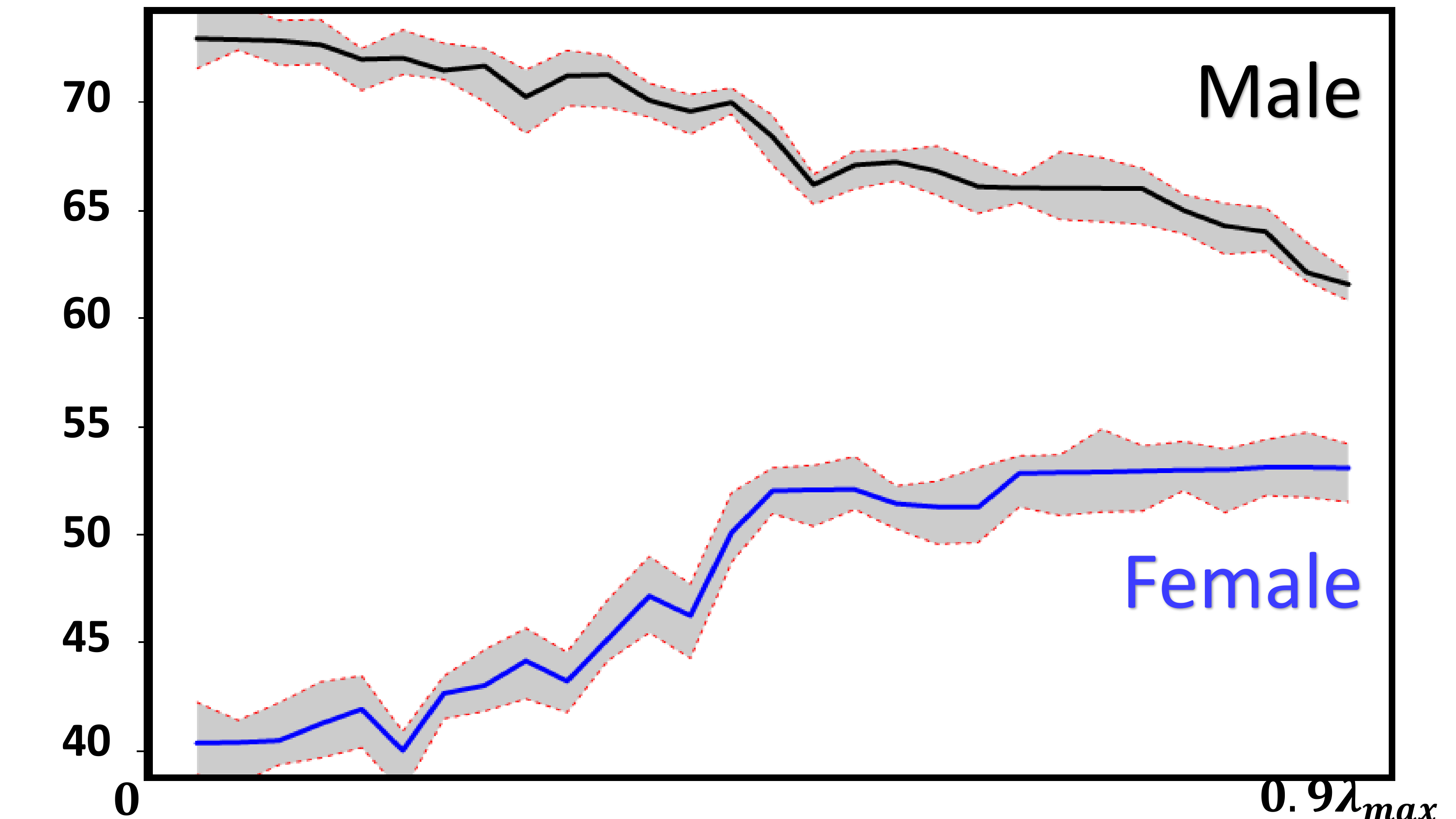}}
    \caption{Sensitivity with respect to $\lambda$ (Shakespeare Dataset).}
    \label{fig:lambda}
    \vskip -0.1in
\end{wrapfigure}

In this section, we use \texttt{GIFAIR-FL-Global} to study the effect of the tuning parameter $\lambda\in[0,\lambda_{max})$ using the Shakespeare dataset. A similar conclusion holds for \texttt{GIFAIR-FL-Per} and we therefore omit it. Results are reported in Figure \ref{fig:lambda}. It can be seen that as $\lambda$ increases, the discrepancy between male and female groups decreases accordingly. However, after $\lambda$ passes a certain threshold, the averaged testing accuracy of the female group remained flat yet the performance of the male group significantly dropped. Therefore, in practice, it is recommended to consider a moderate $\lambda$ value. Intuitively, when $\lambda=0$, \texttt{GIFAIR-FL} becomes \texttt{FedAvg}. When $\lambda$ is close to $\lambda_{max}$, the coefficient (i.e., $(1+\lambda\frac{1}{p_k|{\cal A}_{s_k}|}r_k)$) of devices with good performance will be close to zero and the updating is, therefore, impeded. A moderate $\lambda$ balances those two situations well. Besides this example, we also conducted additional sensitivity analysis. Due to space limitation, we defer those results to the Appendix. 

\section{Conclusion}
\label{sec:con}

In this paper, we propose \texttt{GIFAIR-FL}: a framework that imposes group and individual fairness to FL. Experiments show that \texttt{GIFAIR-FL} can lead to more fair solutions compared to recent state-of-the-art fair and personalized FL algorithms while retaining similar testing performance. To the best of our knowledge, fairness in FL is an under underinvestigated area and we hope our work will help inspire continued exploration into fair FL algorithms.


Also, real-life FL datasets for specific engineering or health science applications are still scarce. This is understandable as FL efforts have mainly focused on mobile applications. As such we only test on image classification and text prediction datasets. However, as FL is expected to infiltrate many applications, we hope that more real-life datasets will be generated to provide a means for model validation within different domains. We plan to actively pursue this direction in future research. 

\newpage

\section*{Appendix}

\section{Assumptions}
\label{app:assumption}
We make the following assumptions. 
\begin{assumption}
\label{assumption:11}
$F_k$ is $L$-smooth and $\mu$-strongly convex for all $k\in[K]$.
\end{assumption}

\begin{assumption}
\label{assumption:22}
Denote by $\zeta^{(t)}_k$ the batched data from client $k$ and $ g_k(\bm{\theta}^{(t)}_k;\zeta^{(t)}_k)$ the stochastic gradient calculated on this batched data. The variance of stochastic gradients are bounded. Specifically,
\begin{align*}
    \mathbb{E}\bigg\{\norm{g_k(\bm{\theta}^{(t)}_k;\zeta^{(t)}_k)-\nabla F_k(\bm{\theta}^{(t)}_k)}^2\bigg\}\leq\sigma_k^2, \forall k\in[K].
\end{align*}
\end{assumption}
It can be shown that, at local iteration $t$ during communication round $c$, 
\begin{align*}
    &\mathbb{E}\bigg\{\norm{\nabla H_k(\bm{\theta}^{(t)}_k;\zeta^{(t)}_k)-\nabla H_k(\bm{\theta}^{(t)}_k)}^2\bigg\}\\
    &=\mathbb{E}\bigg\{\norm{(1+\frac{\lambda  r^c_k}{p_k|{\cal A}_{s_k}|})g_k(\bm{\theta}^{(t)}_k;\zeta^{(t)}_k)-(1+\frac{\lambda r_k^c}{p_k|{\cal A}_{s_k}|})\nabla F_k(\bm{\theta}^{(t)}_k)}^2\bigg\}\\
    &\leq(1+\frac{\lambda}{p_k|{\cal A}_{s_k}|} r_k^c)^2\sigma_k^2, \forall k\in[K].
\end{align*}
Here, $\nabla H_k(\bm{\theta}^{(t)}_k;\zeta^{(t)}_k)$ denotes the stochastic gradient of $H_k$ evaluated on the batched data $\zeta^{(t)}_k$.
\begin{assumption}
\label{assumption:33}
The expected squared norm of stochastic gradient is bounded. Specifically,
\begin{align*}
    \mathbb{E}\bigg\{\norm{g_k(\bm{\theta}^{(t)}_k;\zeta^{(t)}_k)}^2\bigg\}\leq G^2, \forall k\in[K].
\end{align*}
\end{assumption}
It can be shown that, at local iteration $t$ during communication round $c$, 
\begin{align*}
    &\mathbb{E}\bigg\{\norm{\nabla H_k(\bm{\theta}^{(t)}_k;\zeta^{(t)}_k)}\bigg\}=\mathbb{E}\bigg\{\norm{(1+\frac{\lambda}{p_k|{\cal A}_{s_k}|} r_k^c)g_k(\bm{\theta}^{(t)}_k;\zeta^{(t)}_k)}^2\bigg\}\\
    &\leq (1+\frac{\lambda}{p_k|{\cal A}_{s_k}|} r_k^c)^2G^2, \forall k\in[K].
\end{align*}

For the non-convex setting, we replace Assumption \ref{assumption:11} by the following assumption.
\begin{assumption}
\label{assumption:44}
$F_k$ is $L$-smooth for all $k\in[K]$.
\end{assumption}
In our proof, for the sake of neatness, we drop the superscript of $r^c_k$.

We use the definition in \citet{li2019convergence} to roughly quantify the degree of non-$i.i.d.$-ness. Specifically, 
\begin{align*}
    \Gamma_K = H^* - \sum_{k=1}^Kp_kH^*_k= \sum_{k=1}^K p_k(H^*-H^*_k).
\end{align*}
If data from all sensitive attributes are $i.i.d.$, then $\Gamma_K=0$ as number of clients grows. Otherwise, $\Gamma_K\neq 0$ \citep{li2019convergence}.

\section{Detailed Proof}
\label{app:proof}

\subsection{Proof of Lemma}
\begin{lemma}
For any given $\bm{\theta}$, the global objective function $H(\bm{\theta})$ defined in the main paper can be expressed as
\[H(\bm{\theta}) = \sum_{k=1}^K \left(p_k + \dfrac{\lambda}{|{\cal A}_{s_k}|} r_k(\bm{\theta}) \right)F_k(\bm{\theta}),  \]
where 
\begin{align*}
r_k(\bm{\theta}) \triangleq \sum_{1\leq j \neq s_k \leq d } {\normalfont\mbox{sign}}(L_{s_k}(\bm{\theta}) - L_j(\bm{\theta})) 
\end{align*}
and $s_k\in[d]$ is the group index of device $k$. Consequently, \[H(\bm{\theta}) = \sum_{k=1}^K \, p_k H_k(\bm{\theta}). \]
\end{lemma}

\textbf{Proof}
By definition, at communication round $c$,
\begin{align*}
    H(\bm{\theta})&=\sum_{k=1}^K  p_kF_k(\bm{\theta})  + \lambda\sum_{1\leq i < j\leq d}| L_{i}(\bm{\theta})- L_{j}(\bm{\theta})|\\
    &=\sum_{k=1}^K  p_kF_k(\bm{\theta})  + \lambda\sum_{1\leq i < j\leq d}\bigg| \dfrac{1}{|{\cal A}_i|} \sum_{k \in {\cal A}_i} F_k(\bm{\theta})- \dfrac{1}{|{\cal A}_j|} \sum_{k \in {\cal A}_j} F_k(\bm{\theta})\bigg|\\
    &=\sum_{k=1}^K  p_kF_k(\bm{\theta})  + \lambda\sum_{1\leq i < j\leq d}\mbox{sign}(L_{i}(\bm{\theta})- L_{j}(\bm{\theta}))\bigg( \dfrac{1}{|{\cal A}_i|} \sum_{k \in {\cal A}_i} F_k(\bm{\theta})- \dfrac{1}{|{\cal A}_j|} \sum_{k \in {\cal A}_j} F_k(\bm{\theta})\bigg)\\
    &=\sum_{k=1}^K  p_kF_k(\bm{\theta})  + \lambda\sum_{u=1}^{d-1}\sum_{u < j\leq d}\mbox{sign}(L_{u}(\bm{\theta})- L_{j}(\bm{\theta}))\bigg( \dfrac{1}{|{\cal A}_u|} \sum_{k \in {\cal A}_u} F_k(\bm{\theta})- \dfrac{1}{|{\cal A}_j|} \sum_{k \in {\cal A}_j} F_k(\bm{\theta})\bigg)\\
    &=\sum_{k=1}^K  p_kF_k(\bm{\theta}) + \lambda\sum_{u=1}^{d}\sum_{k\in{\cal A}_u}\sum_{u \neq j\leq d}\mbox{sign}(L_{u}(\bm{\theta})- L_{j}(\bm{\theta}))\frac{F_k(\bm{\theta})}{|{\cal A}_u|}\\
    &=\sum_{k=1}^K  p_kF_k(\bm{\theta})  + \sum_{k=1}^K\frac{\lambda}{|{\cal A}_{s_k}|}\sum_{1\leq j \neq s_k \leq d } \mbox{sign}(L_{s_k}(\bm{\theta}) - L_j(\bm{\theta})) F_k(\bm{\theta})\\
    &= \sum_{k=1}^K \left(p_k + \dfrac{\lambda}{|{\cal A}_{s_k}|} r^c_k(\bm{\theta}) \right)F_k(\bm{\theta}).
\end{align*}
The fifth equality is achieved by rearranging the equation and merging items with the same group label. By definition of $H_k$, we thus proved
\begin{align*}
    H(\bm{\theta}) = \sum_{k=1}^K \, p_k H_k(\bm{\theta}).
\end{align*}

\subsection{Learning Bound}

We present a generalization bound for our learning model. Denote by $\mathcal{G}$ the family of the losses associated to a hypothesis set $\mathcal{H}:\mathcal{G}=\{(x,y)\mapsto \ell(h(x),y):h\in\mathcal{H}\}$. The weighted Rademacher complexity \citep{mohri2019agnostic} is defined as
\begin{align*}
    \mathfrak{R}_{\bm{m}}(\mathcal{G}, \bm{p})\coloneqq\mathop{\mathbb{E}}_{\bm{\sigma}}\bigg[\sup_{h\in\mathcal{H}}\sum_{k=1}^K\frac{p_k}{N_k}\sum_{n=1}^{N_k}\sigma_{k,n}\ell(h(x_{k,n}),y_{k,n})\bigg]
\end{align*}
where $\bm{m}=(N_1,N_2,\ldots,N_k)$, $\bm{p}=(p_1,\ldots,p_K)$ and $\bm{\sigma}=(\sigma_{k,n})_{k\in[K],n\in[N_k]}$ is a collection of Rademacher variables taking values in $\{-1,+1\}$. Denote by $\mathcal{L}_{\mathcal{D}^{\lambda}_{\bm{p}}}(h)$ the expected loss according to our fairness formulation. Denote by $\hat{\mathcal{L}}_{\mathcal{D}^{\lambda}_{\bm{p}}}(h)$ the expected empirical loss (See Appendix for a detailed expression).

\begin{theorem}
\label{theorem:learning_bound}
Assume that the loss $\ell$ is bounded above by $M>0$. Fix $\epsilon_0>0$ and $\bm{m}$. Then, for any $\delta_0>0$, with probability at least $1-\delta_0$ over samples $D_k\sim\mathcal{D}_k$, the following holds for all $h\in\mathcal{H}$:
\begin{align*}
    \mathcal{L}_{\mathcal{D}^{\lambda}_{\bm{p}}}(h)\leq \hat{\mathcal{L}}_{\mathcal{D}^{\lambda}_{\bm{p}}}(h)+ \sqrt{\frac{1}{2}\sum_{k=1}^K( \frac{p_k}{N_k}M+\lambda\frac{d(d-1)}{2}M)^2\log\frac{1}{\delta_0}} + 2\mathfrak{R}_{\bm{m}}(\mathcal{G}, \bm{p}) +\lambda\frac{d(d-1)}{2}M.
\end{align*}
\end{theorem}
It can be seen that, given a sample of data, we can bound the generalization error $\mathcal{L}_{\mathcal{D}^{\lambda}_{\bm{p}}}(h)-\hat{\mathcal{L}}_{\mathcal{D}^{\lambda}_{\bm{p}}}(h)$ with high probability. When $\lambda=0$, the bound is same as the generalization bound in \texttt{FedAvg} \citep{mohri2018foundations}. When we consider the worst combination of $p_k$ by taking the supremum of the upper bound in Theorem \ref{theorem:learning_bound} and let $\lambda=0$, then our generalization bound is same as the one in \texttt{AFL} \citep{mohri2019agnostic}.

\textbf{Proof} Define
\begin{align*}
    \Phi(D_1,\ldots,D_K)=\sup_{h\in\mathcal{H}}\bigg(\mathcal{L}_{\mathcal{D}^{\lambda}_{\bm{p}}}(h)-\hat{\mathcal{L}}_{\mathcal{D}^{\lambda}_{\bm{p}}}(h)\bigg).
\end{align*}

Let $D'=(D_1',\ldots,D_K')$ be a sample differing from $D=(D_1,\ldots,D_K)$ only by one point $x_{k,n}'$. Therefore, we have
\begin{align*}
    \Phi(D')-\Phi(D)&=\sup_{h\in\mathcal{H}}\bigg(\mathcal{L}_{\mathcal{D}^{\lambda}_{\bm{p}}}(h)-\hat{\mathcal{L}}_{\mathcal{D}^{'\lambda}_{\bm{p}}}(h)\bigg)-\sup_{h\in\mathcal{H}}\bigg(\mathcal{L}_{\mathcal{D}^{\lambda}_{\bm{p}}}(h)-\hat{\mathcal{L}}_{\mathcal{D}^{\lambda}_{\bm{p}}}(h)\bigg)\\
    &\leq\sup_{h\in\mathcal{H}}\bigg(\mathcal{L}_{\mathcal{D}^{\lambda}_{\bm{p}}}(h)-\hat{\mathcal{L}}_{\mathcal{D}^{'\lambda}_{\bm{p}}}(h)\bigg)-\bigg(\mathcal{L}_{\mathcal{D}^{\lambda}_{\bm{p}}}(h)-\hat{\mathcal{L}}_{\mathcal{D}^{\lambda}_{\bm{p}}}(h)\bigg)\\
    &\leq\sup_{h\in\mathcal{H}}\big\{\sup_{h\in\mathcal{H}}\mathcal{L}_{\mathcal{D}^{\lambda}_{\bm{p}}}(h)-\sup_{h\in\mathcal{H}}\hat{\mathcal{L}}_{\mathcal{D}^{'\lambda}_{\bm{p}}}(h)-\mathcal{L}_{\mathcal{D}^{\lambda}_{\bm{p}}}(h)+\hat{\mathcal{L}}_{\mathcal{D}^{\lambda}_{\bm{p}}}(h)\big\}\\
    &=\sup_{h\in\mathcal{H}}\big\{\hat{\mathcal{L}}_{\mathcal{D}^{\lambda}_{\bm{p}}}(h)-\hat{\mathcal{L}}_{\mathcal{D}^{'\lambda}_{\bm{p}}}(h)\big\}
\end{align*}
By definition,
\begin{align*}
    \hat{\mathcal{L}}_{\mathcal{D}^{'\lambda}_{\bm{p}}}(h)&=\sum_{k=1}^K\frac{p_k}{N_k}\sum_{n=1}^{N_k}\ell(h(x'_{k,n}),y'_{k,n}) + \\
    & \lambda\sum_{1\leq i < j\leq d}| \frac{\sum_{k\in{\cal A}_{i}}\frac{1}{N_k}\sum_{n=1}^{N_k}\ell(h(x'_{k,n}),y'_{k,n})}{|{\cal A}_{i}|}- \frac{\sum_{k\in{\cal A}_j}\frac{1}{N_k}\sum_{n=1}^{N_k}\ell(h(x'_{k,n}),y'_{k,n})}{|{\cal A}_j|}|. 
\end{align*}
Therefore,
\begin{align*}
    &\sup_{h\in\mathcal{H}}\big\{\hat{\mathcal{L}}_{\mathcal{D}^{\lambda}_{\bm{p}}}(h)-\hat{\mathcal{L}}_{\mathcal{D}^{'\lambda}_{\bm{p}}}(h)\big\}\\
    &\leq\sup_{h\in\mathcal{H}}\bigg[\frac{p_k}{N_k}(\ell(h(x'_{k,n}),y'_{k,n})-\ell(h(x_{k,n}),y_{k,n}))+\lambda\frac{d(d-1)}{2}M\bigg]\\
    &\leq \frac{p_k}{N_k}M+\lambda\frac{d(d-1)}{2}M.
\end{align*}
By McDiarmid's inequality, for $\delta_0=\exp\bigg(\frac{-2\epsilon_0^2}{\sum_{k=1}^K( \frac{p_k}{N_k}M+\lambda\frac{d(d-1)}{2}M)^2}\bigg)$, the following holds with probability at least $1-\delta_0$
\begin{align*}
    \Phi(D)-\mathbb{E}_D[\Phi(D)]\leq\epsilon_0=\sqrt{\frac{1}{2}\sum_{k=1}^K( \frac{p_k}{N_k}M+\lambda\frac{d(d-1)}{2}M)^2\log\frac{1}{\delta_0}}.
\end{align*}
Our next goal is to bound $\mathbb{E}[\Phi(D)]$. We have
\begin{align*}
    \mathbb{E}_{D}[\Phi(D)]&=\mathbb{E}_D\bigg[\sup_{h\in\mathcal{H}}\bigg(\mathcal{L}_{\mathcal{D}^{\lambda}_{\bm{p}}}(h)-\hat{\mathcal{L}}_{\mathcal{D}^{\lambda}_{\bm{p}}}(h)\bigg)\bigg]\\
    &=\mathbb{E}_D\bigg[\sup_{h\in\mathcal{H}}\mathbb{E}_{D'}\bigg(\hat{\mathcal{L}}_{\mathcal{D}^{'\lambda}_{\bm{p}}}(h)-\hat{\mathcal{L}}_{\mathcal{D}^{\lambda}_{\bm{p}}}(h)\bigg)\bigg]\\
    &\leq\mathbb{E}_D\mathbb{E}_{D'}\sup_{h\in\mathcal{H}}\bigg(\hat{\mathcal{L}}_{\mathcal{D}^{'\lambda}_{\bm{p}}}(h)-\hat{\mathcal{L}}_{\mathcal{D}^{\lambda}_{\bm{p}}}(h)\bigg)\\
    &\leq\mathbb{E}_D\mathbb{E}_{D'}\sup_{h\in\mathcal{H}}\bigg[\sum_{k=1}^K\frac{p_k}{N_k}\sum_{n=1}^{N_k}\ell(h(x'_{k,n}),y'_{k,n})-\sum_{k=1}^K\frac{p_k}{N_k}\sum_{n=1}^{N_k}\ell(h(x_{k,n}),y_{k,n})+\lambda\frac{d(d-1)}{2}M\bigg]\\
    &\leq\mathbb{E}_D\mathbb{E}_{D'}\mathbb{E}_{\bm{\sigma}}\sup_{h\in\mathcal{H}}\bigg[\sum_{k=1}^K\frac{p_k}{N_k}\sum_{n=1}^{N_k}\sigma_{k,n}\ell(h(x'_{k,n}),y'_{k,n})-\sum_{k=1}^K\frac{p_k}{N_k}\sum_{n=1}^{N_k}\sigma_{k,n}\ell(h(x_{k,n}),y_{k,n})+\lambda\frac{d(d-1)}{2}M\bigg]\\
    &\leq 2\mathfrak{R}_{\bm{m}}(\mathcal{G}, \bm{p}) +\lambda\frac{d(d-1)}{2}M.
\end{align*}
Therefore,
\begin{align*}
    \Phi(D)\leq \sqrt{\frac{1}{2}\sum_{k=1}^K( \frac{p_k}{N_k}M+\lambda\frac{d(d-1)}{2}M)^2\log\frac{1}{\delta_0}} + 2\mathfrak{R}_{\bm{m}}(\mathcal{G}, \bm{p}) +\lambda\frac{d(d-1)}{2}M.
\end{align*}

\subsection{Convergence (Strongly Convex)}
\label{app:convex}

Our proof is based on the convergence result of \texttt{FedAvg} \citep{li2019convergence}.

\begin{theorem}
\label{theorem:conv_full_1}
Assume Assumptions in the main paper hold and $|\mathcal{S}_c|=K$. For $\gamma,\mu>0$ and $\eta^{(t)}$ is decreasing in a rate of $\mathcal{O}(\frac{1}{t})$. If $\eta^{(t)}\leq\mathcal{O}(\frac{1}{L})$, we have
\begin{align*}
    \mathbb{E}\bigg\{H(\bm{\bar{\theta}}^{(T)})\bigg\} - H^*\leq \frac{L}{2}\frac{1}{\gamma+T}\bigg\{ \frac{4\xi}{\epsilon^2\mu^2}+(\gamma+1)\norm{\bm{\bar{\theta}}^{(0)}-\bm{\theta}^*}^2\bigg\},
\end{align*}
where $\xi=8(E-1)^2G^2+4L\Gamma_K+2\frac{\Gamma_{max}}{\eta^{(t)}}+4\sum_{k=1}^Kp_k^2\sigma_k^2$ and $\Gamma_{max}\coloneqq\sum_{k=1}^Kp_k|(H^*-H^*_k)|\geq|\sum_{k=1}^K p_k(H^*-H^*_k)|= |\Gamma_K|$.
\end{theorem}

\textbf{Proof} 

For each device $k$, we introduce an intermediate model parameter $\bm{w}^{(t+1)}_k=\bm{\theta}^{(t)}_k-\eta^{(t)}\nabla H_k(\bm{\theta}^{(t)}_k)$. If iteration $t+1$ is in the communication round, then $\bm{\theta}^{(t+1)}_k=\sum_{k=1}^Kp_k\bm{w}^{(t+1)}_k$ (i.e., aggregation). Otherwise, $\bm{\theta}^{(t+1)}_k=\bm{w}^{(t+1)}_k$. Define $\bm{\bar{w}}^{(t)}=\sum_{k=1}^Kp_k\bm{w}^{(t)}_k$ and $\bm{\bar{\theta}}^{(t)}=\sum_{k=1}^Kp_k\bm{\theta}^{(t)}_k$. Also, define $\bm{g}^{(t)}=\sum_{k=1}^Kp_k\nabla H_k(\bm{\theta}^{(t)}_k;\zeta^{(t)}_k)$ and $\bm{\bar{g}}^{(t)}=\mathbb{E}(\bm{g}^{(t)})=\sum_{k=1}^Kp_k\nabla H_k(\bm{\theta}^{(t)}_k)$.

Denote by $\bm{\theta}^*$ the optimal model parameter of the global objective function $H(\cdot)$. At iteration $t$, we have
\begin{align*}
    &\mathbb{E}\bigg\{\norm{\bm{\bar{\theta}}^{(t+1)}-\bm{\theta}^*}^2\bigg\}=\mathbb{E}\bigg\{\norm{\bm{\bar{\theta}}^{(t)}-\eta^{(t)}\bm{g}^{(t)} -\bm{\theta}^*-\eta^{(t)}\bm{\bar{g}}^{(t)}+\eta^{(t)}\bm{\bar{g}}^{(t)}}^2\bigg\}\\
    &=\mathbb{E}\bigg\{\norm{\bm{\bar{\theta}}^{(t)}-\bm{\theta}^*-\eta^{(t)}\bm{\bar{g}}^{(t)}}^2\bigg\} + \mathbb{E}\bigg\{2\eta^{(t)}\langle\bm{\bar{\theta}}^{(t)}-\bm{\theta}^*-\eta^{(t)}\bm{\bar{g}}^{(t)},\bm{\bar{g}}^{(t)}-\bm{g}^{(t)}\rangle\bigg\} +\mathbb{E}\bigg\{\eta^{(t)2}\norm{\bm{g}^{(t)}-\bm{\bar{g}}^{(t)}}^2\bigg\}\\
    &= \underbrace{\mathbb{E}\bigg\{\norm{\bm{\bar{\theta}}^{(t)}-\bm{\theta}^*-\eta^{(t)}\bm{\bar{g}}^{(t)}}^2\bigg\}}_{A} +  \underbrace{ \mathbb{E}\bigg\{\eta^{(t)2}\norm{\bm{g}^{(t)}-\bm{\bar{g}}^{(t)}}^2\bigg\}}_B,
\end{align*}
since $\mathbb{E}\bigg\{2\eta^{(t)}\langle\bm{\bar{\theta}}^{(t)}-\bm{\theta}^*-\eta^{(t)}\bm{\bar{g}}^{(t)},\bm{\bar{g}}^{(t)}-\bm{g}^{(t)}\rangle\bigg\}=0$. Our remaining work is to bound term $A$ and term $B$.

\paragraph{Part I: Bounding Term $A$} 

We can split term $A$ above into three parts:
\begin{align*}
    \mathbb{E}\bigg\{\norm{\bm{\bar{\theta}}^{(t)}-\bm{\theta}^*-\eta^{(t)}\bm{\bar{g}}^{(t)}}^2\bigg\}=\mathbb{E}\bigg\{\norm{\bm{\bar{\theta}}^{(t)}-\bm{\theta}^*}^2\bigg\}\underbrace{-2\eta^{(t)}\mathbb{E}\bigg\{\langle\bm{\bar{\theta}}^{(t)}-\bm{\theta}^*,\bm{\bar{g}}^{(t)}\rangle\bigg\}}_{\text{C}}+\underbrace{\eta^{(t)2}\mathbb{E}\bigg\{\norm{\bm{\bar{g}}^{(t)}}^2\bigg\}}_{\text{D}}.
\end{align*}
For part C, We have
\begin{align*}
    \text{C}&=-2\eta^{(t)}\mathbb{E}\bigg\{\langle\bm{\bar{\theta}}^{(t)}-\bm{\theta}^*,\bm{\bar{g}}^{(t)}\rangle\bigg\}=-2\eta^{(t)}\mathbb{E}\bigg\{\sum_{k=1}^Kp_k\langle\bm{\bar{\theta}}^{(t)}-\bm{\theta}^*,\nabla H_k(\bm{\theta}_k^{(t)})\rangle\bigg\}\\
    &=-2\eta^{(t)}\mathbb{E}\bigg\{\sum_{k=1}^Kp_k\langle\bm{\bar{\theta}}^{(t)}-\bm{\theta}_k^{(t)},\nabla H_k(\bm{\theta}_k^{(t)})\rangle\bigg\}-2\eta^{(t)}\mathbb{E}\bigg\{\sum_{k=1}^Kp_k\langle\bm{\theta}_k^{(t)}-\bm{\theta}^*,\nabla H_k(\bm{\theta}_k^{(t)})\rangle\bigg\}
\end{align*}
To bound C, we need to use Cauchy-Schwarz inequality, inequality of arithmetic and geometric means. Specifically, the Cauchy-Schwarz inequality indicates that 
\begin{align*}
    \langle\bm{\bar{\theta}}^{(t)}-\bm{\theta}_k^{(t)},\nabla H_k(\bm{\theta}_k^{(t)})\rangle\geq - \norm{\bm{\bar{\theta}}^{(t)}-\bm{\theta}_k^{(t)}}\norm{\nabla H_k(\bm{\theta}_k^{(t)})}
\end{align*}
and inequality of arithmetic and geometric means further implies
\begin{align*}
    - \norm{\bm{\bar{\theta}}^{(t)}-\bm{\theta}_k^{(t)}}\norm{\nabla H_k(\bm{\theta}_k^{(t)})} \geq - \frac{\norm{\bm{\bar{\theta}}^{(t)}-\bm{\theta}_k^{(t)}}^2+\norm{\nabla H_k(\bm{\theta}_k^{(t)})}^2}{2}.
\end{align*}

Therefore, we obtain
\begin{align*}
    \text{C}&=-2\eta^{(t)}\mathbb{E}\bigg\{\langle\bm{\bar{\theta}}^{(t)}-\bm{\theta}^*,\bm{\bar{g}}^{(t)}\rangle\bigg\}=-2\eta^{(t)}\mathbb{E}\bigg\{\sum_{k=1}^Kp_k\langle\bm{\bar{\theta}}^{(t)}-\bm{\theta}^*,\nabla H_k(\bm{\theta}_k^{(t)})\rangle\bigg\}\\
    &=-2\eta^{(t)}\mathbb{E}\bigg\{\sum_{k=1}^Kp_k\langle\bm{\bar{\theta}}^{(t)}-\bm{\theta}_k^{(t)},\nabla H_k(\bm{\theta}_k^{(t)})\rangle\bigg\}-2\eta^{(t)}\mathbb{E}\bigg\{\sum_{k=1}^Kp_k\langle\bm{\theta}_k^{(t)}-\bm{\theta}^*,\nabla H_k(\bm{\theta}_k^{(t)})\rangle\bigg\}\\
    &\leq \mathbb{E}\bigg\{\eta^{(t)}\sum_{k=1}^Kp_k\frac{1}{\eta^{(t)}}\norm{\bm{\bar{\theta}}^{(t)}-\bm{\theta}_k^{(t)}}^2+\eta^{(t)^2}\sum_{k=1}^Kp_k\norm{\nabla H_k(\bm{\theta}_k^{(t)})}^2 \\
    &\qquad -2\eta^{(t)}\sum_{k=1}^Kp_k(H_k(\bm{\theta}_k^{(t)})-H_k(\bm{\theta}^*))-2\eta^{(t)}\sum_{k=1}^Kp_k\frac{(1+\frac{\lambda}{p_k|{\cal A}_{s_k}|} r_k(\bm{\theta}))\mu}{2}\norm{\bm{\theta}_k^{(t)}-\bm{\theta}^*}^2\bigg\},
\end{align*}
where $-2\eta^{(t)}\mathbb{E}\bigg\{\sum_{k=1}^Kp_k\langle\bm{\theta}_k^{(t)}-\bm{\theta}^*,\nabla H_k(\bm{\theta}_k^{(t)})\rangle\bigg\}$ is bounded by the property of strong convexity of $H_k$.

Since $H_k$ is $(1+\frac{\lambda}{p_k|{\cal A}_{s_k}|} r_k(\bm{\theta}))L$-smooth, we know
\begin{align*}
    \norm{\nabla H_k(\bm{\theta}_k^{(t)})}^2\leq 2(1+\frac{\lambda}{p_k|{\cal A}_{s_k}|} r_k(\bm{\theta}))L(H_k(\bm{\theta}_k^{(t)})-H_k^*)
\end{align*}
and therefore
\begin{align*}
    \text{D}&=\eta^{(t)2}\mathbb{E}\bigg\{\norm{\bm{\bar{g}}^{(t)}}^2\bigg\}\leq \eta^{(t)2}\mathbb{E}\bigg\{\sum_{k=1}^Kp_k\norm{\nabla H_k(\bm{\theta}_k^{(t)})}^2\bigg\}\\
    &\leq 2\eta^{(t)2}\mathbb{E}\bigg\{ \sum_{k=1}^Kp_k(1+\frac{\lambda}{p_k|{\cal A}_{s_k}|} r_k(\bm{\theta}))L(H_k(\bm{\theta}_k^{(t)})-H_k^*)\bigg\}
\end{align*}
by convexity of norm. 

Therefore, combining C and D, we have
\begin{align*}
    A&=\mathbb{E}\bigg\{\norm{\bm{\bar{\theta}}^{(t)}-\bm{\theta}^*-\eta^{(t)}\bm{\bar{g}}^{(t)}}^2\bigg\} \\
    &\leq \mathbb{E}\bigg\{\norm{\bm{\bar{\theta}}^{(t)}-\bm{\theta}^*}^2\bigg\} + 2\eta^{(t)2}\mathbb{E}\bigg\{\sum_{k=1}^K p_k(1+\frac{\lambda}{p_k|{\cal A}_{s_k}|} r_k(\bm{\theta}))L(H_k(\bm{\theta}_k^{(t)})-H_k^*)\bigg\}\\
    &\qquad +\eta^{(t)}\mathbb{E}\bigg\{\sum_{k=1}^Kp_k\frac{1}{\eta^{(t)}}\norm{\bm{\bar{\theta}}^{(t)}-\bm{\theta}_k^{(t)}}^2\bigg\}+\eta^{(t)^2}\mathbb{E}\bigg\{\sum_{k=1}^Kp_k\norm{\nabla H_k(\bm{\theta}_k^{(t)})}^2\bigg\} \\
    &\qquad -2\eta^{(t)}\mathbb{E}\bigg\{\sum_{k=1}^Kp_k(H_k(\bm{\theta}_k^{(t)})-H_k(\bm{\theta}^*))\bigg\} - 2\eta^{(t)}\mathbb{E}\bigg\{\sum_{k=1}^Kp_k\frac{(1+\frac{\lambda}{p_k|{\cal A}_{s_k}|} r_k(\bm{\theta}))\mu}{2}\norm{\bm{\theta}_k^{(t)}-\bm{\theta}^*}^2\bigg\}\\
    &\leq\mathbb{E}\bigg\{\norm{\bm{\bar{\theta}}^{(t)}-\bm{\theta}^*}^2\bigg\} -\eta^{(t)}\mathbb{E}\bigg\{\sum_{k=1}^Kp_k(1+\frac{\lambda}{p_k|{\cal A}_{s_k}|} r_k(\bm{\theta}))\mu\norm{\bm{\theta}_k^{(t)}-\bm{\theta}^*}^2\bigg\} + \sum_{k=1}^Kp_k\norm{\bm{\bar{\theta}}^{(t)}-\bm{\theta}_k^{(t)}}^2\\
    &\qquad +\underbrace{4\eta^{(t)2}\mathbb{E}\bigg\{\sum_{k=1}^K p_k(1+\frac{\lambda}{p_k|{\cal A}_{s_k}|} r_k(\bm{\theta}))L(H_k(\bm{\theta}_k^{(t)})-H_k^*)\bigg\}-2\eta^{(t)}\mathbb{E}\bigg\{\sum_{k=1}^Kp_k(H_k(\bm{\theta}_k^{(t)})-H_k(\bm{\theta}^*))\bigg\}}_{\text{E}}.
\end{align*}
In the last inequality, we simply rearrange other terms and use the fact that $\norm{\nabla H_k(\bm{\theta}_k^{(t)})}^2\leq 2(1+\frac{\lambda}{p_k|{\cal A}_{s_k}|} r_k(\bm{\theta}))L(H_k(\bm{\theta}_k^{(t)})-H_k^*)$ as aforementioned.

To bound E, we define $\gamma^{(t)}_k=2\eta^{(t)}(1-2(1+\frac{\lambda}{p_k|{\cal A}_{s_k}|} r_k(\bm{\theta}))L\eta^{(t)})$. Assume $\eta^{(t)}\leq\frac{1}{4(1+\frac{(d-1)}{\min\{p_k|{\cal A}_{s_k}|\}}\lambda)L}$, then we know $\eta^{(t)}\leq\gamma_k^{(t)}\leq 2\eta^{(t)}$. 

Therefore, we have
\begin{align*}
    \text{E}&=4\eta^{(t)2}\mathbb{E}\bigg\{\sum_{k=1}^K p_k(1+\frac{\lambda}{p_k|{\cal A}_{s_k}|} r_k(\bm{\theta}))L(H_k(\bm{\theta}_k^{(t)})-H_k^*)\bigg\}-2\eta^{(t)}\mathbb{E}\bigg\{\sum_{k=1}^Kp_k(H_k(\bm{\theta}_k^{(t)})-H_k(\bm{\theta}^*))\bigg\}\\
    &=4\eta^{(t)2}\mathbb{E}\bigg\{\sum_{k=1}^K p_k(1+\frac{\lambda}{p_k|{\cal A}_{s_k}|} r_k(\bm{\theta}))L(H_k(\bm{\theta}_k^{(t)})-H_k^*)\bigg\} -2\eta^{(t)}\mathbb{E}\bigg\{\sum_{k=1}^Kp_k(H_k(\bm{\theta}_k^{(t)})-H_k^*+H_k^*-H_k(\bm{\theta}^*))\bigg\}\\
    &=-2\eta^{(t)}\mathbb{E}\bigg\{\sum_{k=1}^K p_k(1-2(1+\frac{\lambda}{p_k|{\cal A}_{s_k}|} r_k(\bm{\theta}))L\eta^{(t)})(H_k(\bm{\theta}_k^{(t)})-H_k^*)\bigg\} + 2\eta^{(t)}\mathbb{E}\bigg\{\sum_{k=1}^K p_k(H_k(\bm{\theta}^*)-H_k^*)\bigg\}\\
    &=-\mathbb{E}\bigg\{\sum_{k=1}^K \gamma^{(t)}_k p_k(H_k(\bm{\theta}_k^{(t)})-H^*+H^*-H^*_k)\bigg\} +  2\eta^{(t)}\mathbb{E}\bigg\{H^*-\sum_{k=1}^K p_kH_k^*\bigg\}\\
    &= \underbrace{-\mathbb{E}\bigg\{\sum_{k=1}^K \gamma_k^{(t)}p_k(H_k(\bm{\theta}_k^{(t)})-H^*)\bigg\}}_{\text{F}} + \underbrace{\  \mathbb{E}\bigg\{\sum_{k=1}^K (2\eta^{(t)}-\gamma_k^{(t)}) p_k(H^*-H^*_k)\bigg\}}_{\text{G}}.
\end{align*}
If $H^*-H^*_k\geq 0$ for some $k$, then $(2\eta^{(t)}-\gamma_k^{(t)}) p_k(H^*-H^*_k)\leq2\eta^{(t)} p_k(H^*-H^*_k)$. If $H^*-H^*_k< 0$ otherwise, then $(2\eta^{(t)}-\gamma_k^{(t)}) p_k(H^*-H^*_k)$ is negative and $(2\eta^{(t)}-\gamma_k^{(t)}) p_k(H^*-H^*_k)\leq -2\eta^{(t)}p_k(H^*-H^*_k)$. Therefore, by definition of $\Gamma_{max}$,
\begin{align*}
    \text{G}\leq 2\eta^{(t)}\mathbb{E}\bigg\{\sum_{k=1}^K  p_k|H^*-H^*_k|\bigg\}=2\eta^{(t)}\Gamma_{max}.
\end{align*}

The remaining goal of Part I is to bound term F. Note that
\begin{align*}
    \text{F}&=-\mathbb{E}\bigg\{\sum_{k=1}^K\gamma_k^{(t)} p_k(H_k(\bm{\theta}_k^{(t)})-H^*)\bigg\}\\
    &=-\mathbb{E}\bigg\{\bigg(\sum_{k=1}^K p_k\gamma_k^{(t)}(H_k(\bm{\theta}_k^{(t)})-H_k(\bm{\bar{\theta}}^{(t)})) + \sum_{k=1}^K p_k\gamma_k^{(t)}(H_k(\bm{\bar{\theta}}^{(t)})-H^*)\bigg)\bigg\}\\
    &\leq-\mathbb{E}\bigg\{\bigg(\sum_{k=1}^K p_k\gamma_k^{(t)}\langle\nabla H_k(\bm{\bar{\theta}}^{(t)}),\bm{\theta}_k^{(t)} -\bm{\bar{\theta}}^{(t)}\rangle + \sum_{k=1}^K p_k\gamma_k^{(t)}(H_k(\bm{\bar{\theta}}^{(t)})-H^*) \bigg)\bigg\}\\
    &\leq \mathbb{E}\bigg\{\sum_{k=1}^K \frac{1}{2}\gamma_k^{(t)}p_k\bigg[\eta^{(t)}\norm{\nabla H_k(\bm{\bar{\theta}}^{(t)})}^2 + \frac{1}{\eta^{(t)}}\norm{\bm{\theta}_k^{(t)} -\bm{\bar{\theta}}^{(t)}}^2 \bigg] - \sum_{k=1}^K p_k\gamma_k^{(t)}(H_k(\bm{\bar{\theta}}^{(t)})-H^*) \bigg\}\\
    &\leq \mathbb{E}\bigg\{\sum_{k=1}^K \gamma_k^{(t)}p_k\bigg[\eta^{(t)}(1+\frac{\lambda}{p_k|{\cal A}_{s_k}|} r_k(\bm{\theta}))L(H_k(\bm{\bar{\theta}}^{(t)})-H_k^*) +\frac{1}{2\eta^{(t)}}\norm{\bm{\theta}_k^{(t)} - \bm{\bar{\theta}}^{(t)}}^2 \bigg] \\
    &\qquad
    -\sum_{k=1}^K p_k\gamma_k^{(t)}(H_k(\bm{\bar{\theta}}^{(t)})-H^*) \bigg\}.
\end{align*}
In the second inequality, we again use the Cauchy–Schwarz inequality and Inequality of arithmetic and geometric means. In the last inequality, we use the fact that $\norm{\nabla H_k(\bm{\theta}_k^{(t)})}^2\leq 2(1+\frac{\lambda}{p_k|{\cal A}_{s_k}|} r_k(\bm{\theta}))L(H_k(\bm{\theta}_k^{(t)})-H_k^*)$.

Since $\eta^{(t)}\leq\gamma_k^{(t)}\leq 2\eta^{(t)}$, we can bound E as
\begin{align*}
    &\text{E} \leq \text{F} + \mathbb{E}\bigg\{2\eta^{(t)}\Gamma_{max}\bigg\}\\
    &= (\eta^{(t)}(1+\frac{\lambda}{p_k|{\cal A}_{s_k}|} r_k(\bm{\theta}))L-1)\mathbb{E}\bigg\{\sum_{k=1}^K \gamma_k^{(t)}p_k\bigg[(H_k(\bm{\bar{\theta}}^{(t)})-H^*)  \bigg]\bigg\} \\
    &\qquad +\mathbb{E}\bigg\{\sum_{k=1}^K \eta^{(t)}\gamma_k^{(t)}p_k(1+\frac{\lambda}{p_k|{\cal A}_{s_k}|} r_k(\bm{\theta}))L(H^*-H^*_k)\bigg\}\\
    &\qquad +\frac{1}{2\eta^{(t)}}\sum_{k=1}^K  \gamma_k^{(t)}p_k\bigg\{\norm{\bm{\theta}_k^{(t)} -\bm{\bar{\theta}}^{(t)}}^2\bigg\} + 2\eta^{(t)}\Gamma_{max}\\
    &\leq  \mathbb{E}\bigg\{\sum_{k=1}^K \eta^{(t)}\gamma_k^{(t)}p_k(1+\frac{\lambda}{p_k|{\cal A}_{s_k}|} r_k(\bm{\theta}))L(H^*-H^*_k)\bigg\}+\frac{ 1}{2\eta^{(t)}}\sum_{k=1}^K \gamma_k^{(t)}p_k\mathbb{E}\bigg\{\norm{\bm{\theta}_k^{(t)} -\bm{\bar{\theta}}^{(t)}}^2\bigg\} +2\eta^{(t)}\Gamma_{max}\\
    &\leq \sum_{k=1}^K p_k\mathbb{E}\bigg\{\norm{\bm{\theta}_k^{(t)} -\bm{\bar{\theta}}^{(t)}}^2\bigg\} + \mathbb{E}\bigg\{\sum_{k=1}^K \eta^{(t)}\gamma_k^{(t)}p_k(1+\frac{d-1}{p_k|{\cal A}_{s_k}|}\lambda)L(H^*-H^*_k)\bigg\}+2\eta^{(t)}\Gamma_{max}\\
    &\leq  \sum_{k=1}^K p_k\mathbb{E}\bigg\{\norm{\bm{\theta}_k^{(t)} -\bm{\bar{\theta}}^{(t)}}^2\bigg\} + 4\eta^{(t)2}L\mathbb{E}\bigg\{\sum_{k=1}^K p_k(H^*-H^*_k)\bigg\}+2\eta^{(t)}\Gamma_{max}\\
    &= \sum_{k=1}^K p_k\mathbb{E}\bigg\{\norm{\bm{\theta}_k^{(t)} -\bm{\bar{\theta}}^{(t)}}^2\bigg\} + 4\eta^{(t)2}L\Gamma_K+2\eta^{(t)}\Gamma_{max}
\end{align*}
The second inequality holds because $(\eta^{(t)}(1+\frac{\lambda}{p_k|{\cal A}_{s_k}|} r_k(\bm{\theta}))L-1)\leq 0$ and the fourth inequality uses the fact that $1+\frac{d-1}{p_k|{\cal A}_{s_k}|}\lambda\leq 2$ based on the constraint of $\lambda$.

Therefore,
\begin{align*}
    A&\leq \mathbb{E}\bigg\{\norm{\bm{\bar{\theta}}^{(t)}-\bm{\theta}^*}^2\bigg\} -\eta^{(t)}\mathbb{E}\bigg\{\sum_{k=1}^Kp_k(1+\frac{\lambda}{p_k|{\cal A}_{s_k}|} r_k(\bm{\theta}))\mu\norm{\bm{\theta}_k^{(t)}-\bm{\theta}^*}^2\bigg\} + \sum_{k=1}^Kp_k\norm{\bm{\bar{\theta}}^{(t)}-\bm{\theta}_k^{(t)}}^2+\text{E}\\
    &\leq 2\sum_{k=1}^Kp_k\mathbb{E}\bigg\{\norm{\bm{\bar{\theta}}^{(t)}-\bm{\theta}_k^{(t)}}^2\bigg\} + 4\eta^{(t)2}L\Gamma_K+2\eta^{(t)}\Gamma_{max} + \mathbb{E}\bigg\{\norm{\bm{\bar{\theta}}^{(t)}-\bm{\theta}^*}^2\bigg\} \\
    &\qquad -\eta^{(t)}\mathbb{E}\bigg\{\sum_{k=1}^Kp_k(1-\frac{d-1}{p_k|{\cal A}_{s_k}|}\lambda)\mu\norm{\bm{\theta}_k^{(t)}-\bm{\theta}^*}^2\bigg\} \\
    &\leq 2\sum_{k=1}^Kp_k\mathbb{E}\bigg\{\norm{\bm{\bar{\theta}}^{(t)}-\bm{\theta}_k^{(t)}}^2\bigg\} + 4\eta^{(t)2}L\Gamma_K+2\eta^{(t)}\Gamma_{max} + \mathbb{E}\bigg\{\norm{\bm{\bar{\theta}}^{(t)}-\bm{\theta}^*}^2\bigg\} \\
    &\qquad -\eta^{(t)}\mathbb{E}\bigg\{\sum_{k=1}^Kp^2_k(1-\frac{d-1}{p_k|{\cal A}_{s_k}|}\lambda)\mu\norm{\bm{\theta}_k^{(t)}-\bm{\theta}^*}^2\bigg\} \\
    &\leq 2\sum_{k=1}^Kp_k\mathbb{E}\bigg\{\norm{\bm{\bar{\theta}}^{(t)}-\bm{\theta}_k^{(t)}}^2\bigg\} + 4\eta^{(t)2}L\Gamma_K+2\eta^{(t)}\Gamma_{max} + \mathbb{E}\bigg\{\norm{\bm{\bar{\theta}}^{(t)}-\bm{\theta}^*}^2\bigg\} \\
    &\qquad -\eta^{(t)}\mathbb{E}\bigg\{(1-\frac{d-1}{\min\{p_k|{\cal A}_{s_k}|\}}\lambda)\mu\frac{1}{K}\norm{\sum_{k=1}^Kp_k\bm{\theta}_k^{(t)}-\bm{\theta}^*}^2\bigg\} \\
    &= 2\sum_{k=1}^Kp_k\mathbb{E}\bigg\{\norm{\bm{\bar{\theta}}^{(t)}-\bm{\theta}_k^{(t)}}^2\bigg\} +4\eta^{(t)2}L\Gamma_K+2\eta^{(t)}\Gamma_{max} + (1-\eta^{(t)}(1-\frac{d-1}{\min\{p_k|{\cal A}_{s_k}|\}}\lambda)\frac{\mu}{K}) \mathbb{E}\bigg\{\norm{\bm{\bar{\theta}}^{(t)}-\bm{\theta}^*}^2\bigg\}
\end{align*}
The third inequality uses the fact that $0\leq p_k\leq 1$ and $-p_k^2\geq -p_k$. The last inequality uses the fact that $\norm{\sum_{k=1}^Kp_k\bm{\theta}_k}^2\leq K \sum_{k=1}^K\norm{p_k\bm{\theta}_k}^2=K\sum_{k=1}^Kp^2_k\norm{\bm{\theta}_k}^2$ and $1-\frac{d-1}{p_k|{\cal A}_{s_k}|}\lambda\geq1-\frac{d-1}{\min\{p_k|{\cal A}_{s_k}|\}}\lambda$.

\paragraph{Part II: Bounding Term $\sum_{k=1}^Kp_k\mathbb{E}\bigg\{\norm{\bm{\bar{\theta}}^{(t)}-\bm{\theta}_k^{(t)}}^2\bigg\}$ in Term \text{A}} For any iteration $t\geq 0$, denote by $t_0\leq t$ the index of previous communication iteration before $t$. Since the FL algorithm requires one communication each $E$ steps, we know $t-t_0\leq E-1$ and $\bm{\theta}_k^{(t_0)}=\bm{\bar{\theta}}^{(t_0)}$. Assume $\eta^{(t)}\leq 2\eta^{(t+E)}$. Since $\eta^{(t)}$ is decreasing, we have
\begin{align*}
    \mathbb{E}\bigg\{\sum_{k=1}^Kp_k\norm{\bm{\bar{\theta}}^{(t)}-\bm{\theta}_k^{(t)}}^2\bigg\}&=\mathbb{E}\bigg\{\sum_{k=1}^Kp_k\norm{(\bm{\theta}_k^{(t)}- \bm{\bar{\theta}}^{(t_0)}) -
    (\bm{\bar{\theta}}^{(t)}-\bm{\bar{\theta}}^{(t_0)})}^2\bigg\}\\
    &\leq \mathbb{E}\bigg\{\sum_{k=1}^Kp_k\norm{\bm{\theta}_k^{(t)}- \bm{\bar{\theta}}^{(t_0)}}^2\bigg\}\\
    &= \mathbb{E}\bigg\{\sum_{k=1}^Kp_k\norm{\sum_{t=0}^{t-1}\eta^{(t)}g_k(\bm{\theta}_k^{(t)};\zeta^{(t)}_k)}^2\bigg\}\\
    &\leq\mathbb{E}\bigg\{\sum_{k=1}^Kp_k (t-t_0)\sum_{t=0}^{t-1}\eta^{(t)2}\norm{g_k(\bm{\theta}_k^{(t)};\zeta^{(t)}_k)}^2\bigg\}\\
    &\leq\sum_{k=1}^Kp_k\sum_{t=t_0}^{t-1}(E-1)\eta^{(t)2}G^2 \leq \sum_{k=1}^Kp_k\sum_{t=t_0}^{t-1}(E-1)\eta^{(t_0)2}G^2 \\
    &\leq \sum_{k=1}^Kp_k(E-1)^2\eta^{(t_0)2}G^2 \leq 4\eta^{(t)2}(E-1)^2G^2.
\end{align*}

\paragraph{Part III: Bounding Term B} 
By assumption, it is easy to show
\begin{align*}
    \mathbb{E}\bigg\{\eta^{(t)2}\norm{\bm{g}^{(t)}-\bm{\bar{g}}^{(t)}}^2\bigg\}\leq \eta^{(t)2}\sum_{k=1}^Kp_k^2(1+\frac{\lambda}{p_k|{\cal A}_{s_k}|} r_k(\bm{\theta}))^2\sigma_k^2.
\end{align*}

\paragraph{Part IV: Proving Convergence} 
So far, we have shown that
\begin{align*}
    &\mathbb{E}\bigg\{\norm{\bm{\bar{\theta}}^{(t+1)}-\bm{\theta}^*}^2\bigg\}\leq \text{A} + \text{B}\\
    &\leq 8\eta^{(t)2}(E-1)^2G^2 +4\eta^{(t)2}L\Gamma_K+2\eta^{(t)}\Gamma_{max} + (1-\eta^{(t)}(1-\frac{d-1}{p_k|{\cal A}_{s_k}|}\lambda)\mu) \mathbb{E}\bigg\{\norm{\bm{\bar{\theta}}^{(t)}-\bm{\theta}^*}^2\bigg\}  \\
    &\qquad +\eta^{(t)2}\sum_{k=1}^Kp_k^2(1+\frac{\lambda}{p_k|{\cal A}_{s_k}|} r_k(\bm{\theta}))^2\sigma_k^2\\
    &=  (1-\eta^{(t)}(1-\frac{d-1}{\min\{p_k|{\cal A}_{s_k}|\}}\lambda)\frac{\mu}{K}) \mathbb{E}\bigg\{\norm{\bm{\bar{\theta}}^{(t)}-\bm{\theta}^*}^2\bigg\} + \eta^{(t)^2}\xi
\end{align*}
where $\xi=8(E-1)^2G^2+4L\Gamma_K+2\frac{\Gamma_{max}}{\eta^{(t)}}+\sum_{k=1}^Kp_k^2(1+\frac{\lambda}{p_k|{\cal A}_{s_k}|} r_k(\bm{\theta}))^2\sigma_k^2$.

Let $\eta^{(t)}=\frac{\beta}{t+\gamma}$ with $\beta>\frac{1}{(1-\frac{d-1}{min\{p_k|{\cal A}_{s_k}|\}}\lambda)\frac{\mu}{K}}$ and $\gamma>0$. Define $\epsilon\coloneqq(1-\frac{d-1}{min\{p_k|{\cal A}_{s_k}|\}}\lambda)$. Let $v=\max\{\frac{\beta^2\xi}{\beta\epsilon\mu-1},(\gamma+1)\norm{\bm{\bar{\theta}}^{(0)}-\bm{\theta}^*}^2\}$. We will show that $\norm{\bm{\bar{\theta}}^{(t)}-\bm{\theta}^*}^2\leq\frac{v}{\gamma+t}$ by induction. For $t=0$, we have $\norm{\bm{\bar{\theta}}^{(0)}-\bm{\theta}^*}^2\leq(\gamma+1)\norm{\bm{\bar{\theta}}^{(0)}-\bm{\theta}^*}^2\leq\frac{v}{\gamma+1}$. Now assume this is true for some $t$, then
\begin{align*}
    \mathbb{E}\bigg\{\norm{\bm{\bar{\theta}}^{(t+1)}-\bm{\theta}^*}^2\bigg\}&\leq (1-\eta^{(t)}\epsilon\mu)\mathbb{E}\bigg\{\norm{\bm{\bar{\theta}}^{(t)}-\bm{\theta}^*}^2\bigg\} + \eta^{(t)^2}\xi\\
    &\leq(1-\frac{\beta\epsilon\mu}{t+\gamma})\frac{v}{t+\gamma} + \frac{\beta^2\xi}{(t+\gamma)^2}\\
    &=\frac{t+\gamma-1}{(t+\gamma)^2}v + \frac{\beta^2\xi}{(t+\gamma)^2}-\frac{\beta\epsilon\mu-1}{(t+\gamma)^2}v.
\end{align*}
It is easy to show $\frac{t+\gamma-1}{(t+\gamma)^2}v + \frac{\beta^2\xi}{(t+\gamma)^2}-\frac{\beta\epsilon\mu-1}{(t+\gamma)^2}v\leq\frac{v}{t+\gamma+1}$ by definition of $v$. Therefore, we proved $\norm{\bm{\bar{\theta}}^{(t)}-\bm{\theta}^*}^2\leq\frac{v}{\gamma+t}$.

By definition, we know $H$ is $\sum_{k=1}^K p_k\frac{(1+\frac{\lambda}{p_k|{\cal A}_{s_k}|} r_k(\bm{\theta}))}{2}L$-smooth. Therefore, 
\begin{align*}
    \mathbb{E}\bigg\{H(\bm{\bar{\theta}}^{(t)})\bigg\} - H^* &\leq \frac{\sum_{k=1}^K p_k\frac{(1+\frac{\lambda}{p_k|{\cal A}_{s_k}|} r_k(\bm{\theta}))}{2}L}{2}\mathbb{E}\bigg\{\norm{\bm{\bar{\theta}}^{(t)}-\bm{\theta}^*}^2\bigg\}\\
    &\qquad \leq \frac{\sum_{k=1}^K p_k\frac{(1+\frac{\lambda}{p_k|{\cal A}_{s_k}|} r_k(\bm{\theta}))}{2}L}{2}\frac{v}{\gamma+t}.
\end{align*}

By choosing $\beta=\frac{2}{\epsilon\frac{\mu}{K}}$
We have
\begin{align*}
    v=\max\{\frac{\beta^2\xi}{\beta\epsilon\mu-1},(\gamma+1)\norm{\bm{\bar{\theta}}^{(0)}-\bm{\theta}^*}^2\}\leq \frac{\beta^2\xi}{\beta\epsilon\mu-1}+(\gamma+1)\norm{\bm{\bar{\theta}}^{(0)}-\bm{\theta}^*}^2\leq \frac{4\xi}{\epsilon^2\mu^2}+(\gamma+1)\norm{\bm{\bar{\theta}}^{(0)}-\bm{\theta}^*}^2.
\end{align*}
Therefore, 
\begin{align*}
    \mathbb{E}\bigg\{H(\bm{\bar{\theta}}^{(T)})\bigg\} - H^* &\leq\frac{\sum_{k=1}^K p_k\frac{(1+\frac{\lambda}{p_k|{\cal A}_{s_k}|} r_k(\bm{\theta}))}{2}L}{2}\frac{1}{\gamma+T}\bigg\{ \frac{4\xi}{\epsilon^2\mu^2}+(\gamma+1)\norm{\bm{\bar{\theta}}^{(0)}-\bm{\theta}^*}^2\bigg\}\\
    &\leq\frac{\sum_{k=1}^K p_k\frac{(1+\frac{\lambda(d-1)}{p_k|{\cal A}_{s_k}|})}{2}L}{2}\frac{1}{\gamma+T}\bigg\{ \frac{4\xi}{\epsilon^2\mu^2}+(\gamma+1)\norm{\bm{\bar{\theta}}^{(0)}-\bm{\theta}^*}^2\bigg\}\\
    &\leq \frac{L}{2}\frac{1}{\gamma+T}\bigg\{ \frac{4\xi}{\epsilon^2\mu^2}+(\gamma+1)\norm{\bm{\bar{\theta}}^{(0)}-\bm{\theta}^*}^2\bigg\}.
\end{align*}
We thus proved our convergence result.

\begin{theorem}
\label{theorem:conv_partial_1}
Assume at each communication round, central server sampled a fraction $\alpha$ of devices and those local devices are sampled according to the sampling probability $p_k$. Additionally, assume Assumptions in the main paper hold. For $\gamma,\mu,\epsilon>0$, we have
\begin{align*}
    &\mathbb{E}\bigg\{H(\bm{\bar{\theta}}^{(T)})\bigg\} - H^*\leq\frac{L}{2}\frac{1}{\gamma+T}\bigg\{ \frac{4(\xi+\tau)}{\epsilon^2\mu^2}+(\gamma+1)\norm{\bm{\bar{\theta}}^{(0)}-\bm{\theta}^*}^2\bigg\},
\end{align*}
$\tau=\frac{E^2}{\ceil*{\alpha K}}\sum_{k=1}^Kp_k(1+\frac{\lambda}{p_k|{\cal A}_{s_k}|} r_k(\bm{\theta}))^2G^2$.
\end{theorem}

\textbf{Proof} \begin{align*}
    \mathbb{E}\bigg\{\norm{\bm{\bar{\theta}}^{(t+1)}-\bm{\theta}^*}^2\bigg\}&=\mathbb{E}\bigg\{\norm{\bm{\bar{\theta}}^{(t+1)}-\bar{\bm{w}}^{(t+1)}+\bar{\bm{w}}^{(t+1)}-\bm{\theta}^*}^2\bigg\}\\
    &=\mathbb{E}\bigg\{\norm{\bm{\bar{\theta}}^{(t+1)}-\bar{\bm{w}}^{(t+1)}}^2+\norm{\bar{\bm{w}}^{(t+1)}-\bm{\theta}^*}^2+2\langle\bm{\bar{\theta}}^{(t+1)}-\bar{\bm{w}}^{(t+1)},\bar{\bm{w}}^{(t+1)}-\bm{\theta}^* \rangle\bigg\}\\
    &=\mathbb{E}\bigg\{\norm{\bm{\bar{\theta}}^{(t+1)}-\bar{\bm{w}}^{(t+1)}}^2+\norm{\bar{\bm{w}}^{(t+1)}-\bm{\theta}^*}^2\bigg\}.
\end{align*}
Note that the expectation is taken over subset $\mathcal{S}_c$.
\paragraph{Part I: Bounding Term $\mathbb{E}\bigg\{\norm{\bm{\bar{\theta}}^{(t+1)}-\bar{\bm{w}}^{(t+1)}}^2\bigg\}$}

Assume $\ceil*{\alpha K}$ number of local devices are sampled according to sampling probability $p_k$. During the communication round, we have $\bar{\bm{\theta}}^{t+1}=\frac{1}{\ceil*{\alpha K}}\sum_{l=1}^{\ceil*{\alpha K}}\bm{w}^{(t+1)}_{l}$. Therefore,
\begin{align*}
    \mathbb{E}\bigg\{\norm{\bm{\bar{\theta}}^{(t+1)}-\bar{\bm{w}}^{(t+1)}}^2\bigg\}&=\mathbb{E}\bigg\{\frac{1}{\ceil*{\alpha K}^2}\norm{\sum_{l=1}^{\ceil*{\alpha K}}\bm{w}^{(t+1)}_{l}-\bar{\bm{w}}^{(t+1)}}^2\bigg\}\\
    &=\mathbb{E}\bigg\{\frac{1}{\ceil*{\alpha K}^2}\sum_{l=1}^{\ceil*{\alpha K}}\norm{\bm{w}^{(t+1)}_{l}-\bar{\bm{w}}^{(t+1)}}^2\bigg\}\\
    &=\frac{1}{\ceil*{\alpha K}}\sum_{k=1}^Kp_k\norm{\bm{w}^{(t+1)}_{k}-\bar{\bm{w}}^{(t+1)}}^2.
\end{align*}
We know
\begin{align*}
    \sum_{k=1}^Kp_k\norm{\bm{w}^{(t+1)}_{k}-\bar{\bm{w}}^{(t+1)}}^2=\sum_{k=1}^Kp_k\norm{(\bm{w}^{(t+1)}_{k}-\bar{\bm{\theta}}^{(t_0)})-(\bar{\bm{w}}^{(t+1)}-\bar{\bm{\theta}}^{(t_0)})}^2\leq\sum_{k=1}^Kp_k\norm{(\bm{w}^{(t+1)}_{k}-\bar{\bm{\theta}}^{(t_0)})}^2,
\end{align*}
where $t_0=t-E+1$. Similarly,
\begin{align*}
    \mathbb{E}\bigg\{\norm{\bm{\bar{\theta}}^{(t+1)}-\bar{\bm{w}}^{(t+1)}}^2\bigg\}&\leq\frac{1}{\ceil*{\alpha K}}\mathbb{E}\bigg\{\sum_{k=1}^Kp_k\norm{(\bm{w}^{(t+1)}_{k}-\bar{\bm{\theta}}^{(t_0)})}^2\bigg\}\\
    &\leq\frac{1}{\ceil*{\alpha K}}\mathbb{E}\bigg\{\sum_{k=1}^Kp_k\norm{(\bm{w}^{(t+1)}_{k}-\bm{\theta}_k^{(t_0)})}^2\bigg\}\\
    &\leq\frac{1}{\ceil*{\alpha K}}\mathbb{E}\bigg\{\sum_{k=1}^Kp_kE\sum_{m=t_o}^t\norm{\eta^{(m)}\nabla H_k(\bm{\theta}_k^{(m)};\zeta^{(t)}_k)}^2 \bigg\}\\
    &\leq \frac{E^2\eta^{(t_0)2}}{\ceil*{\alpha K}}\sum_{k=1}^Kp_k(1+\frac{\lambda}{p_k|{\cal A}_{s_k}|} r_k(\bm{\theta}))^2G^2\\
    &\leq  \frac{E^2\eta^{(t)2}}{\ceil*{\alpha K}}\sum_{k=1}^Kp_k(1+\frac{\lambda}{p_k|{\cal A}_{s_k}|} r_k(\bm{\theta}))^2G^2
\end{align*}
using the fact that $\eta^{(t)}$ is non-increasing in $t$.

\paragraph{Part II: Convergence Result}
As aforementioned,
\begin{align*}
    \mathbb{E}\bigg\{\norm{\bm{\bar{\theta}}^{(t+1)}-\bm{\theta}^*}^2\bigg\}&=\mathbb{E}\bigg\{\norm{\bm{\bar{\theta}}^{(t+1)}-\bar{\bm{w}}^{(t+1)}}^2+\norm{\bar{\bm{w}}^{(t+1)}-\bm{\theta}^*}^2\bigg\}\\
    &\leq \frac{E^2\eta^{(t)2}}{\ceil*{\alpha K}}\sum_{k=1}^Kp_k(1+\frac{\lambda}{p_k|{\cal A}_{s_k}|} r_k(\bm{\theta}))^2G^2 + (1-\eta^{(t)}\epsilon\frac{\mu}{K})\mathbb{E}\bigg\{\norm{\bm{\bar{\theta}}^{(t)}-\bm{\theta}^*}^2\bigg\} + \eta^{(t)^2}\xi\\
    &=(1-\eta^{(t)}\epsilon\frac{\mu}{K})\mathbb{E}\bigg\{\norm{\bm{\bar{\theta}}^{(t)}-\bm{\theta}^*}^2\bigg\}+\eta^{(t)^2}\bigg(\xi+\frac{E^2}{\ceil*{\alpha K}}\sum_{k=1}^Kp_k(1+\frac{\lambda}{p_k|{\cal A}_{s_k}|} r_k(\bm{\theta}))^2G^2\bigg).
\end{align*}
Let $\tau=\frac{E^2}{\ceil*{\alpha K}}\sum_{k=1}^Kp_k(1+\frac{\lambda}{p_k|{\cal A}_{s_k}|} r_k(\bm{\theta}))^2G^2$. Let $\eta^{(t)}=\frac{\beta}{t+\gamma}$ with $\beta>\frac{1}{\epsilon\frac{\mu}{K}}$ and $\gamma>0$. Let $v=\max\{\frac{\beta^2(\xi+\tau)}{\beta\epsilon\mu-1},(\gamma+1)\norm{\bm{\bar{\theta}}^{(0)}-\bm{\theta}^*}^2\}$. Similar to the full device participation scenario, we can show that $\mathbb{E}\bigg\{\norm{\bm{\bar{\theta}}^{(t)}-\bm{\theta}^*}^2\bigg\}\leq\frac{v}{\gamma+t}$ by induction.

By definition, we know $H$ is $\sum_{k=1}^K p_k\frac{(1+\frac{\lambda}{p_k|{\cal A}_{s_k}|} r_k(\bm{\theta}))}{2}L$-smooth. Therefore, 
\begin{align*}
    \mathbb{E}\bigg\{H(\bm{\bar{\theta}}^{(t)})\bigg\} - H^* &\leq \frac{\sum_{k=1}^K p_k\frac{(1+\frac{\lambda}{p_k|{\cal A}_{s_k}|} r_k(\bm{\theta}))}{2}L}{2}\mathbb{E}\bigg\{\norm{\bm{\bar{\theta}}^{(t)}-\bm{\theta}^*}^2\bigg\}\\
    &\leq \frac{\sum_{k=1}^K p_k\frac{(1+\frac{\lambda}{p_k|{\cal A}_{s_k}|} r_k(\bm{\theta}))}{2}L}{2}\frac{v}{\gamma+t}.
\end{align*}

By choosing $\beta=\frac{2}{\epsilon\frac{\mu}{K}}$
We have
\begin{align*}
    v=\max\{\frac{\beta^2\xi}{\beta\epsilon\mu-1},(\gamma+1)\norm{\bm{\bar{\theta}}^{(0)}-\bm{\theta}^*}^2\}\leq \frac{\beta^2\xi}{\beta\epsilon\mu-1}+(\gamma+1)\norm{\bm{\bar{\theta}}^{(0)}-\bm{\theta}^*}^2\leq \frac{4\xi}{\epsilon^2\mu^2}+(\gamma+1)\norm{\bm{\bar{\theta}}^{(0)}-\bm{\theta}^*}^2.
\end{align*}
Therefore, 
\begin{align*}
    \mathbb{E}\bigg\{H(\bm{\bar{\theta}}^{(T)})\bigg\} - H^*&\leq\frac{\sum_{k=1}^K p_k\frac{(1+\frac{\lambda}{p_k|{\cal A}_{s_k}|} r_k(\bm{\theta}))}{2}L}{2}\frac{1}{\gamma+T}\bigg\{ \frac{4(\xi+\tau)}{\epsilon^2\mu^2}+(\gamma+1)\norm{\bm{\bar{\theta}}^{(0)}-\bm{\theta}^*}^2\bigg\}\\
    &\leq \frac{L}{2}\frac{1}{\gamma+T}\bigg\{ \frac{4(\xi+\tau)}{\epsilon^2\mu^2}+(\gamma+1)\norm{\bm{\bar{\theta}}^{(0)}-\bm{\theta}^*}^2\bigg\}
\end{align*}

\subsection{Convergence (Non-convex)}
\label{app:nonconvex}

\begin{lemma}
\label{lemma:1}
If $\eta^{(t)}\leq\frac{2}{L}$, then $\mathbb{E}\bigg\{H(\bm{\bar{\theta}}^{(t)})\bigg\}\leq \mathbb{E}\bigg\{H(\bm{\bar{\theta}}^{(0)})\bigg\}$.
\end{lemma}
\textbf{Proof}
\begin{align*}
    \mathbb{E}\bigg\{H(\bm{\bar{\theta}}^{(t+1)})\bigg\}&=\mathbb{E}\bigg\{H(\bm{\bar{\theta}}^{(t)}-\eta^{(t)}\sum_{k=1}^Kp_k\nabla H_k(\bm{\theta}^{(t)}_k;\zeta^{(t)}_k))\bigg\}\\
    &=\mathbb{E}\bigg\{H(\bm{\bar{\theta}}^{(t)}-\eta^{(t)}\sum_{k=1}^Kp_k\nabla H_k(\bm{\bar{\theta}}^{(t)};\zeta^{(t)}_k))\bigg\}\\
    &=\mathbb{E}\bigg\{H(\bm{\bar{\theta}}^{(t)}-\eta^{(t)}g^{(t)}(\bm{\bar{\theta}}^{(t)}) )\bigg\}
\end{align*}
Here we used the fact that $\bm{\bar{\theta}}^{(t)}=\bm{\theta}^{(t)}_k$ since the aggregated model parameter has been distributed to local devices. By Taylor's theorem, there exists a $\bm{w}^{(t)}$ such that
\begin{align*}
    \mathbb{E}\bigg\{H(\bm{\bar{\theta}}^{(t+1)})\bigg\}&=\mathbb{E}\bigg\{H(\bm{\bar{\theta}}^{(t)}) - \eta^{(t)}g^{(t)}(\bm{\bar{\theta}}^{(t)})^Tg^{(t)}(\bm{\bar{\theta}}^{(t)}) + \frac{1}{2}(\eta^{(t)}g^{(t)}(\bm{\bar{\theta}}^{(t)}))^Tg^{(t)}(\bm{w}^{(t)}) (\eta^{(t)}g^{(t)}(\bm{\bar{\theta}}^{(t)})) \bigg\}\\
    &\leq \mathbb{E}\bigg\{H(\bm{\bar{\theta}}^{(t)}) - \eta^{(t)}g^{(t)}(\bm{\bar{\theta}}^{(t)})^Tg^{(t)}(\bm{\bar{\theta}}^{(t)}) + \eta^{(t)2}\frac{\sum_{k=1}^K p_k\frac{(1+\frac{\lambda}{p_k|{\cal A}_{s_k}|} r_k(\bm{\theta}))}{2}L}{2}\norm{g^{(t)}(\bm{\bar{\theta}}^{(t)}}^2 \bigg\}\\
    &\leq \mathbb{E}\bigg\{H(\bm{\bar{\theta}}^{(t)})\bigg\} - \eta^{(t)}\norm{g^{(t)}(\bm{\bar{\theta}}^{(t)}}^2 + \eta^{(t)2}\frac{L}{2}\norm{g^{(t)}(\bm{\bar{\theta}}^{(t)}}^2 
\end{align*}
since $H$ is $\sum_{k=1}^K p_k\frac{(1+\frac{\lambda}{p_k|{\cal A}_{s_k}|} r_k(\bm{\theta}))}{2}L$-smooth. It can be shown that if $\eta^{(t)}\leq\frac{2}{L}$, we have 
\begin{align*}
    - \eta^{(t)}\norm{g^{(t)}(\bm{\bar{\theta}}^{(t)}}^2 + \eta^{(t)2}\frac{L}{2}\norm{g^{(t)}(\bm{\bar{\theta}}^{(t)}}^2 \leq 0.
\end{align*}
Therefore, By choosing $\eta^{(t)}\leq\frac{2}{L}$, we proved $\mathbb{E}\bigg\{H(\bm{\bar{\theta}}^{(t)})\bigg\}\leq \mathbb{E}\bigg\{H(\bm{\bar{\theta}}^{(0)})\bigg\}$.

\begin{theorem}
\label{theorem:nonconvex_full_1}
Assume Assumptions in the main paper hold and $|\mathcal{S}_c|=K$. If $\eta^{(t)}=\mathcal{O}(\frac{1}{\sqrt{t}})$ and $\eta^{(t)}\leq\mathcal{O}(\frac{1}{L})$, then for $>0$
\begin{align*}
    \min_{t=1,\ldots,T} \mathbb{E}\bigg\{\norm{\nabla H(\bm{\bar{\theta}}^{(t)})}^2 \bigg\} \leq \frac{1}{\sqrt{T} }\bigg\{ 2(1+2KL^2\sum_{t=1}^T\eta^{(t)2})\mathbb{E}\bigg\{H(\bm{\bar{\theta}}^{(0)}) -H^*\bigg\} + 2\sum_{t=1}^T\xi^{(t)}\bigg\},
\end{align*}
where $\xi^{(t)}=2KL^2\eta^{(t)2}\Gamma_K+(8\eta^{(t)3}KL^2(E-1)+8KL\eta^{(t)2}+4(2+4L)KL\eta^{(t)4}(E-1))G^2 + (2L\eta^{(t)2}+8KL\eta^{(t)2})\sum_{k=1}^Kp_k\sigma_k^2$
\end{theorem}

\textbf{Proof} Since $H$ is $\sum_{k=1}^K p_k\frac{(1+\frac{\lambda}{p_k|{\cal A}_{s_k}|} r_k(\bm{\theta}))}{2}L$-smooth, we have
\begin{align*}
    \mathbb{E}\bigg\{H(\bm{\bar{\theta}}^{(t+1)})\bigg\} &\leq \mathbb{E}\bigg\{H(\bm{\bar{\theta}}^{(t)})\bigg\} + \underbrace{\mathbb{E}\bigg\{\langle \nabla H(\bm{\bar{\theta}}^{(t)}),\bm{\bar{\theta}}^{(t+1)}-\bm{\bar{\theta}}^{(t)}\rangle \bigg\}}_{\text{A}} + \\
    &\qquad \frac{\sum_{k=1}^K p_k\frac{(1+\frac{\lambda}{p_k|{\cal A}_{s_k}|} r_k(\bm{\theta}))}{2}L}{2}\underbrace{\mathbb{E}\bigg\{\norm{\bm{\bar{\theta}}^{(t+1)}-\bm{\bar{\theta}}^{(t)}}^2\bigg\}}_{\text{B}}.
\end{align*}

\paragraph{Part I: Bounding Term \text{A}} We have
\begin{align*}
    \text{A}&=-\eta^{(t)}\mathbb{E}\bigg\{\langle \nabla H(\bm{\bar{\theta}}^{(t)}),\sum_{k=1}^Kp_k\nabla H_k(\bm{\theta}_k^{(t)};\zeta^{(t)}_k)  \rangle\bigg\}=-\eta^{(t)}\mathbb{E}\bigg\{\langle \nabla H(\bm{\bar{\theta}}^{(t)}),\sum_{k=1}^Kp_k\nabla H_k(\bm{\theta}_k^{(t)})  \rangle\bigg\}\\
    &=-\frac{1}{2}\eta^{(t)} \mathbb{E}\bigg\{\norm{\nabla H(\bm{\bar{\theta}}^{(t)})}^2 \bigg\} -\frac{1}{2}\eta^{(t)}\mathbb{E}\bigg\{\norm{\sum_{k=1}^Kp_k\nabla H_k(\bm{\theta}_k^{(t)})}^2  \bigg\} + \frac{1}{2}\eta^{(t)}\mathbb{E}\bigg\{ \norm{\nabla H(\bm{\bar{\theta}}^{(t)})-\sum_{k=1}^Kp_k\nabla H_k(\bm{\theta}_k^{(t)})  }^2\bigg\}\\
    &=-\frac{1}{2}\eta^{(t)} \mathbb{E}\bigg\{\norm{\nabla H(\bm{\bar{\theta}}^{(t)})}^2 \bigg\} -\frac{1}{2}\eta^{(t)}\mathbb{E}\bigg\{\norm{\sum_{k=1}^Kp_k\nabla H_k(\bm{\theta}_k^{(t)})}^2  \bigg\} + \frac{1}{2}\eta^{(t)}\mathbb{E}\bigg\{ \norm{\sum_{k=1}^Kp_k\nabla H_k(\bm{\bar{\theta}}^{(t)})-\sum_{k=1}^Kp_k\nabla H_k(\bm{\theta}_k^{(t)})  }^2\bigg\}\\
    &\leq -\frac{1}{2}\eta^{(t)} \mathbb{E}\bigg\{\norm{\nabla H(\bm{\bar{\theta}}^{(t)})}^2 \bigg\} -\frac{1}{2}\eta^{(t)}\mathbb{E}\bigg\{\norm{\sum_{k=1}^Kp_k\nabla H_k(\bm{\theta}_k^{(t)})}^2  \bigg\} + \frac{1}{2}\eta^{(t)}\mathbb{E}\bigg\{ K\sum_{k=1}^Kp_k\norm{\nabla H_k(\bm{\bar{\theta}}^{(t)})-\nabla H_k(\bm{\theta}_k^{(t)})  }^2\bigg\}\\
    &\leq -\frac{1}{2}\eta^{(t)} \mathbb{E}\bigg\{\norm{\nabla H(\bm{\bar{\theta}}^{(t)})}^2 \bigg\} -\frac{1}{2}\eta^{(t)}\mathbb{E}\bigg\{\norm{\sum_{k=1}^Kp_k\nabla H_k(\bm{\theta}_k^{(t)})}^2  \bigg\} +  \\
    &\qquad\frac{1}{2}\eta^{(t)}\mathbb{E}\bigg\{ K\sum_{k=1}^Kp_k((1+\frac{\lambda}{p_k|{\cal A}_{s_k}|} r_k(\bm{\theta}))L)^2\underbrace{\norm{\bm{\bar{\theta}}^{(t)}-\bm{\theta}_k^{(t)}}^2}_{\text{C}} \bigg\}.
\end{align*}
In the convex setting, we proved that
\begin{align*}
    \text{C}\leq 4\eta^{(t)2}(E-1)G^2.
\end{align*}
This is also true for the non-convex setting since we do not use any property of convex functions.

\paragraph{Part II: Bounding Term \text{B}} We have
\begin{align*}
    \text{B}&=\mathbb{E}\bigg\{\norm{ \eta^{(t)}g^{(t)} }^2\bigg\}=\mathbb{E}\bigg\{\norm{ \eta^{(t)}\sum_{k=1}^Kp_k\nabla H_k(\bm{\theta}_k^{(t)};\zeta^{(t)}_k) }^2\bigg\}\\
    &=\mathbb{E}\bigg\{\norm{ \eta^{(t)}\sum_{k=1}^Kp_k(\nabla H_k(\bm{\theta}_k^{(t)};\zeta^{(t)}_k) -\nabla H_k(\bm{\theta}_k^{(t)}) ) }^2\bigg\} + \mathbb{E}\bigg\{\norm{ \eta^{(t)}\sum_{k=1}^Kp_k\nabla H_k(\bm{\theta}_k^{(t)}) }^2\bigg\}\\
    &=\eta^{(t)2}\sum_{k=1}^Kp_k^2\mathbb{E}\bigg\{\norm{\nabla H_k(\bm{\theta}_k^{(t)};\zeta^{(t)}_k)-\nabla H_k(\bm{\theta}_k^{(t)}) }^2 \bigg\} + \mathbb{E}\bigg\{\norm{ \eta^{(t)}\sum_{k=1}^Kp_k\nabla H_k(\bm{\theta}_k^{(t)}) }^2\bigg\}\\
    &\leq \eta^{(t)2}\sum_{k=1}^Kp_k^2(1+\frac{\lambda}{p_k|{\cal A}_{s_k}|} r_k(\bm{\theta}))^2\sigma_k^2+ \eta^{(t)2}\mathbb{E}\bigg\{K\sum_{k=1}^Kp_k^2\norm{ \nabla H_k(\bm{\theta}_k^{(t)}) }^2\bigg\}.
\end{align*}

Since $H_k$ is $(1+\frac{\lambda}{p_k|{\cal A}_{s_k}|} r_k(\bm{\theta}))L$-smooth, we know
\begin{align*}
    \norm{\nabla H_k(\bm{\theta}_k^{(t)})}^2\leq 2(1+\frac{\lambda}{p_k|{\cal A}_{s_k}|} r_k(\bm{\theta}))L(H_k(\bm{\theta}_k^{(t)})-H_k^*).
\end{align*}
Therefore, 
\begin{align*}
    \text{B}&\leq\eta^{(t)2}\sum_{k=1}^Kp_k^2(1+\frac{\lambda}{p_k|{\cal A}_{s_k}|} r_k(\bm{\theta}))^2\sigma_k^2+ \\
    &\qquad \eta^{(t)2}\mathbb{E}\bigg\{K\sum_{k=1}^K2p_k^2 (1+\frac{\lambda}{p_k|{\cal A}_{s_k}|} r_k(\bm{\theta}))L(H_k(\bm{\theta}_k^{(t)})-H_k^*) \bigg\}\\
    &=\eta^{(t)2}\sum_{k=1}^Kp_k^2(1+\frac{\lambda}{p_k|{\cal A}_{s_k}|} r_k(\bm{\theta}))^2\sigma_k^2+ \\
    &\qquad \eta^{(t)2}\mathbb{E}\bigg\{K\sum_{k=1}^K2p_k^2 (1+\frac{\lambda}{p_k|{\cal A}_{s_k}|} r_k(\bm{\theta}))L(H_k(\bm{\theta}_k^{(t)})-H^*+H^*-H_k^*) \bigg\}\\
    &\leq\eta^{(t)2}\sum_{k=1}^Kp_k^2(1+\frac{\lambda}{p_k|{\cal A}_{s_k}|} r_k(\bm{\theta}))^2\sigma_k^2+ \\
    &\qquad \eta^{(t)2}\mathbb{E}\bigg\{K\sum_{k=1}^K2p_k (1+\frac{\lambda}{p_k|{\cal A}_{s_k}|} r_k(\bm{\theta}))L(H_k(\bm{\theta}_k^{(t)})-H^*+H^*-H_k^*) \bigg\}
\end{align*}
since $0\leq p_k\leq 1$ and $p_k^2\leq p_k$.

Therefore,
\begin{align*}
    \mathbb{E}\bigg\{H(\bm{\bar{\theta}}^{(t+1)})\bigg\} &\leq \mathbb{E}\bigg\{H(\bm{\bar{\theta}}^{(t)})\bigg\}-\frac{1}{2}\eta^{(t)} \mathbb{E}\bigg\{\norm{\nabla H(\bm{\bar{\theta}}^{(t)})}^2 \bigg\} \underbrace{-\frac{1}{2}\eta^{(t)}\mathbb{E}\bigg\{\norm{\sum_{k=1}^Kp_k\nabla H_k(\bm{\theta}_k^{(t)})}^2  \bigg\} }_{\text{D}<0} + \\
    &\qquad\frac{1}{2}\eta^{(t)}\mathbb{E}\bigg\{ K\sum_{k=1}^Kp_k((1+\frac{\lambda}{p_k|{\cal A}_{s_k}|} r_k(\bm{\theta}))L)^2 4\eta^{(t)2}(E-1)G^2 \bigg\} \\
    &\qquad + \frac{\sum_{k=1}^K p_k\frac{(1+\frac{\lambda}{p_k|{\cal A}_{s_k}|} r_k(\bm{\theta}))}{2}L}{2} \bigg[  \eta^{(t)2}\sum_{k=1}^Kp_k^2(1+\frac{\lambda}{p_k|{\cal A}_{s_k}|} r_k(\bm{\theta}))^2\sigma_k^2 \\
    &\qquad +\eta^{(t)2}\mathbb{E}\bigg\{K\sum_{k=1}^K2p_k (1+\frac{\lambda}{p_k|{\cal A}_{s_k}|} r_k(\bm{\theta}))L(H_k(\bm{\theta}_k^{(t)})-H^*+H^*-H_k^*) \bigg\} \bigg]\\
    &\leq \mathbb{E}\bigg\{H(\bm{\bar{\theta}}^{(t)})\bigg\}-\frac{1}{2}\eta^{(t)} \mathbb{E}\bigg\{\norm{\nabla H(\bm{\bar{\theta}}^{(t)})}^2 \bigg\}\\
    &\qquad\frac{1}{2}\eta^{(t)}\mathbb{E}\bigg\{ K\sum_{k=1}^Kp_k((1+\frac{\lambda}{p_k|{\cal A}_{s_k}|} r_k(\bm{\theta}))L)^2 4\eta^{(t)2}(E-1)G^2 \bigg\} \\
    &\qquad  +\frac{\sum_{k=1}^K p_k\frac{(1+\frac{\lambda}{p_k|{\cal A}_{s_k}|} r_k(\bm{\theta}))}{2}L}{2} \bigg[  \eta^{(t)2}\sum_{k=1}^Kp_k^2(1+\frac{\lambda}{p_k|{\cal A}_{s_k}|} r_k(\bm{\theta}))^2\sigma_k^2+ \\
    &\qquad \underbrace{4KL\eta^{(t)2}\mathbb{E}\bigg\{\sum_{k=1}^Kp_k(H_k(\bm{\theta}_k^{(t)})-H^*)+\sum_{k=1}^Kp_k(H^*-H_k^*) \bigg\}}_{\text{E}} \bigg] \\
    &\leq \mathbb{E}\bigg\{H(\bm{\bar{\theta}}^{(t)})\bigg\}-\frac{1}{2}\eta^{(t)} \mathbb{E}\bigg\{\norm{\nabla H(\bm{\bar{\theta}}^{(t)})}^2 \bigg\}+\frac{1}{2}\eta^{(t)}\mathbb{E}\bigg\{ K\sum_{k=1}^K4p_kL^2 4\eta^{(t)2}(E-1)G^2 \bigg\} \\
    & \qquad +\frac{L}{2} \bigg[  \eta^{(t)2}\sum_{k=1}^K4p_k^2\sigma_k^2+  \underbrace{4KL\eta^{(t)2}\mathbb{E}\bigg\{\sum_{k=1}^Kp_k(H_k(\bm{\theta}_k^{(t)})-H^*)+\sum_{k=1}^Kp_k(H^*-H_k^*) \bigg\}}_{\text{E}} \bigg]
\end{align*}
Here
\begin{align*}
    \text{E}&=4KL\eta^{(t)2}\mathbb{E}\bigg\{\sum_{k=1}^Kp_k(H_k(\bm{\theta}_k^{(t)})-H^*)\bigg\} + 4KL\eta^{(t)2}\mathbb{E}\bigg\{\sum_{k=1}^Kp_k(H^*-H_k^*) \bigg\}\\
    &=4KL\eta^{(t)2}\mathbb{E}\bigg\{\sum_{k=1}^Kp_k(H_k(\bm{\theta}_k^{(t)})-H_k(\bm{\bar{\theta}}^{(t)}))\bigg\} + 4KL\eta^{(t)2}\mathbb{E}\bigg\{\sum_{k=1}^Kp_k (H_k(\bm{\bar{\theta}}^{(t)}) -H^*) \bigg\}+ 4KL\eta^{(t)2}\Gamma_K\\
    &=4KL\eta^{(t)2}\underbrace{\mathbb{E}\bigg\{\sum_{k=1}^Kp_k(H_k(\bm{\theta}_k^{(t)})-H_k(\bm{\bar{\theta}}^{(t)}))\bigg\}}_{\text{F}} + 4KL\eta^{(t)2}\mathbb{E}\bigg\{H(\bm{\bar{\theta}}^{(t)}) -H^*\bigg\}+ 4KL\eta^{(t)2}\Gamma_K.
\end{align*}
We can bound term F as
\begin{align*}
    \text{F}&=\mathbb{E}\bigg\{\sum_{k=1}^Kp_k(H_k(\bm{\theta}_k^{(t)})-H_k(\bm{\bar{\theta}}^{(t)}))\bigg\}\\
    &\leq\mathbb{E}\bigg\{\sum_{k=1}^Kp_k( \langle \nabla H_k(\bm{\bar{\theta}}^{(t)}),\bm{\theta}_k^{(t)}-\bm{\bar{\theta}}^{(t)}\rangle + \frac{(1+\frac{\lambda}{p_k|{\cal A}_{s_k}|} r_k(\bm{\theta}))L}{2}\underbrace{\norm{\bm{\theta}_k^{(t)}-\bm{\bar{\theta}}^{(t)}}^2}_{\leq4\eta^{(t)2}(E-1)G^2}  )\bigg\}
\end{align*}
where we use the fact that $H_k$ is $(1+\frac{\lambda}{p_k|{\cal A}_{s_k}|} r_k(\bm{\theta}))L$-smooth. To bound the inner product, we again use the inequality of arithmetic and geometric means and Cauchy–Schwarz inequality:
\begin{align*}
     \langle \nabla H_k(\bm{\bar{\theta}}^{(t)}),\bm{\theta}_k^{(t)}-\bm{\bar{\theta}}^{(t)}\rangle \leq \norm{\nabla H_k(\bm{\bar{\theta}}^{(t)})}\norm{\bm{\theta}_k^{(t)}-\bm{\bar{\theta}}^{(t)}}\leq \frac{\norm{\nabla H_k(\bm{\bar{\theta}}^{(t)})}^2+\norm{\bm{\theta}_k^{(t)}-\bm{\bar{\theta}}^{(t)}}^2}{2}.
\end{align*}
It can be shown that
\begin{align*}
    \mathbb{E}\bigg\{\norm{\nabla H_k(\bm{\bar{\theta}}^{(t)})}^2\bigg\} &= \mathbb{E}\bigg\{\norm{\nabla F_k(\bm{\theta}^{(t)}_k,D^{(t)}_k)}\bigg\}^2 +  \mathbb{E}\bigg\{\norm{\nabla F_k(\bm{\theta}^{(t)}_k;\zeta^{(t)}_k)-\nabla F_k(\bm{\theta}^{(t)}_k)}^2\bigg\}\\
    &\leq \mathbb{E}\bigg\{\norm{\nabla F_k(\bm{\theta}^{(t)}_k;\zeta^{(t)}_k)}^2\bigg\} +  \mathbb{E}\bigg\{\norm{\nabla F_k(\bm{\theta}^{(t)}_k;\zeta^{(t)}_k)-\nabla F_k(\bm{\theta}^{(t)}_k)}^2\bigg\}\\
    &\leq (1+\frac{\lambda}{p_k|{\cal A}_{s_k}|} r_k(\bm{\theta}))^2(G^2 +\sigma_k^2)\leq 4(G^2 +\sigma_k^2)
\end{align*}
Therefore, we can simplify F as
\begin{align*}
    \text{F}&\leq\mathbb{E}\bigg\{\sum_{k=1}^Kp_k( \frac{\norm{\nabla H_k(\bm{\bar{\theta}}^{(t)})}^2+\norm{\bm{\theta}_k^{(t)}-\bm{\bar{\theta}}^{(t)}}^2}{2} + \frac{(1+\frac{\lambda}{p_k|{\cal A}_{s_k}|} r_k(\bm{\theta}))L}{2}\underbrace{\norm{\bm{\theta}_k^{(t)}-\bm{\bar{\theta}}^{(t)}}^2}_{\leq4\eta^{(t)2}(E-1)G^2}  )\bigg\}\\
    &\leq \mathbb{E}\bigg\{\sum_{k=1}^Kp_k( \frac{4(G^2 +\sigma_k^2)+4\eta^{(t)2}(E-1)G^2}{2} + 4L\eta^{(t)2}(E-1)G^2 )\bigg\}\\
    &= 2\mathbb{E}\bigg\{\sum_{k=1}^Kp_k\sigma_k^2\bigg\} + 2G^2 + (2+4L)\eta^{(t)2}(E-1)G^2
\end{align*}
Combining with E, we obtain
\begin{align*}
    \text{E}&\leq 4KL\eta^{(t)2}\bigg(2\sum_{k=1}^Kp_k\sigma_k^2+ 2G^2 + (2+4L)\eta^{(t)2}(E-1)G^2\bigg) + 4KL\eta^{(t)2}\mathbb{E}\bigg\{H(\bm{\bar{\theta}}^{(t)}) -H^*\bigg\}+ 4KL\eta^{(t)2}\Gamma_K
\end{align*}

\paragraph{Part III: Proving Convergence}

Therefore,
\begin{align*}
     &\frac{1}{2}\eta^{(t)} \mathbb{E}\bigg\{\norm{\nabla H(\bm{\bar{\theta}}^{(t)})}^2 \bigg\}\\
     &\leq\mathbb{E}\bigg\{H(\bm{\bar{\theta}}^{(t)})\bigg\} - \mathbb{E}\bigg\{H(\bm{\bar{\theta}}^{(t+1)})\bigg\} + \frac{1}{2}\eta^{(t)3}\mathbb{E}\bigg\{ K\sum_{k=1}^K4p_kL^2 4(E-1)G^2 \bigg\} + \\
     &\frac{L}{2} \bigg[  \eta^{(t)2}\sum_{k=1}^K4p_k^2\sigma_k^2+  4KL\eta^{(t)2}\bigg(2\sum_{k=1}^Kp_k\sigma_k^2+ 2G^2 + (2+4L)\eta^{(t)2}(E-1)G^2\bigg) + 4KL\eta^{(t)2}\mathbb{E}\bigg\{H(\bm{\bar{\theta}}^{(t)}) -H^*\bigg\}\\
     &\qquad +4KL\eta^{(t)2}\Gamma_K\bigg]\\
     &=\mathbb{E}\bigg\{H(\bm{\bar{\theta}}^{(t)})\bigg\} - \mathbb{E}\bigg\{H(\bm{\bar{\theta}}^{(t+1)})\bigg\} +  2KL^2\eta^{(t)2}\mathbb{E}\bigg\{H(\bm{\bar{\theta}}^{(t)}) -H^*\bigg\}+ 2KL^2\eta^{(t)2}\Gamma_K\\
     &\qquad +(8\eta^{(t)3}KL^2(E-1)+8KL\eta^{(t)2}+4(2+4L)KL\eta^{(t)4}(E-1))G^2 + (2L\eta^{(t)2}+8KL\eta^{(t)2})\sum_{k=1}^Kp_k\sigma_k^2. 
\end{align*}
Let $\xi^{(t)}=2KL^2\eta^{(t)2}\Gamma_K+(8\eta^{(t)3}KL^2(E-1)+8KL\eta^{(t)2}+4(2+4L)KL\eta^{(t)4}(E-1))G^2 + (2L\eta^{(t)2}+8KL\eta^{(t)2})\sum_{k=1}^Kp_k\sigma_k^2$, then
\begin{align*}
    \frac{1}{2}\eta^{(t)} \mathbb{E}\bigg\{\norm{\nabla H(\bm{\bar{\theta}}^{(t)})}^2 \bigg\}&\leq\mathbb{E}\bigg\{H(\bm{\bar{\theta}}^{(t)})\bigg\} - \mathbb{E}\bigg\{H(\bm{\bar{\theta}}^{(t+1)})\bigg\} +  2KL^2\eta^{(t)2}\mathbb{E}\bigg\{H(\bm{\bar{\theta}}^{(t)}) -H^*\bigg\} + \xi^{(t)}\\
    &\leq \mathbb{E}\bigg\{H(\bm{\bar{\theta}}^{(t)})\bigg\} - \mathbb{E}\bigg\{H(\bm{\bar{\theta}}^{(t+1)})\bigg\} +  2KL^2\eta^{(t)2}\mathbb{E}\bigg\{H(\bm{\bar{\theta}}^{(0)}) -H^*\bigg\} + \xi^{(t)}
\end{align*}
since $\eta^{(t)}\leq\frac{1}{\sqrt{2K}L}$ and $\mathbb{E}\bigg\{H(\bm{\bar{\theta}}^{(t)})\bigg\}\leq \mathbb{E}\bigg\{H(\bm{\bar{\theta}}^{(0)})\bigg\}$ by Lemma \ref{lemma:1}. By taking summation on both side, we obtain
\begin{align*}
    \sum_{t=1}^T\frac{1}{2}\eta^{(t)} \mathbb{E}\bigg\{\norm{\nabla H(\bm{\bar{\theta}}^{(t)})}^2 \bigg\} &\leq  \mathbb{E}\bigg\{H(\bm{\bar{\theta}}^{(0)})\bigg\} - \mathbb{E}\bigg\{H(\bm{\bar{\theta}}^{(t+1)})\bigg\} + 2KL^2\sum_{t=1}^T\eta^{(t)2}\mathbb{E}\bigg\{H(\bm{\bar{\theta}}^{(0)}) -H^*\bigg\} + \sum_{t=1}^T\xi^{(t)}\\
    &\leq  \mathbb{E}\bigg\{H(\bm{\bar{\theta}}^{(0)})\bigg\} - \mathbb{E}\bigg\{H(\bm{\bar{\theta}}^{*})\bigg\} + 2KL^2\sum_{t=1}^T\eta^{(t)2}\mathbb{E}\bigg\{H(\bm{\bar{\theta}}^{(0)}) -H^*\bigg\} + \sum_{t=1}^T\xi^{(t)}\\
    &=(1+2KL^2\sum_{t=1}^T\eta^{(t)2})\mathbb{E}\bigg\{H(\bm{\bar{\theta}}^{(0)}) -H^*\bigg\} + \sum_{t=1}^T\xi^{(t)}.
\end{align*}
This implies
\begin{align*}
    \min_{t=1,\ldots,T} \mathbb{E}\bigg\{\norm{\nabla H(\bm{\bar{\theta}}^{(t)})}^2 \bigg\} \sum_{t=1}^T\eta^{(t)} \leq 2(1+2KL^2\sum_{t=1}^T\eta^{(t)2})\mathbb{E}\bigg\{H(\bm{\bar{\theta}}^{(0)}) -H^*\bigg\} + 2\sum_{t=1}^T\xi^{(t)}
\end{align*}
and therefore
\begin{align*}
    \min_{t=1,\ldots,T} \mathbb{E}\bigg\{\norm{\nabla H(\bm{\bar{\theta}}^{(t)})}^2 \bigg\} \leq \frac{1}{\sum_{t=1}^T\eta^{(t)} }\bigg\{ 2(1+2KL^2\sum_{t=1}^T\eta^{(t)2})\mathbb{E}\bigg\{H(\bm{\bar{\theta}}^{(0)}) -H^*\bigg\} + 2\sum_{t=1}^T\xi^{(t)}\bigg\}.
\end{align*}
Let $\eta^{(t)}=\frac{1}{\sqrt{t}}$, then we have $\sum_{t=1}^T\eta^{(t)}=\mathcal{O}(\sqrt{T})$ and $\sum_{t=1}^T\eta^{(t)2}=\mathcal{O}(\log(T+1))$. Therefore,
\begin{align*}
    \min_{t=1,\ldots,T} \mathbb{E}\bigg\{\norm{\nabla H(\bm{\bar{\theta}}^{(t)})}^2 \bigg\} \leq \frac{1}{\sqrt{T} }\bigg\{ 2(1+2KL^2\sum_{t=1}^T\eta^{(t)2})\mathbb{E}\bigg\{H(\bm{\bar{\theta}}^{(0)}) -H^*\bigg\} + 2\sum_{t=1}^T\xi^{(t)}\bigg\}.
\end{align*}


\section{Additional Experiments}
\begin{figure*}[!htbp]
    \vskip -0.1in
    \centering
    \centerline{\includegraphics[width=0.8\columnwidth]{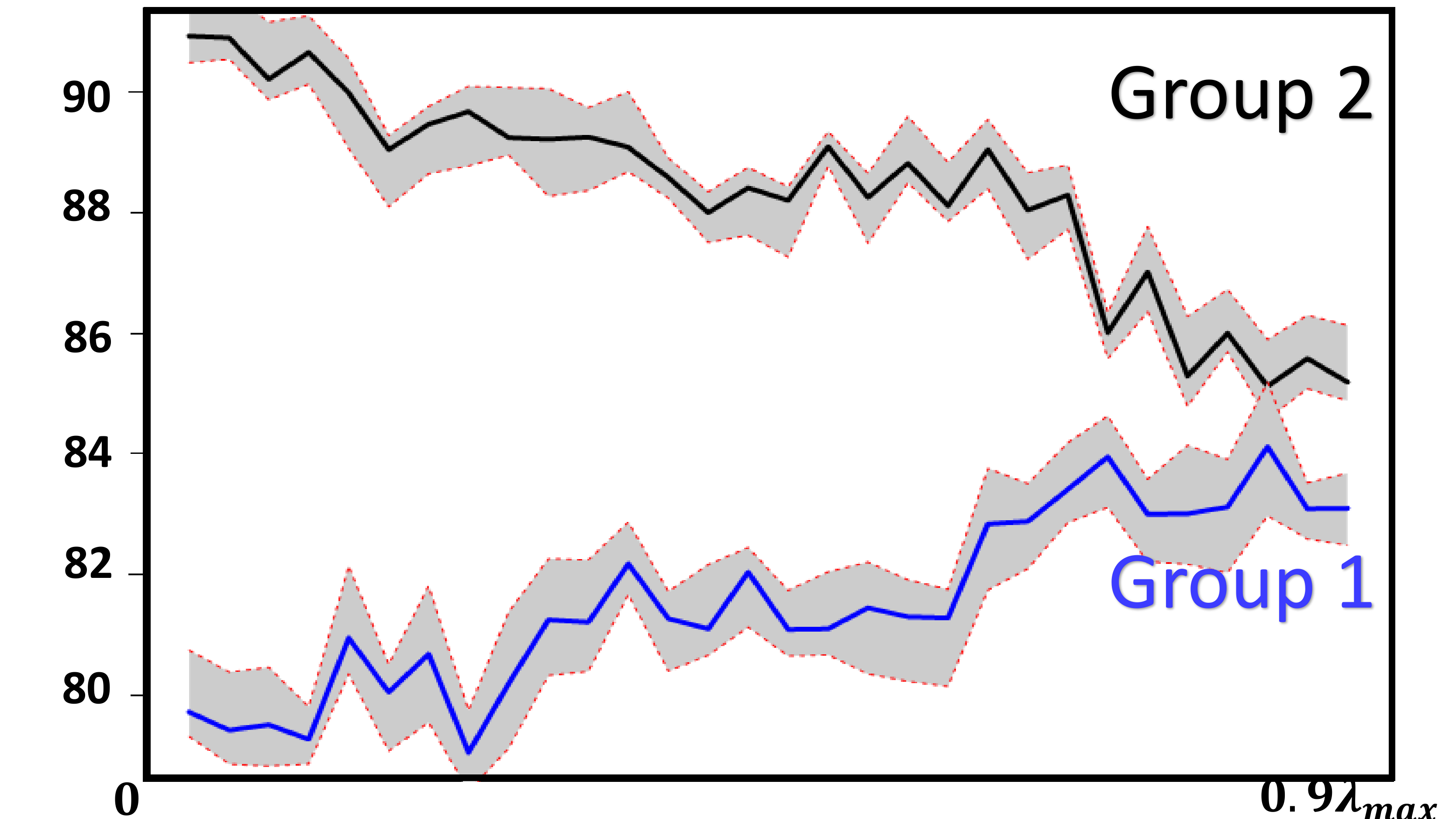}}
    \caption{Sensitivity analysis on FEMNIST}
    \label{fig:femnist1}
    \vskip -0.1in
\end{figure*}

We conduct a sensitivity analysis using the FEMNIST-3-groups setting. Results are reported in Figure \ref{fig:femnist1}. Similar to the observation in the main paper, it can be seen that as $\lambda$ increases, the discrepancy between two groups decreases accordingly. Here kindly note that we did not plot group 3 for the sake of neatness. The line of group should stay in the middle of two lines.

\label{app:exp}

\newpage 

\bibliography{mybib}

\begin{thebibliography}{56}
\providecommand{\natexlab}[1]{#1}
\providecommand{\url}[1]{\texttt{#1}}
\providecommand{\urlprefix}{URL }

\bibitem[{Al-Ali et~al.(2018)Al-Ali, Gupta, \protect\BIBand{}
  Nabulsi}]{al2018cyber}
Al-Ali A, Gupta R, Nabulsi AA (2018) Cyber physical systems role in
  manufacturing technologies. \emph{AIP Conference Proceedings}, volume 1957,
  050007 (AIP Publishing LLC).

\bibitem[{Arivazhagan et~al.(2019)Arivazhagan, Aggarwal, Singh,
  \protect\BIBand{} Choudhary}]{arivazhagan2019federated}
Arivazhagan MG, Aggarwal V, Singh AK, Choudhary S (2019) Federated learning
  with personalization layers. \emph{arXiv preprint arXiv:1912.00818} .

\bibitem[{Bhagoji et~al.(2019)Bhagoji, Chakraborty, Mittal, \protect\BIBand{}
  Calo}]{bhagoji2019analyzing}
Bhagoji AN, Chakraborty S, Mittal P, Calo S (2019) Analyzing federated learning
  through an adversarial lens. \emph{International Conference on Machine
  Learning}, 634--643 (PMLR).

\bibitem[{Caldas et~al.(2018)Caldas, Duddu, Wu, Li, Kone{\v{c}}n{\`y}, McMahan,
  Smith, \protect\BIBand{} Talwalkar}]{caldas2018leaf}
Caldas S, Duddu SMK, Wu P, Li T, Kone{\v{c}}n{\`y} J, McMahan HB, Smith V,
  Talwalkar A (2018) Leaf: A benchmark for federated settings. \emph{arXiv
  preprint arXiv:1812.01097} .

\bibitem[{Chen et~al.(2018)Chen, Luo, Dong, Li, \protect\BIBand{}
  He}]{chen2018federated}
Chen F, Luo M, Dong Z, Li Z, He X (2018) Federated meta-learning with fast
  convergence and efficient communication. \emph{arXiv preprint
  arXiv:1802.07876} .

\bibitem[{CleanTechnica(2021)}]{Tesla}
CleanTechnica (2021) Tesla fsd hardware has 150 million times more computer
  power than apollo 11 computer.
  \url{https://cleantechnica.com/2021/05/24/tesla-fsd-hardware-has-150-million-times-more-computer-power-than-apollo-11-computer/},
  accessed: 2021-05-24.

\bibitem[{Cohen et~al.(2017)Cohen, Afshar, Tapson, \protect\BIBand{}
  Van~Schaik}]{cohen2017emnist}
Cohen G, Afshar S, Tapson J, Van~Schaik A (2017) Emnist: Extending mnist to
  handwritten letters. \emph{2017 International Joint Conference on Neural
  Networks (IJCNN)}, 2921--2926 (IEEE).

\bibitem[{Dai et~al.(2020)Dai, Low, \protect\BIBand{}
  Jaillet}]{dai2020federated}
Dai Z, Low KH, Jaillet P (2020) Federated bayesian optimization via thompson
  sampling. \emph{arXiv preprint arXiv:2010.10154} .

\bibitem[{Dinh et~al.(2020)Dinh, Tran, \protect\BIBand{}
  Nguyen}]{dinh2020personalized}
Dinh CT, Tran NH, Nguyen TD (2020) Personalized federated learning with moreau
  envelopes. \emph{Conference on Neural Information Processing Systems} .

\bibitem[{Du et~al.(2020)Du, Xu, Wu, \protect\BIBand{} Tong}]{du2020fairness}
Du W, Xu D, Wu X, Tong H (2020) Fairness-aware agnostic federated learning.
  \emph{arXiv preprint arXiv:2010.05057} .

\bibitem[{Fallah et~al.(2020)Fallah, Mokhtari, \protect\BIBand{}
  Ozdaglar}]{fallah2020personalized}
Fallah A, Mokhtari A, Ozdaglar A (2020) Personalized federated learning: A
  meta-learning approach. \emph{Advances in Neural Information Processing
  Systems 33} .

\bibitem[{Feldman et~al.(2015)Feldman, Friedler, Moeller, Scheidegger,
  \protect\BIBand{} Venkatasubramanian}]{feldman2015certifying}
Feldman M, Friedler SA, Moeller J, Scheidegger C, Venkatasubramanian S (2015)
  Certifying and removing disparate impact. \emph{proceedings of the 21th ACM
  SIGKDD international conference on knowledge discovery and data mining},
  259--268.

\bibitem[{Ghosh et~al.(2019)Ghosh, Hong, Yin, \protect\BIBand{}
  Ramchandran}]{ghosh2019robust}
Ghosh A, Hong J, Yin D, Ramchandran K (2019) Robust federated learning in a
  heterogeneous environment. \emph{arXiv preprint arXiv:1906.06629} .

\bibitem[{Hard et~al.(2018)Hard, Rao, Mathews, Ramaswamy, Beaufays, Augenstein,
  Eichner, Kiddon, \protect\BIBand{} Ramage}]{hard2018federated}
Hard A, Rao K, Mathews R, Ramaswamy S, Beaufays F, Augenstein S, Eichner H,
  Kiddon C, Ramage D (2018) Federated learning for mobile keyboard prediction.
  \emph{arXiv preprint arXiv:1811.03604} .

\bibitem[{Hardt et~al.(2016)Hardt, Price, \protect\BIBand{}
  Srebro}]{hardt2016equality}
Hardt M, Price E, Srebro N (2016) Equality of opportunity in supervised
  learning. \emph{Conference on Neural Information Processing Systems} .

\bibitem[{Hu et~al.(2018)Hu, Gao, Liu, \protect\BIBand{} Ma}]{hu2018federated}
Hu B, Gao Y, Liu L, Ma H (2018) Federated region-learning: An edge computing
  based framework for urban environment sensing. \emph{2018 IEEE Global
  Communications Conference (GLOBECOM)}, 1--7 (IEEE).

\bibitem[{Hu et~al.(2020)Hu, Shaloudegi, Zhang, \protect\BIBand{}
  Yu}]{hu2020fedmgda+}
Hu Z, Shaloudegi K, Zhang G, Yu Y (2020) Fedmgda+: Federated learning meets
  multi-objective optimization. \emph{arXiv preprint arXiv:2006.11489} .

\bibitem[{Huang et~al.(2020)Huang, Li, Wang, Du, \protect\BIBand{}
  Zhang}]{huang2020fairness}
Huang W, Li T, Wang D, Du S, Zhang J (2020) Fairness and accuracy in federated
  learning. \emph{arXiv preprint arXiv:2012.10069} .

\bibitem[{Jiang et~al.(2020)Jiang, Hu, Hu, Liu, \protect\BIBand{}
  Wang}]{jiang2020bacombo}
Jiang J, Hu L, Hu C, Liu J, Wang Z (2020) Bacombo—bandwidth-aware
  decentralized federated learning. \emph{Electronics} 9(3):440.

\bibitem[{Jiang et~al.(2019)Jiang, Kone{\v{c}}n{\`y}, Rush, \protect\BIBand{}
  Kannan}]{jiang2019improving}
Jiang Y, Kone{\v{c}}n{\`y} J, Rush K, Kannan S (2019) Improving federated
  learning personalization via model agnostic meta learning. \emph{arXiv
  preprint arXiv:1909.12488} .

\bibitem[{Jones \protect\BIBand{} Mewhort(2004)}]{jones2004case}
Jones MN, Mewhort DJ (2004) Case-sensitive letter and bigram frequency counts
  from large-scale english corpora. \emph{Behavior research methods,
  instruments, \& computers} 36(3):388--396.

\bibitem[{Kairouz et~al.(2019)Kairouz, McMahan, Avent, Bellet, Bennis, Bhagoji,
  Bonawitz, Charles, Cormode, Cummings et~al.}]{kairouz2019advances}
Kairouz P, McMahan HB, Avent B, Bellet A, Bennis M, Bhagoji AN, Bonawitz K,
  Charles Z, Cormode G, Cummings R, et~al. (2019) Advances and open problems in
  federated learning. \emph{arXiv preprint arXiv:1912.04977} .

\bibitem[{Karimireddy et~al.(2020)Karimireddy, Kale, Mohri, Reddi, Stich,
  \protect\BIBand{} Suresh}]{karimireddy2020scaffold}
Karimireddy SP, Kale S, Mohri M, Reddi S, Stich S, Suresh AT (2020) Scaffold:
  Stochastic controlled averaging for federated learning. \emph{International
  Conference on Machine Learning}, 5132--5143 (PMLR).

\bibitem[{Khodak et~al.(2021)Khodak, Tu, Li, Li, Balcan, Smith,
  \protect\BIBand{} Talwalkar}]{khodak2021federated}
Khodak M, Tu R, Li T, Li L, Balcan MF, Smith V, Talwalkar A (2021) Federated
  hyperparameter tuning: Challenges, baselines, and connections to
  weight-sharing. \emph{arXiv preprint arXiv:2106.04502} .

\bibitem[{Kontar et~al.(2021)Kontar, Shi, Yue, Chung, Byon, Chowdhury, Jin,
  Kontar, Masoud, Nouiehed et~al.}]{kontar2021internet}
Kontar R, Shi N, Yue X, Chung S, Byon E, Chowdhury M, Jin J, Kontar W, Masoud
  N, Nouiehed M, et~al. (2021) The internet of federated things (ioft).
  \emph{IEEE Access} 9:156071--156113.

\bibitem[{Li \protect\BIBand{} Wang(2019)}]{li2019fedmd}
Li D, Wang J (2019) Fedmd: Heterogenous federated learning via model
  distillation. \emph{arXiv preprint arXiv:1910.03581} .

\bibitem[{Li et~al.(2020{\natexlab{a}})Li, Fan, Tse, \protect\BIBand{}
  Lin}]{li2020review}
Li L, Fan Y, Tse M, Lin KY (2020{\natexlab{a}}) A review of applications in
  federated learning. \emph{Computers \& Industrial Engineering} 106854.

\bibitem[{Li et~al.(2020{\natexlab{b}})Li, Beirami, Sanjabi, \protect\BIBand{}
  Smith}]{li2020tilted}
Li T, Beirami A, Sanjabi M, Smith V (2020{\natexlab{b}}) Tilted empirical risk
  minimization. \emph{arXiv preprint arXiv:2007.01162} .

\bibitem[{Li et~al.(2021)Li, Hu, Beirami, \protect\BIBand{}
  Smith}]{li2021ditto}
Li T, Hu S, Beirami A, Smith V (2021) Ditto: Fair and robust federated learning
  through personalization. \emph{International Conference on Machine Learning}
  .

\bibitem[{Li et~al.(2018)Li, Sahu, Zaheer, Sanjabi, Talwalkar,
  \protect\BIBand{} Smith}]{li2018federated}
Li T, Sahu AK, Zaheer M, Sanjabi M, Talwalkar A, Smith V (2018) Federated
  optimization in heterogeneous networks. \emph{Conference on Machine Learning
  and Systems} .

\bibitem[{Li et~al.(2019{\natexlab{a}})Li, Sanjabi, Beirami, \protect\BIBand{}
  Smith}]{li2019fair}
Li T, Sanjabi M, Beirami A, Smith V (2019{\natexlab{a}}) Fair resource
  allocation in federated learning. \emph{International Conference on Learning
  Representations} .

\bibitem[{Li et~al.(2019{\natexlab{b}})Li, Huang, Yang, Wang, \protect\BIBand{}
  Zhang}]{li2019convergence}
Li X, Huang K, Yang W, Wang S, Zhang Z (2019{\natexlab{b}}) On the convergence
  of fedavg on non-iid data. \emph{arXiv preprint arXiv:1907.02189} .

\bibitem[{Li et~al.(2019{\natexlab{c}})Li, Yang, Wang, \protect\BIBand{}
  Zhang}]{li2019communication}
Li X, Yang W, Wang S, Zhang Z (2019{\natexlab{c}}) Communication-efficient
  local decentralized sgd methods. \emph{arXiv preprint arXiv:1910.09126} .

\bibitem[{Liang et~al.(2020)Liang, Liu, Ziyin, Allen, Auerbach, Brent,
  Salakhutdinov, \protect\BIBand{} Morency}]{liang2020think}
Liang PP, Liu T, Ziyin L, Allen NB, Auerbach RP, Brent D, Salakhutdinov R,
  Morency LP (2020) Think locally, act globally: Federated learning with local
  and global representations. \emph{arXiv preprint arXiv:2001.01523} .

\bibitem[{Lyu et~al.(2020)Lyu, Xu, Wang, \protect\BIBand{}
  Yu}]{lyu2020collaborative}
Lyu L, Xu X, Wang Q, Yu H (2020) Collaborative fairness in federated learning.
  \emph{Federated Learning}, 189--204 (Springer).

\bibitem[{McMahan et~al.(2017)McMahan, Moore, Ramage, Hampson,
  \protect\BIBand{} y~Arcas}]{mcmahan2017communication}
McMahan B, Moore E, Ramage D, Hampson S, y~Arcas BA (2017)
  Communication-efficient learning of deep networks from decentralized data.
  \emph{Artificial Intelligence and Statistics}, 1273--1282 (PMLR).

\bibitem[{Mohri et~al.(2018)Mohri, Rostamizadeh, \protect\BIBand{}
  Talwalkar}]{mohri2018foundations}
Mohri M, Rostamizadeh A, Talwalkar A (2018) \emph{Foundations of machine
  learning} (MIT press).

\bibitem[{Mohri et~al.(2019)Mohri, Sivek, \protect\BIBand{}
  Suresh}]{mohri2019agnostic}
Mohri M, Sivek G, Suresh AT (2019) Agnostic federated learning.
  \emph{International Conference on Machine Learning}, 4615--4625 (PMLR).

\bibitem[{Ramaswamy et~al.(2019)Ramaswamy, Mathews, Rao, \protect\BIBand{}
  Beaufays}]{ramaswamy2019federated}
Ramaswamy S, Mathews R, Rao K, Beaufays F (2019) Federated learning for emoji
  prediction in a mobile keyboard. \emph{arXiv preprint arXiv:1906.04329} .

\bibitem[{Reddi et~al.(2020)Reddi, Charles, Zaheer, Garrett, Rush,
  Kone{\v{c}}n{\`y}, Kumar, \protect\BIBand{} McMahan}]{reddi2020adaptive}
Reddi S, Charles Z, Zaheer M, Garrett Z, Rush K, Kone{\v{c}}n{\`y} J, Kumar S,
  McMahan HB (2020) Adaptive federated optimization. \emph{arXiv preprint
  arXiv:2003.00295} .

\bibitem[{Samsung(2019)}]{phones}
Samsung SfB (2019) Your phone is now more powerful than your pc.
  \url{https://insights.samsung.com/2021/08/19/your-phone-is-now-more-powerful-than-your-pc-3/},
  accessed: 2019-02-19.

\bibitem[{Sattler et~al.(2019)Sattler, Wiedemann, M{\"u}ller, \protect\BIBand{}
  Samek}]{sattler2019robust}
Sattler F, Wiedemann S, M{\"u}ller KR, Samek W (2019) Robust and
  communication-efficient federated learning from non-iid data. \emph{IEEE
  transactions on neural networks and learning systems} 31(9):3400--3413.

\bibitem[{Smith et~al.(2017)Smith, Chiang, Sanjabi, \protect\BIBand{}
  Talwalkar}]{smith2017federated}
Smith V, Chiang CK, Sanjabi M, Talwalkar AS (2017) Federated multi-task
  learning. \emph{Advances in Neural Information Processing Systems},
  4424--4434.

\bibitem[{Wang et~al.(2020{\natexlab{a}})Wang, Sreenivasan, Rajput,
  Vishwakarma, Agarwal, Sohn, Lee, \protect\BIBand{}
  Papailiopoulos}]{wang2020attack}
Wang H, Sreenivasan K, Rajput S, Vishwakarma H, Agarwal S, Sohn Jy, Lee K,
  Papailiopoulos D (2020{\natexlab{a}}) Attack of the tails: Yes, you really
  can backdoor federated learning. \emph{Conference on Neural Information
  Processing Systems} .

\bibitem[{Wang et~al.(2020{\natexlab{b}})Wang, Yurochkin, Sun, Papailiopoulos,
  \protect\BIBand{} Khazaeni}]{wang2020federated}
Wang H, Yurochkin M, Sun Y, Papailiopoulos D, Khazaeni Y (2020{\natexlab{b}})
  Federated learning with matched averaging. \emph{International Conference on
  Learning Representations} .

\bibitem[{Wang et~al.(2019)Wang, Mathews, Kiddon, Eichner, Beaufays,
  \protect\BIBand{} Ramage}]{wang2019federated}
Wang K, Mathews R, Kiddon C, Eichner H, Beaufays F, Ramage D (2019) Federated
  evaluation of on-device personalization. \emph{arXiv preprint
  arXiv:1910.10252} .

\bibitem[{Xu \protect\BIBand{} Lyu(2020)}]{xu2020towards}
Xu X, Lyu L (2020) Towards building a robust and fair federated learning
  system. \emph{arXiv preprint arXiv:2011.10464} .

\bibitem[{Yang et~al.(2020)Yang, Jiang, Shi, \protect\BIBand{}
  Ding}]{yang2020federated}
Yang K, Jiang T, Shi Y, Ding Z (2020) Federated learning via over-the-air
  computation. \emph{IEEE Transactions on Wireless Communications}
  19(3):2022--2035.

\bibitem[{Yu et~al.(2020)Yu, Bagdasaryan, \protect\BIBand{}
  Shmatikov}]{yu2020salvaging}
Yu T, Bagdasaryan E, Shmatikov V (2020) Salvaging federated learning by local
  adaptation. \emph{arXiv preprint arXiv:2002.04758} .

\bibitem[{Yuan \protect\BIBand{} Ma(2020)}]{yuan2020federated}
Yuan H, Ma T (2020) Federated accelerated stochastic gradient descent.
  \emph{Conference on Neural Information Processing Systems} .

\bibitem[{Yue \protect\BIBand{} Kontar(2021)}]{yue2021federated}
Yue X, Kontar RA (2021) Federated gaussian process: Convergence, automatic
  personalization and multi-fidelity modeling. \emph{arXiv preprint
  arXiv:2111.14008} .

\bibitem[{Zafar et~al.(2017)Zafar, Valera, Gomez~Rodriguez, \protect\BIBand{}
  Gummadi}]{zafar2017fairness}
Zafar MB, Valera I, Gomez~Rodriguez M, Gummadi KP (2017) Fairness beyond
  disparate treatment \& disparate impact: Learning classification without
  disparate mistreatment. \emph{Proceedings of the 26th international
  conference on world wide web}, 1171--1180.

\bibitem[{Zhang et~al.(2020{\natexlab{a}})Zhang, Kou, \protect\BIBand{}
  Wang}]{zhang2020fairfl}
Zhang DY, Kou Z, Wang D (2020{\natexlab{a}}) Fairfl: A fair federated learning
  approach to reducing demographic bias in privacy-sensitive classification
  models. \emph{2020 IEEE International Conference on Big Data (Big Data)},
  1051--1060 (IEEE).

\bibitem[{Zhang et~al.(2020{\natexlab{b}})Zhang, Li, Robles-Kelly,
  \protect\BIBand{} Kankanhalli}]{zhang2020hierarchically}
Zhang J, Li C, Robles-Kelly A, Kankanhalli M (2020{\natexlab{b}})
  Hierarchically fair federated learning. \emph{arXiv preprint
  arXiv:2004.10386} .

\bibitem[{Zhang et~al.(2020{\natexlab{c}})Zhang, Zhu, Wang, Yan, Chen,
  \protect\BIBand{} Bao}]{zhang2020federated}
Zhang X, Zhu X, Wang J, Yan H, Chen H, Bao W (2020{\natexlab{c}}) Federated
  learning with adaptive communication compression under dynamic bandwidth and
  unreliable networks. \emph{Information Sciences} 540:242--262.

\bibitem[{Zhao et~al.(2018)Zhao, Li, Lai, Suda, Civin, \protect\BIBand{}
  Chandra}]{zhao2018federated}
Zhao Y, Li M, Lai L, Suda N, Civin D, Chandra V (2018) Federated learning with
  non-iid data. \emph{arXiv preprint arXiv:1806.00582} .

\end{thebibliography}
\bibliographystyle{informs2014}
\end{document}